%% file: 19-441.tex
\documentclass[twoside, 11pt]{article}

\usepackage{jmlr2e_mod}

\usepackage[linesnumbered,ruled]{algorithm2e}
\usepackage{float}
\DontPrintSemicolon

\usepackage{color}
\usepackage{multirow}

\usepackage{amsfonts}
\usepackage{amsmath}

\usepackage{tabularx}%

\usepackage{enumitem}
\usepackage{booktabs,makecell}

\usepackage{tikz}
\usetikzlibrary{positioning}
\usetikzlibrary{arrows.meta}

\usepackage{etoc}

\usepackage{mathtools}

\usepackage{lastpage}

\input{macros}

\jmlrheading{21}{2020}{1-\pageref{LastPage}}{5/20}{5/20}{20-499}{Yuansi Chen, Raaz Dwivedi, Martin Wainwright and Bin Yu}
\ShortHeadings{Fast mixing of Metropolized HMC}{Chen, Dwivedi, Wainwright and Yu}

\begin{document}
\etocdepthtag.toc{mtchapter}
\etocsettagdepth{mtchapter}{subsection}
\etocsettagdepth{mtappendix}{none}


\title{Fast mixing of Metropolized Hamiltonian Monte Carlo:
        Benefits of multi-step gradients}

\author{\name Yuansi Chen \email yuansi.chen@berkeley.edu \\
       \addr Department of Statistics\\
       University of California\\
       Berkeley, CA 94720-1776, USA
       \AND
       \name Raaz Dwivedi \email raaz.rsk@berkeley.edu \\
       \addr Department of Electrical Engineering and Computer Sciences\\
       University of California\\
       Berkeley, CA 94720-1776, USA
      \AND
      \name Martin J. Wainwright \email wainwrig@berkeley.edu\\
      \name Bin Yu \email binyu@berkeley.edu\\
      \addr Department of Statistics\\
      Department of Electrical Engineering and Computer Sciences\\
      University of California\\
      Berkeley, CA 94720-1776, USA
       }

\editor{Mohammad Emtiyaz Khan}

\maketitle

\begin{abstract}
  Hamiltonian Monte Carlo (HMC) is a state-of-the-art Markov chain Monte Carlo sampling algorithm for drawing samples from smooth probability
  densities over continuous spaces. We study the variant most widely
  used in practice, Metropolized HMC with the St\"{o}rmer-Verlet or
  leapfrog integrator, and make two primary contributions.  First, we
  provide a non-asymptotic upper bound on the mixing time of the
  Metropolized HMC with explicit choices of step-size and number of
  leapfrog steps.  This bound gives a precise quantification of the
  faster convergence of Metropolized HMC relative to simpler MCMC
  algorithms such as the Metropolized random walk, or Metropolized
  Langevin algorithm.  Second, we provide a general framework for
  sharpening mixing time bounds of Markov chains initialized at a
  substantial distance from the target distribution over continuous spaces.  We apply this sharpening device to the Metropolized random walk and Langevin
  algorithms, thereby obtaining improved mixing time bounds from a
  non-warm initial distribution.

\end{abstract}



\section{Introduction}
\label{sec:introduction}

Markov Chain Monte Carlo (MCMC) methods date back to the seminal work
of~\cite{metropolis1953equation}, and are the method
of choice for drawing samples from high-dimensional
distributions. They are widely used in practice, including in Bayesian
statistics for exploring posterior
distributions~\citep{carpenter2017stan,smith2014mamba}, in
simulation-based methods for reinforcement learning, and in image
synthesis in computer vision, among other areas.  Since their origins
in the 1950s, many MCMC algorithms have been introduced, applied and
studied; we refer the reader to the handbook by~\cite{brooks2011handbook}
for a survey of known results and contemporary developments.

There are a variety of MCMC methods for sampling from target
distributions with smooth
densities~\citep{Robert,roberts2004general,roberts2002langevin,
  brooks2011handbook}.  Among them, the method of Hamiltonian Monte
Carlo (HMC) stands out among practitioners: it is the default sampler for sampling from complex distributions
in many popular software packages, including
Stan~\citep{carpenter2017stan}, Mamba~\citep{smith2014mamba}, and
Tensorflow~\citep{tensorflow2015-whitepaper}.  We refer the reader to
the papers~\citep{neal2011mcmc,hoffman2014no,durmus2017convergence} for
further examples and discussion of the HMC method.  There are a number
of variants of HMC, but the most popular choice involves a combination
of the leapfrog integrator with Metropolis-Hastings correction.
Throughout this paper, we reserve the terminology HMC to refer to this
particular Metropolized algorithm.  The idea of using Hamiltonian
dynamics in simulation first appeared in~\cite{alder1959studies}. \cite{duane1987hybrid} introduced MCMC with Hamiltonian dynamics,
and referred to it as Hybrid Monte Carlo. The algorithm was further
refined by~\cite{Neal1994}, and later re-christened in the statistics
community as Hamiltonian Monte Carlo.  We refer the reader to~\cite{neal2011mcmc} for an illuminating overview of the history
of HMC and a discussion of contemporary work.


\subsection{Past work on HMC}

While HMC enjoys fast convergence in practice, a theoretical
understanding of this behavior remains incomplete. Some intuitive
explanations are based on its ability to maintain a constant
asymptotic accept-reject rate with large step-size
\citep{creutz1988global}.  Others suggest, based on intuition from the continuous-time limit of the Hamiltonian
dynamics, that HMC can suppress random walk behavior using
momentum~\citep{neal2011mcmc}.  However, these intuitive arguments do not provide rigorous
or quantitative justification for the fast convergence of the
discrete-time HMC used in practice.

More recently, general asymptotic conditions under which HMC will or
will not be geometrically ergodic have been established in some recent
papers~\citep{durmus2017convergence,livingstone2016geometric}.  Other
work has yielded some insight into the mixing properties of different
variants of HMC, but it has focused mainly on \emph{unadjusted}
versions of the algorithm. \cite{mangoubi2017rapid}
and~\cite{mangoubi2018dimensionally} study versions of unadjusted HMC
based on Euler discretization or leapfrog integrator (but omitting the
Metropolis-Hastings step), and provide explicit bounds on the mixing
time as a function of dimension $\dims$, condition number $\condition$
and error tolerance $\errorparam> 0$. \cite{lee2018algorithmic}
studied an extended version of HMC that involves applying an ordinary
differential equation (ODE) solver; they established bounds with
sublinear dimension dependence, and even polylogarithmic for certain
densities (e.g., those arising in Bayesian logistic regression). The
mixing time for the same algorithm is further refined in recent work
by \cite{chen2019optimal}. In a similar spirit, \cite{LeeV18} studied
the Riemannian variant of HMC (RHMC) with an ODE solver focusing on
sampling uniformly from a polytope. While their result could be
extended to log-concave sampling, the practical implementation of
their ODE solver for log-concave sampling is unclear. Moreover, it
requires a regularity condition on all the derivatives of density. One
should note that such unadjusted HMC methods behave differently from
the Metropolized version most commonly used in practice.  In the
absence of the Metropolis-Hastings correction, the resulting Markov
chain no longer converges to the correct target distribution, but
instead exhibits a persistent bias, even in the limit of infinite
iterations.  Consequently, the analysis of such sampling methods
requires controlling this bias; doing so leads to mixing times that
scale polynomially in $1/\errorparam$, in sharp contrast with the
$\log(1/\errorparam)$ that is typical for Metropolis-Hastings
corrected methods.

Most closely related to our paper is the recent work by~\cite{bou2018coupling}, which studies the same Metropolized HMC
algorithm that we analyze in this paper.  They use coupling
methods to analyze HMC for a class of distributions that are strongly
log-concave outside of a compact set.  In the strongly log-concave
case, they prove a mixing-time bound that scales at least as $\dims^{3/2}$ in the dimension $\dims$.  We note that with a ``warm''
initialization, this dimension dependence grows more quickly than known bounds for the MALA algorithm~\citep{dwivedi2018log,eberle2014error}, and so does not explain the superiority of HMC in practice.

In practice, it is known that Metropolized HMC is fairly sensitive to the choice of its parameters---namely the step-size $\step$ used in the discretization scheme, and the number of leapfrog steps $\internsteps$.  At one extreme, taking a single leapfrog step $\internsteps = 1$, the
algorithm reduces to the Metropolis adjusted Langevin algorithm
(MALA). More generally, if too few leapfrog steps are taken, HMC
is likely to exhibit a random walk behavior similar to that of MALA.  At the
other extreme, if $\internsteps$ is too large, the leapfrog steps tend
to wander back to a neighborhood of the initial state, which leads to
wasted computation as well as slower
mixing~\citep{betancourt2014optimizing}. In terms of the step size
$\step$, when overly large step size is chosen, the discretization
diverges from the underlying continuous dynamics leading to a drop in the
Metropolis acceptance probability, thereby slowing down the mixing of the algorithm.  On the other hand, an overly small choice of $\step$ does
not allow the algorithm to explore the state space rapidly enough.
While it is difficult to characterize the necessary and sufficient conditions on $\internsteps$ and $\step$ to ensure fast convergence, many works suggest the choice of these two parameters based on the necessary conditions such as maintaining a constant acceptance rate~\citep{chen2001exploring}. For instance, \cite{beskos2013optimal} showed that in the simplified scenario of target density with independent, identically distributed components, the number of leapfrog steps should scale as $\dims^{1/4}$ to achieve a constant acceptance rate.
Besides, instead of setting the two parameters explicitly, various automatic strategies for tuning these two parameters have been
proposed~\citep{wang2013adaptive,hoffman2014no,robert2019faster}. Despite being introduced via heuristic arguments and with additional computational cost, these methods, such as the No-U-Turn (NUTS) sampler~\citep{hoffman2014no}, have shown promising empirical evidence of its effectiveness on a wide range of simple target distributions.


\subsection{Past work on mixing time dependency on initialization}

Many proof techniques for the convergence of continuous-state Markov
chains are inspired by the large body of work on discrete-state Markov
chains; for instance, see the
surveys~\citep{lovasz1993randomsurvey,aldous2002reversible} and
references therein. Historically, much work has been devoted to
improving the mixing time dependency on the initial distribution. For
discrete-state Markov chains, \cite{diaconis1996logarithmic} were the
first to show that the logarithmic dependency of the mixing time of a
Markov chain on the warmness parameter\footnote{See
  equation~\eqref{eq:def_warm} for a formal definition.}  of the
starting distribution can be improved to doubly logarithmic.  This
improvement---from logarithmic to doubly logarithmic---allows for a
good bound on the mixing time even when a good starting distribution
is not available.  The innovation underlying this improvement is the
use of log-Sobolev inequalities in place of the usual isoperimetric
inequality. Later, closely related ideas such as average
conductance~\citep{lovasz1999faster,kannan2006blocking}, evolving
sets~\citep{morris2005evolving} and spectral
profile~\citep{goel2006mixing} were shown to be effective for reducing
dependence on initial conditions for discrete space chains. Thus far,
only the notion of average
conductance~\citep{lovasz1999faster,kannan2006blocking} has been
adapted to continuous-state Markov chains so as to sharpen the mixing
time analysis of the Ball walk~\citep{lovasz1990ballwalk}.


\subsection{Our contributions}

This paper makes two primary contributions.  First, we provide a
non-asymptotic upper bound on the mixing time of the Metropolized HMC
algorithm for smooth densities (see
Theorem~\ref{thm:hmc_mixing_general}).  This theorem applies to the
form of Metropolized HMC (based on the leapfrog integrator) that is
most widely used in practice.  To the best of our knowledge,
Theorem~\ref{thm:hmc_mixing_general} is the first rigorous
confirmation of the faster non-asymptotic convergence of the
Metropolized HMC as compared to MALA and other simpler
Metropolized algorithms.\footnote{As noted earlier, previous results by~\cite{bou2018coupling} on Metropolized HMC do not establish
  that it mixes more rapidly than MALA.} Other related works on HMC
consider either its unadjusted version (without accept-reject step)
with different
integrators~\citep{mangoubi2017rapid,mangoubi2018dimensionally} or the
HMC based on an ODE solver~\citep{lee2018algorithmic,LeeV18}.  While
the dimension dependency for these algorithms is usually better than
MALA, they have polynomial dependence on the inverse error tolerance
$1/\errorparam$ while MALA's mixing time scales as
$\log(1/\errorparam)$.  Moreover, our direct analysis of the
Metropolized HMC with a leapfrog integrator provides explicit choices
of the hyper-parameters for the sampler, namely, the step-size and the number of leapfrog updates in each step. Our theoretical choices of
the hyper-parameters could potentially provide guidelines for
parameter tuning in practical HMC implementations

Our second main contribution is formalized in
Lemmas~\ref{prop:mixing_bound_using_conductanceprofile}
and~\ref{prop:conductanceprofile_via_overlaps}: we develop results
based on the conductance profile in order to prove quantitative
convergence guarantees general continuous-state space Markov
chains. Doing so involves non-trivial extensions of ideas from
discrete-state Markov chains to those in continuous state spaces. Our
results not only enable us to establish the mixing time bounds for HMC
with different classes of target distributions, but also allow
simultaneous improvements on mixing time bounds of several Markov
chains (for general continuous-state space) when the starting
distribution is far from the stationary distribution.
Consequentially, we improve upon previous mixing time bounds for
Metropolized Random Walk (MRW) and MALA~\citep{dwivedi2018log}, when the starting distribution is not \emph{warm} with respect to the
target distribution (see Theorem~\ref{thm:mala_mrw_mixing}).

While this high-level road map is clear, a number of technical
challenges arise en route in particular in controlling the conductance
profile of HMC.  The use of multiple gradient steps in each iteration of HMC helps it mix
faster but also complicates the analysis; in particular, a key step is
to control the overlap between the transition distributions of HMC chain at two nearby points; doing so requires a delicate argument (see
Lemma~\ref{lem:transition_closeness} and
Section~\ref{sub:lem:transition_closeness} for further details).

\begin{figure}[ht]
\begin{tikzpicture}[>=stealth,every node/.style={shape=rectangle,draw,rounded corners, minimum width=4.5cm,},]
    \node[very thick, fill=red!30] (t1) {\begin{tabular}{c} Theorem~\ref{thm:hmc_mixing_general}
    \\ general mixing-time  \\ result for HMC \end{tabular}};
    \node[very thick, fill=orange!30] (l1) [right=of
    t1]{\begin{tabular}
    {c} Lemma~\ref{prop:mixing_bound_using_conductanceprofile} \\ mixing-time bound \\ via conductance profile \end{tabular}} ;
    \node[very thick, fill=orange!30] (l2) [below=of l1]{\begin{tabular}{c} Lemma~\ref{prop:conductanceprofile_via_overlaps}
    \\ conductance profile \\ via overlap bounds \end{tabular}};
    \node[very thick] (l3) [below=of t1]{\begin{tabular}{c} Lemma~\ref{lem:transition_closeness}
    \\ overlap bounds \\ for HMC\end{tabular}};
    \node[very thick] (l4) [right=of l2]{\begin{tabular}{c}
    known overlap bounds\\ for MALA and MRW \\
    \cite{dwivedi2018log} \end{tabular}};
    \node[very thick] (c1) [above=of t1]{\begin{tabular}{c}
    Corollaries~\ref{cor:hmc_mixing_sc_fixedK}~and~\ref{cor:hmc_mixing_wc}
    \\ HMC mixing-time for \\ different target distributions \end{tabular}};
    \node[very thick, fill=red!30] (t2) [right=of l1]{\begin{tabular}{c} Theorem~\ref{thm:mala_mrw_mixing}
    \\ improved  mixing-time\\  for MALA and MRW\end{tabular}};
    \draw[->, line width=.6mm] (l1) to[out=180,in=0](t1);
    \draw[->,  line width=.6mm] (l2) to[out=180,in=353] (t1);
    \draw[->,  line width=.6mm] (l3) to[out=0,in=345] (t1);
    \draw[->, dashed,  line width=0.6mm] (t1) to[out=90,in=270] (c1);
    \draw[->, dashed,  line width=0.6mm] (l1) to[out=0,in=180] (t2);
    \draw[->, dashed,  line width=0.6mm] (l2) to[out=0,in=187] (t2);
    \draw[->, dashed,  line width=0.6mm] (l4) to[out=180,in=195] (t2);
\end{tikzpicture}
  \caption{High-level sketch diagram of the results and the proof techniques
  developed in this paper.
  Lemmas~\ref{prop:mixing_bound_using_conductanceprofile}
  and \ref{prop:conductanceprofile_via_overlaps} of this paper provide general
  machinery to develop mixing time bounds for several Markov chains, provided
  that we have a control over the overlap bounds of the respective transition
  distributions. Establishing these overlap
  bounds for Metropolized HMC requires non-trivial machinery which we develop
  in Lemma~\ref{lem:transition_closeness}. Putting Lemmas~\ref{prop:mixing_bound_using_conductanceprofile},
  \ref{prop:conductanceprofile_via_overlaps} and \ref{lem:transition_closeness}
  together yields the general mixing time bound for HMC in
  Theorem~\ref{thm:hmc_mixing_general}. Moreover, Theorem~\ref{thm:hmc_mixing_general}
  easily implies a faster mixing time bound for HMC over MALA and MRW for different
  target distributions as shown in Corollaries~\ref{cor:hmc_mixing_sc_fixedK}
  and \ref{cor:hmc_mixing_wc}, and summarized in Table~\ref{tab:mixing_times_all}.
  On the other hand, Lemmas~\ref{prop:mixing_bound_using_conductanceprofile}
  and \ref{prop:conductanceprofile_via_overlaps} in conjunction with
  overlap bounds from~\cite{dwivedi2018log} immediately
  yield Theorem~\ref{thm:mala_mrw_mixing} that provides improved mixing
  time
  bounds for MALA and MRW from a non-warm start.
  }
    \label{fig:proof_sketch}
\end{figure}
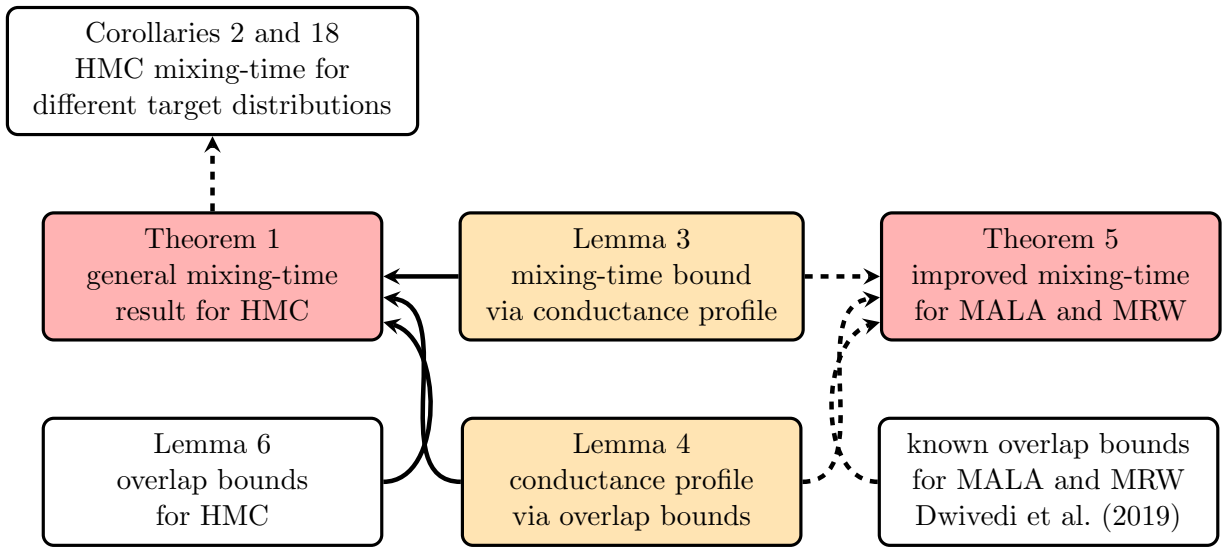

Table~\ref{tab:mixing_times_all} provides an informal summary of our
mixing time bounds of HMC and how they compare with known bounds for
MALA when applied to log-concave target distributions.  From the
table, we see that Metropolized HMC takes fewer gradient evaluations
than MALA to mix to the same accuracy for log-concave distributions.  Note
that our current analysis establishes logarithmic dependence on the
target error $\errorparam$ for strongly-log-concave as well as for a
sub-class of weakly log-concave distributions.\footnote{For comparison with previous results on unadjusted HMC or
  ODE based HMC refer to the discussion after
  Corollary~\ref{cor:hmc_mixing_sc_fixedK} and
  Table~\ref{tab:mixing_times_with_poly} in
  Appendix~\ref{sub:comparison_with_guarantees_for_other_versions_of_hmc}.}
Moreover, in Figure~\ref{fig:proof_sketch} we provide a sketch-diagram
of how the different results in this paper interact to provide mixing time
bounds for different Markov chains.

\begin{table}[ht]
    \centering
    {\renewcommand{\arraystretch}{.5}
    \begin{tabular}{cccc}
        \toprule
        & \thead{\bf Strongly
        log-concave} &\multicolumn{2}{c}{\thead{\bf \footnotesize
        Weakly log-concave}} \\ \thead{ \bf Sampling algorithm} &
    \thead{\bf Assumption~\ref{itm:assumptionB}
      ($\condition\ll\dims$)} & \thead{\bf
      Assumption~\ref{itm:assumptionC}} & \thead{ \bf
      Assumption~\ref{itm:assumptionD}} \\ \midrule \thead{ MALA
      \\ (improved bound in \\ Thm~\ref{thm:mala_mrw_mixing} in this
      paper)} &
    \thead{$\displaystyle\dims\condition\log\frac{1}{\errorparam}$\\[3pt]
    \cite{dwivedi2018log}}
    &
    \thead{$\displaystyle\frac{\dims^{2}}{\errorparam^{\frac{3}{2}}}\log
      \frac{1} {\errorparam}$\\[3pt] \cite{dwivedi2018log}} &
    \thead{$\displaystyle
      \dims^{\frac{3}{2}}\log\frac{1}{\errorparam}$\\[3pt] \cite{mangoubi2019nonconvex}}
    \\[7mm] \thead{Metropolized HMC with \\ leapfrog integrator
      \\ \ [this paper]} &
    \thead{$\displaystyle\dims^{\frac{11}{12}}\condition\log\frac{1}{\errorparam}$
    \\[3pt] {(Corollary~\ref{cor:hmc_mixing_sc_fixedK})}} &
    \thead{$\displaystyle\frac{\dims^{\frac{11}{6}}}{\errorparam}
      \log\frac{1}{\errorparam}$\\[3pt] {(Corollary~\ref{cor:hmc_mixing_wc})}}
    & \thead{$\displaystyle
      \dims^{\frac{4}{3}}\log\frac{1}{\errorparam}$
      \\[3pt] {(Corollary~\ref{cor:hmc_mixing_wc})}} \\ [2mm]
    \bottomrule
    \end{tabular}
    }
    \caption{Comparisons of the number of gradient evaluations needed
      by MALA and Metropolized HMC with leapfrog integrator from a
      \emph{warm start} to obtain an $\errorparam$-accurate sample in
      TV distance from a log-concave target distribution on
      $\real^\dims$.  The second column corresponds to strongly
      log-concave densities with condition number $\condition$, and
      the third and fourth column correspond to weakly log-concave
      densities satisfying certain regularity conditions.}
    \label{tab:mixing_times_all}
\end{table}


\paragraph{Organization:}
\label{par:organization_}

The remainder of the paper is organized as follows.
Section~\ref{sec:problem_set_up} is devoted to background on the idea
of Monte Carlo approximation, Markov chains and MCMC algorithms, and
the introduction of the MRW, MALA and HMC algorithms.
Section~\ref{sec:main_results} contains our main results on mixing
time of HMC in
Section~\ref{sub:mixing_time_bounds_for_hmc_for_general_regular_target_distributions},
followed by the general framework for obtaining sharper mixing time
bounds in
Section~\ref{sub:mixing_time_of_a_markov_chain_with_conductance_profile}
and its application to MALA and MRW in
Section~\ref{sub:improved_warmness_dependency_for_mala_and_mrw}.  In
Section~\ref{sec:numerical_experiments}, we describe some numerical
experiments that we performed to explore the sharpness of our
theoretical predictions in some simple scenarios.  In
Section~\ref{sec:proof_of_the_main_theorem}, we prove
Theorem~\ref{thm:hmc_mixing_general} and
Corollary~\ref{cor:hmc_mixing_sc_general}, with the proofs of
technical lemmas and other results deferred to the appendices.  We
conclude in Section~\ref{sec:discussion} with a discussion of our
results and future directions.


\paragraph{Notation:}

For two real-valued sequences $\{a_n\}$ and $\{b_n\}$, we write $a_n =
O(b_n)$ if there exists a universal constant $c > 0$ such that $a_n
\leq c b_n$. We write $a_n = \Ot(b_n)$ if $a_n \leq c_n b_n$, where
$c_n$ grows at most poly-logarithmically in $n$.  We use
$[\internsteps]$ to denote the integers from the set $\braces{1, 2,
  \ldots, \internsteps}$. We denote the Euclidean norm on
$\real^\dims$ as $\vecnorm{\, \cdot \, }{2}$.  We use $\statespace$ to
denote the (general) state space of a Markov chain. We denote
$\borel(\statespace)$ as the Borel $\sigma$-algebra of the state space
$\statespace$.  Throughout we use the notation $c$, $c_1$, $c_2$ to
denote universal constants.  For a function $\targetf: \real^\dims
\rightarrow \real$ that is three times differentiable, we represent
its derivatives at $x\in \real^\dims$ by $\gradf(x) \in \real^\dims$,
$\hessf(x) \in \real^{\dims \times \dims}$ and $\thirdf(x) \in
\real^{\dims^3}$. Here
\begin{align*}
  \brackets{\gradf(x)}_i = \frac{\partial}{\partial x_i}
  \targetf(x),\quad \brackets{\hessf(x)}_{i, j} =
  \frac{\partial^2}{\partial x_i \partial x_j} \targetf(x),\quad
  \brackets{\thirdf}_{i,j,k} = \frac{\partial^3}{\partial x_i \partial
    x_j \partial x_k } f(x).
\end{align*}
Moreover for a square matrix $A$, we define its $\ell_2$-operator norm
$\opnorm{A} \defn \max \limits_{\vecnorm{v}{2}=1}\vecnorm{Av} {2}$.


\section{Background and problem set-up}
\label{sec:problem_set_up}

In this section, we begin by introducing background on Markov chain
Monte Carlo in Section~\ref{sub:monte_carlo_markov_chain_methods},
followed by definitions and terminology for Markov chains in
Section~\ref{sub:markov_chain_terminology}.  In
Section~\ref{sub:hmc_and_related_sampling_algorithms}, we describe
several MCMC algorithms, including the Metropolized random walk (MRW),
the Metropolis-adjusted Langevin algorithm (MALA), and the
Metropolis-adjusted Hamiltonian Monte Carlo (HMC) algorithm.  Readers
familiar with the literature may skip directly to the
Section~\ref{sec:main_results}, where we set up
and state our main results.


\subsection{Monte Carlo Markov chain methods}
\label{sub:monte_carlo_markov_chain_methods}

Consider a distribution $\target$ equipped with a density
$\targetdensity: \statespace \rightarrow \real_+$, specified
explicitly up to a normalization constant as follows
\begin{align}
\label{EqnDefnTarget}
  \targetdensity(x) \propto e^{-\targetf(x)}.
\end{align}
A standard computational task is to estimate the expectation of some
function \mbox{$g: \statespace \rightarrow \real$}---that is, to
approximate $\target(g) = \Exs_{\targetdensity}\brackets{g(X)} =
\int_{\statespace} g(x) \targetdensity(x) dx$.  In general, analytical
computation of this integral is infeasible.  In high dimensions,
numerical integration is not feasible either, due to the well-known
curse of dimensionality.

A Monte Carlo approximation to $\target(g)$ is based on access to a
sampling algorithm that can generate i.i.d. random variables $Z_i \sim
\targetdensity$ for $i = 1, \ldots, N$.  Given such samples, the
random variable $\widehat{\target}(g) \defn \frac{1}{N} \sum_{i=1}^N
g(Z_i)$ is an unbiased estimate of the quantity $\target(g)$, and has
its variance proportional to $1/N$.  The challenge of implementing
such a method is drawing the i.i.d. samples $Z_i$.  If
$\targetdensity$ has a complicated form and the dimension $\dims$ is
large, it is difficult to generate i.i.d. samples from
$\targetdensity$. For example, rejection
sampling~\citep{gilks1992adaptive}, which works well in low dimensions,
fails due to the curse of dimensionality.

The Markov chain Monte Carlo (MCMC) approach is to construct a Markov chain on $\statespace$ that starts from some easy-to-simulate initial
distribution $\initial$, and converges to $\targetdensity$ as its
stationary distribution.  Two natural questions are: (i) methods for
designing such Markov chains; and (ii) how many steps will the Markov chain
take to converge close enough to the stationary distribution?  Over
the years, these questions have been the subject of considerable
research; for instance, see the
reviews by~\cite{tierney1994markov,smith1993bayesian,roberts2004general}
and references therein. In this paper, we are particularly interested
in comparing three popular Metropolis-Hastings adjusted Markov chains
sampling algorithms (MRW, MALA, HMC).  Our primary goal is to tackle
the second question for HMC, in particular via establishing its
concrete non-asymptotic mixing-time bound and thereby characterizing
how HMC converges faster than MRW and MALA.


\subsection{Markov chain basics}
\label{sub:markov_chain_terminology}

Let us now set up some basic notation and definitions on Markov chains
that we use in the sequel.  We consider \emph{time-homogeneous} Markov
chains defined on a measurable state space $\parenth{\statespace,
  \borel(\statespace)}$ with a transition kernel
$\transprob: \statespace \times \borel(\statespace) \rightarrow
\real_+$. By definition, the transition kernel satisfies the following
properties:
\begin{align*}
\transprob(x, dy) \geq 0,\quad \text{for all } x \in \statespace,
\quad \text{and} \quad
\int_{y \in \statespace} \transprob(x, dy) dy = 1 \quad \text{for all
} x \in \statespace.
\end{align*}
The $k$-step transition kernel $\transprob^k$ is defined recursively
as $\transprob^{k+1}(x, dy) = \int_{z \in \statespace} \transprob^k(x,
dz) \transprob(z, dy) dz$.

The Markov chain is \emph{irreducible} means that for all $x, y \in
\statespace$, there is a natural number $k > 0$ such that $
\transprob^k (x, dy) > 0$.  We say that a Markov chain satisfies the
\emph{detailed balance condition} if
\begin{align}
\label{eq:def_reversible}
  \targetdensity(x) \transprob(x, dy)dx = \targetdensity(y)
  \transprob(y, dx)dy \quad\mbox{for all $x, y \in \statespace$}.
\end{align}
Such a Markov chain is also called \emph{reversible}.
Finally, we say that a probability measure $\target$ with density $\targetdensity$
on $\statespace$ is \emph{stationary} (or \emph{invariant}) for a Markov
chain with the transition kernel $\transprob$ if
\begin{align*}
  \int_{x \in \statespace} \targetdensity(x) \transprob(y, dx) =
  \targetdensity (y) \quad \text{for all } y \in \statespace.
\end{align*}


\paragraph{Transition operator:} 
\label{par:transition_operator_}
We use $\transition$ to denote the transition operator of the Markov
chain on the space of probability measures with state space
$\statespace$. In simple words, given a distribution $\initial$ on the
current state of the Markov chain, $\transition(\initial)$ denotes the
distribution of the next state of the chain.  Mathematically, we have
$\transition(\initial)(A) =\int_{\statespace} \transprob (x, A)
\initial (x) dx$ for any $A \in \borel (\statespace)$. In an analogous
fashion, $\transition^k$ stands for the $k$-step transition operator.
We use $\transition_x$ as the shorthand for $\transition(\dirac_x)$,
the \emph{transition distribution at $x$}; here $\dirac_x$ denotes the
Dirac delta distribution at $x \in
 \statespace$.
Note that by definition $\transition_x=\transprob (x, \cdot)$.


\paragraph{Distances between two distributions:}
\label{par:convergence_criterion_}

In order to quantify the convergence of the Markov chain, we study the
mixing time for a class of distances denoted $\Ell_\lp$ for $\lp \geq
1$.  Letting $Q$ be a distribution with density $q$, its
$\Ell_\lp$-divergence with respect to the positive density $\nu$ is
defined as
\begin{subequations}
\label{eq:convergence}
\begin{align}
\label{eq:l_p_distance}
d_{\lp}(Q, \nu) = \parenth{\int_{\statespace}
  \abss{\frac{q(x)}{\nu(x)} - 1}^\lp\nu(x) dx}^{\frac{1}{\lp}}.
\end{align}
Note that for $\lp=2$, we get the $\chi^2$-divergence. For $\lp = 1$,
the distance $d_{1}(Q, \nu)$ represents two times the total variation
distance between $Q$ and $\nu$. In order to make this distinction
clear, we use $\tvdist{Q}{\nu}$ to denote the total variation
distance.


\paragraph{Mixing time of a Markov chain:}
\label{par:mixing_time_of_markov_chains_}

Consider a Markov chain with initial distribution $\initial$,
transition operator $\transition$ and a target distribution $\target$
with density $\targetdensity$. Its $\Ell_\lp$ mixing time with respect
to $\target$ is defined as follows:
\begin{align}
\label{eq:tmix_defn}
  \TMIX_{\lp}(\errorparam; \initial) = \inf\braces{k \in \naturalnum
    \mid d_{\lp}\parenth{\transition^k(\initial), \target} \leq
    \errorparam}.
\end{align}
where $\errorparam > 0$ is an error tolerance.  Since distance
$d_{\lp}(Q, \target)$ increases as $\lp$ increases, we have
\begin{align}
\label{eq:mixing_time_relation}
  \TMIX_{\lp}(\errorparam; \initial) \leq \TMIX_{\lp'}(\errorparam; \initial)
  \quad\text{for any }\quad \lp' \geq \lp \geq 1.
\end{align}
\end{subequations}


\paragraph{Warm initial distribution:}
\label{par:initial_distribution_}

We say that a Markov chain with state space $\statespace$ and
stationary distribution $\target$ has a $\warmparam$-\textit{warm start} if its
initial distribution $\initial$ satisfies
\begin{align}
\label{eq:def_warm}
\sup_{S \in \borel(\statespace)} \frac{\initial(S)}{\target(S)} \leq
\warmparam,
\end{align}
where $\borel(\statespace)$ denotes the Borel $\sigma$-algebra of the
state space $\statespace$. For simplicity, we say that $\initial$ is a
warm start if the warmness parameter $\warmparam$ is a small constant
(e.g., $\warmparam$ does not scale with dimension $\dims$).


\paragraph{Lazy chain:}
\label{par:lazy_chain_}
We say that the Markov chain is $\lazyparam$-\textit{lazy} if, at each iteration, the chain is forced to stay at the previous iterate with
probability~$\lazyparam$.  We study $\frac{1}{2}$-lazy chains in this
paper. In practice, one is not likely to use a lazy chain (since the
lazy steps slow down the convergence rate by a constant factor);
rather, it is a convenient assumption for theoretical
analysis of the mixing rate up to constant factors.\footnote{Any lazy (time-reversible) chain is always
  aperiodic and admits a unique stationary distribution. For more
  details, see the survey~\citep{vempala2005geometric} and references
  therein.}


\subsection{From Metropolized random walk to HMC}
\label{sub:hmc_and_related_sampling_algorithms}

In this subsection, we provide a brief description of the popular
algorithms used for sampling from the space $\statespace=\real^\dims$.
We start with the simpler zeroth-order Metropolized random walk (MRW),
followed by the single-step first-order Metropolis adjusted Langevin
algorithm (MALA) and finally discuss the Hamiltonian Monte Carlo (HMC)
algorithm.

\subsubsection{MRW and MALA algorithms}

One of the simplest Markov chain algorithms for sampling from a
density of the form~\eqref{EqnDefnTarget} defined on $\real^\dims$ is
the Metropolized random walk (MRW).  Given state $x_i \in
\real^\dims$ at iterate $i$, it generates a new proposal vector $z_{i + 1} \sim
\Normal(x_i, 2\step \Ind_\dims)$, where $\step > 0$ is a step-size
parameter.\footnote{The factor $2$ in the step-size definition is a
  convenient notational choice so as to facilitate comparisons with
  other algorithms.}  It then decides to accept or reject $z_{i+1}$
using a Metropolis-Hastings correction; see Algorithm~\ref{algo:mrw}
for the details.  Note that the MRW algorithm uses information about
the function $f$ only via querying function values, but not the
gradients.

The Metropolis-adjusted Langevin algorithm (MALA) is a natural
extension of the MRW algorithm: in addition to the function value
$f(\cdot)$, it also assumes access to its gradient $\nabla f(\cdot)$
at any state $x \in \real^\dims$.  Given state $x_i$ at iterate $i$,
it observes $\parenth{f(x_i), \gradf(x_i)}$ and then generates a new proposal $z_{i+1} \sim \Normal(\displaystyle x_{i} - \step
\gradf(x_{i}), 2\step \Ind_\dims)$, followed by a suitable
Metropolis-Hastings correction; see Algorithm~\ref{algo:mala} for the
details.  The MALA algorithm has an interesting connection to the
Langevin diffusion, a stochastic process whose evolution is
characterized by the stochastic differential equation (SDE)
\begin{align}
  \label{eq:Langevin_SDE}
  dX_t = -\gradf(X_t) dt + \sqrt{2} dW_t.
\end{align}
The MALA proposal can be understood as the Euler-Maruyama
discretization of the SDE~\eqref{eq:Langevin_SDE}.

\begin{algorithm}[ht]
\footnotesize
  \KwIn{Step size $\step > 0$ and a sample $x_0$ from a starting
    distribution $\initial$} \KwOut{Sequence $x_1, x_2,\ldots$}
  \For{$i=0, 1, \ldots $}
  {%
      \textbf{Proposal step}: \emph{Draw} $z_{i+1} \sim \Normal(x_i, 2\step
      \Ind_\dims)$  \\
       \textbf{Accept-reject step}:\\
      \quad compute $\alpha_{i+1} \gets \displaystyle\min \braces{1,
          \frac{\exp\parenth{-f(z_{i+1})}}{\exp\parenth{-f(x_{i})}}}$\\
        \quad With probability $\alpha_{i+1}$ \emph{accept} the proposal:
        $x_{i+1} \gets z_{i+1} $\\
        \quad With probability $1-\alpha_{i+1}$ \emph{reject} the proposal:
        $x_{i+1} \gets x_i$
  }
  \caption{Metropolized Random Walk (MRW)}
  \label{algo:mrw}
\end{algorithm}

\begin{algorithm}[ht]
\footnotesize
  \KwIn{Step size $\step$ and a sample $x_0$ from a starting
    distribution $\initial$} \KwOut{Sequence $x_1, x_2,\ldots$}
  \For{$i=0, 1, \ldots $}
    {%
      \textbf{Proposal step}: \emph{Draw} $z_{i+1} \sim \Normal(\displaystyle
      x_{i} - \step \gradf(x_{i}), 2\step
      \Ind_\dims)$  \\
       \textbf{Accept-reject step}:\\
      \quad compute $\alpha_{i+1} \gets \displaystyle\min \braces{1,
          \frac{\exp\parenth{-f(z_{i+1})-\vecnorm{x_{i}-z_{i+1}+\step\gradf(z_{i+1})}{2}^2/4\step}}{\exp\parenth{-f(x_{i})-\vecnorm{z_{i+1}-x_i+\step\gradf(x_i)}{2}^2/4\step}}}$\\
        \quad With probability $\alpha_{i+1}$ \emph{accept} the proposal:
        $x_{i+1} \gets z_{i+1} $\\
        \quad With probability $1-\alpha_{i+1}$ \emph{reject} the proposal:
        $x_{i+1} \gets x_i$
  }
  \caption{Metropolis adjusted Langevin algorithm (MALA)}
  \label{algo:mala}
\end{algorithm}


\subsubsection{HMC sampling}

The HMC sampling algorithm from the physics literature was introduced to the statistics literature
by Neal; see his survey~\citep{neal2011mcmc} for the historical
background.  The method is inspired by Hamiltonian dynamics, which
describe the evolution of a state vector $\hmcstate(t) \in \real^\dims$ and
its momentum $\hmcnoise(t) \in \real^\dims$ over time $t$ based on a Hamiltonian function $\hamiltonian: \real^\dims \times \real^\dims \rightarrow
\real$ via Hamilton's
equations:
\begin{align}
  \label{eq:hamilton_equations}
  \frac{d\hmcstate}{dt}(t) = \frac{\partial \hamiltonian}{\partial \hmcnoise} (\hmcnoise(t),
  \hmcstate(t)), \quad \mbox{and} \quad \frac{d\hmcnoise}{dt}(t) = - \frac{\partial
    \hamiltonian}{\partial \hmcstate} (\hmcnoise(t), \hmcstate(t)).
\end{align}
A straightforward calculation using the chain rule shows that the
Hamiltonian remains invariant under these dynamics---that is,
$\hamiltonian(\hmcnoise(t), \hmcstate(t)) = C$ for all $t \in \real$.  A typical choice of the
Hamiltonian $\hamiltonian: \real^\dims \times \real^\dims \rightarrow
\real$ is given by
\begin{align}
  \label{eq:hamiltonian}
  \hamiltonian(\hmcnoise, \hmcstate) = \targetf(\hmcstate) + \frac{1}{2} \vecnorm{\hmcnoise}{2}^2.
\end{align}

The ideal HMC algorithm for sampling is based on the continuous
Hamiltonian dynamics; as such, it is not implementable in practice,
but instead a useful algorithm for understanding.  For a given time $T
> 0$ and vectors $u, v \in \real^\dims$, let $\hmcstate_T(u, v)$
denote the $\hmcstate$-solution to Hamilton's equations at time $T$
and with initial conditions $(\hmcnoise(0), \hmcstate(0)) = (u, v)$.
At iteration $k$, given the current iterate $X_k$, the ideal HMC
algorithm generates the next iterate $X_{k+1}$ via the update rule
$X_{k+1} = \hmcstate_T(\hmcnoise_k, X_k)$ where $p_k \sim N(0,
\Ind_\dims)$ is a standard normal random vector, independent of $X_k$
and all past iterates.  It can be shown that with an appropriately
chosen $T$, the ideal HMC algorithm converges to the stationary
distribution $\targetdensity$ without a Metropolis-Hastings adjustment
(see~\cite{neal2011mcmc,mangoubi2018dimensionally} for the existence
of such a solution and its convergence).

However, in practice, it is impossible to compute an exact solution to
Hamilton's equations.  Rather, one must approximate the solution
$\hmcstate_T(\hmcnoise_k, X_k)$ via some discrete process.  There are
many ways to discretize Hamilton's equations other than the simple
Euler discretization; see \cite{neal2011mcmc} for a discussion.  In
particular, using the leapfrog or St\"{o}rmer-Verlet method for
integrating Hamilton's equations leads to the Hamiltonian Monte Carlo
(HMC) algorithm.  It simulates the Hamiltonian dynamics for
$\internsteps$ steps via the leapfrog integrator. At each iteration,
given a integer $\internsteps \geq 1$, a previous state $\hmcstate_0$
and fresh $\hmcnoise_0 \sim \Normal(0, \Ind_\dims)$, it runs the
following updates for $k = 0, 1, \ldots, \internsteps-1$:
\begin{subequations}
\begin{align}
  \hmcnoise_{k+\frac{1}{2}} &= \hmcnoise_k -
  \frac{\step}{2}\gradf(\hmcstate_k) \\
  \hmcstate_{k+1} &=
  \hmcstate_k + \step \hmcnoise_{k+\frac{1}{2}}
  \\
  \hmcnoise_{k+1} &= \hmcnoise_{k+\frac{1}{2}} -
  \frac{\step}{2}\gradf(\hmcstate_{k+1}).
  \label{eq:hmc_leapfrog_integrator}
\end{align}
\end{subequations}
Since discretizing the dynamics generates discretization error at each
iteration, it is followed by a Metropolis-Hastings adjustment where the
proposal $(\hmcnoise_\internsteps, \hmcstate_\internsteps)$ is accepted with
probability
\begin{align}
  \label{eq:hmc_metropolis_step}
  \min\braces{1, \frac{\exp\parenth{-\hamiltonian(\hmcnoise_\internsteps, \hmcstate_\internsteps)}}{\exp\parenth{-\hamiltonian(\hmcnoise_0, \hmcstate_0)}}}.
\end{align}
See Algorithm~\ref{algo:hmc} for a detailed description of the HMC algorithm
with leapfrog integrator.

\begin{algorithm}[ht]
  \footnotesize
  \KwIn{Step size $\step$, number of internal leapfrog updates
    $\internsteps$, \\ and a sample $x_0$ from a starting distribution
    $\initial$} \KwOut{Sequence $x_1, x_2,\ldots$} \For{$i=0, 1,
    \ldots $}
  {%
  \textbf{Proposal step}:\\
   $\hmcstate_0 \gets x_i$\;
   \emph{Draw} $\hmcnoise_0 \sim \Normal(0, \Ind_\dims)$\;
   \For{$k=1, \ldots, \internsteps$}
    {%
      \quad $(\hmcnoise_{k}, \hmcstate_{k}) \gets \text{Leapfrog}(\hmcnoise_{k-1}, \hmcstate_{k-1}, \step)$\;
    } {\footnotesize{\% $\hmcstate_K$ is now the new proposed state}}\;
    \textbf{Accept-reject step}:\\
      \quad compute $\alpha_{i+1} \gets \displaystyle\min\braces{1, \frac{\exp
      \parenth{-\hamiltonian(\hmcnoise_{K}, \hmcstate_{K})}}{\exp\parenth{-\hamiltonian
      (\hmcnoise_0, \hmcstate_0)}}}$\\
        \quad With probability $\alpha_{i+1}$ \emph{accept} the proposal:
        $x_{i+1} \gets \hmcstate_K $\\
        \quad With probability $1-\alpha_{i+1}$ \emph{reject} the proposal:
        $x_{i+1} \gets x_i$

  }
  \SetKwFunction{FMain}{Leapfrog}
  \SetKwProg{Pn}{Program}{:}{\KwRet{$(\tilde{\hmcnoise}, \tilde{\hmcstate})$}}
  \Pn{\FMain{$\hmcnoise$, $\hmcstate$, $\step$}}{
        $\tilde{\hmcnoise} \gets \hmcnoise - \frac{\step}{2}\gradf(\hmcstate)$\;
        $\tilde{\hmcstate} \gets \hmcstate + \step \tilde{\hmcnoise}$\;
        $\tilde{\hmcnoise} \gets \tilde{\hmcnoise} - \frac{\step}{2}\gradf(\tilde{\hmcstate})$\;
  }
  \caption{Metropolized HMC with leapfrog integrator}
  \label{algo:hmc}
\end{algorithm}

\paragraph{Remark:}
The HMC with leapfrog integrator can also be seen as a multi-step
version of a simpler Langevin algorithm. Indeed, running the HMC
algorithm with $\internsteps = 1$ is equivalent to the MALA algorithm
after a re-parametrization of the step-size $\step$.  In practice, one
also uses the HMC algorithm with a modified Hamiltonian, in which the
quadratic term $\vecnorm{\hmcnoise}{2}^2$ is replaced by a more
general quadratic form $\hmcnoise^T M \hmcnoise$.  Here $M$ is a
symmetric positive definite matrix to be chosen by the user; see
Appendix~\ref{sub:the_effect_of_linear_transformations} for further
discussion of this choice.  In the main text, we restrict our analysis
to the case $M = I$.


\section{Main results}
\label{sec:main_results}

We now turn to the statement of our main results.  We remind the
readers that HMC refers to Metropolized HMC with leapfrog integrator,
unless otherwise specified.  We first collect the set of assumptions
for the target distributions in Section~\ref{sub:problem_set_up}.
Following that in Section~\ref{sub:mixing_time_bounds_for_hmc_for_general_regular_target_distributions},
we state our results for HMC: first, we derive the mixing time bounds for
general target distributions in Theorem~\ref{thm:hmc_mixing_general}
and then apply that result to obtain concrete guarantees for HMC with
strongly log-concave target distributions. We defer the discussion of
weakly log-concave target distributions and (non-log-concave) perturbations
of log-concave distributions to Appendix~\ref{sec:beyond_strongly_log_concave}.
In Section~\ref{sub:mixing_time_of_a_markov_chain_with_conductance_profile},
we discuss the underlying results that are used to derive sharper
mixing time bounds using conductance profile (see Lemmas~\ref{prop:mixing_bound_using_conductanceprofile}
and~\ref{prop:conductanceprofile_via_overlaps}).  Besides being
central to the proof of Theorem~\ref{thm:hmc_mixing_general},
these lemmas also enable a sharpening of the mixing time guarantees for
MALA and MRW (without much work), which we state in
Section~\ref{sub:improved_warmness_dependency_for_mala_and_mrw}.


\subsection{Assumptions on the target distribution}
\label{sub:problem_set_up}

In this section, we introduce some regularity notions and state the
assumptions on the target distribution that our results in the next
section rely on.

\paragraph{Regularity conditions:}
\label{par:regularity_conditions_}
\begin{subequations}
A function $\targetf$ is called:
  \begin{align}
    \label{eq:assumption_smoothness}
    \smoothness\text{-smooth}: \quad\ \ \targetf(y) - \targetf(x) -
    \gradf(x)\tp(y-x) & \leq \frac{\smoothness}{2} \vecnorm{x-y}{2}^2,
    \\
          \label{eq:assumption_scparam}
    \scparam\text{-strongly convex}: \quad\ \
    \targetf(y) - \targetf(x) - \gradf(x)\tp(y-x) & \geq
    \frac{\scparam}{2} \vecnorm{x-y}{2}^2, \\
      \label{eq:assumption_hessianLip}
    \hessianLip\text{-Hessian Lipschitz}: \quad\quad\quad\quad\quad
    \matsnorm{\hessf(x) - \hessf(y)}{\text{op}} &\leq \hessianLip
    \vecnorm{x - y}{2},
  \end{align}
  where in all cases, the inequalities hold for all $x, y \in
  \real^\dims$.

A distribution $\Pi$ with support $\mathcal{X} \subset\real^\dims$ is
said to satisfy the \emph{isoperimetric inequality} ($\logisopower =
0$) or the \emph{log-isoperimetric inequality} ($\logisopower =
\frac{1}{2}$) with constant $\isoconst_\logisopower$ if given any
partition $S_1, S_2, S_3$ of $\mathcal{X}$, we have
  \begin{align}
    \label{eq:assumption_isoperimetric}
    \Pi(S_3) \geq \frac{1}{2\isoconst_\logisopower} \cdot d(S_1, S_2)
    \cdot \min\braces{\Pi(S_1), \Pi(S_2)} \cdot
    \log^{\logisopower}\parenth{1 +
      \frac{1}{\min\braces{\Pi(S_1),\Pi(S_2}}}.
  \end{align}
  where the distance between two sets $S_1, S_2$ is defined as $
  d(S_1, S_2) = \inf_{x\in S_1, y\in S_2}\braces{\vecnorm{x - y}{2}}$.
  For a distribution $\Pi$ with density $\pi$ and a given set
  $\convset$, its restriction to $\convset$ is the distribution
  $\Pi_\convset$ with the density $\pi_\convset(x) =
  \frac{\pi(x)\indicator_\convset (x)}{\Pi(\convset)}$.

\paragraph{Assumptions on the target distribution:}
\label{par:assumptions_on_the_target_distribution_}

We introduce two sets of assumptions for the target distribution:
\begin{enumerate}[label=(\Alph*)]
  \item \label{itm:assumptionA} We say that the target distribution
    $\target$ is $(\smoothness, \hessianLip, \res,
    \isoconst_\logisopower, \gradbound)$-\textit{regular} if the
    negative log density $\targetf$ is
    $\smoothness$-smooth~\eqref{eq:assumption_smoothness} and has
    $\hessianLip$-Lipschitz Hessian~\eqref{eq:assumption_hessianLip},
    and there exists a convex measurable set $\convset$ such that the
    distribution $\target_{\convset}$ is
    $\isoconst_\logisopower$-isoperimetric~\eqref{eq:assumption_isoperimetric},
    and the following conditions hold:
    \begin{align}
          \label{eq:assumption_high_mass_gradbound}
    \target(\convset) \geq 1-\res \quad \mbox{and} \quad
    \vecnorm{\gradf(x)}{2} &\leq \gradbound, \quad \text{for all } x
    \in \convset.
  \end{align}
  \item \label{itm:assumptionB} We say that the target distribution
    $\target$ is $(\smoothness,\hessianLip,\scparam)$-\emph{strongly
    log-concave} if the negative log density is
    $\smoothness$-smooth~\eqref{eq:assumption_smoothness},
    $\scparam$-strongly convex~\eqref{eq:assumption_scparam}, and
    $\hessianLip$-Hessian-Lipschitz~\eqref{eq:assumption_hessianLip}.
    Moreover, we use $\xstar$ to denote the unique mode of $\target$
    whenever $\targetf$ is strongly convex.
\end{enumerate}

Assumption~\ref{itm:assumptionB} has appeared in several past papers
on Langevin
algorithms~\citep{dalalyan2016theoretical,dwivedi2018log,cheng2017convergence}
and the Lipschitz-Hessian condition~\eqref{eq:assumption_hessianLip}
has been used in analyzing Langevin algorithms with inaccurate
gradients~\citep{dalalyan2019user} as well as the unadjusted HMC
algorithm~\citep{mangoubi2018dimensionally}.  It is worth noting
Assumption~\ref{itm:assumptionA} is strictly weaker than
Assumption~\ref{itm:assumptionB}, since it allows for distributions
that are not log-concave. In Appendix~\ref{sec:proof_of_corollary_cor:hmc_mixing_sc_fixedk}
(see Lemma~\ref{lem:strongly_logconcave_regular_to_general}),
we show how Assumption~\ref{itm:assumptionB} implies a version of
Assumption~\ref{itm:assumptionA}.

\end{subequations}


\subsection{Mixing time bounds for HMC}
\label{sub:mixing_time_bounds_for_hmc_for_general_regular_target_distributions}

We start with the mixing time bound for HMC applied to any
distribution $\target$ satisfying Assumption~\ref{itm:assumptionA}.
Let \hmc~denote the $\frac{1}{2}$-lazy Metropolized HMC algorithm with
$\step$~step size and $\internsteps$~leapfrog steps in each
iteration. Let $\Tmix_2^\taghmc(\errorparam; \initial)$
denote the $\Ell_2$-mixing time~\eqref{eq:tmix_defn} for this chain with the
starting distribution $\initial$.
We use $c$ to denote a universal constant.
\begin{theorem}
  \label{thm:hmc_mixing_general}
   Consider an $(\smoothness, \hessianLip, \res,
   \isoconst_\logisopower, \gradbound)$-regular target distribution
   (cf. Assumption~\ref{itm:assumptionA}) and a $\warmparam$-warm
   initial distribution $\initial$.  Then for any fixed target
   accuracy $\errorparam \in (0, 1)$ such that $\errorparam^2\geq
   3\warmparam\res$, there exist choices of the parameters
   $(\internsteps, \step)$ such that \hmc chain with $\initial$ start
   satisfies
  \begin{align*}
    \Tmix_2^\taghmc(\errorparam; \initial) \leq
    \begin{cases}
       c\cdot \max\braces{\log\warmparam,
         \displaystyle\frac{\isoconst_\logisopower^2}
                           {\internsteps^2\step^2}
                           \log\parenth{\frac{\log\warmparam}{\errorparam}}}
       &\text{if $\logisopower = \frac{1}{2} \qquad$
         [log-isoperimetric~\eqref{eq:assumption_isoperimetric}]}
       \\[2mm] c\cdot
       \displaystyle\frac{\isoconst_\logisopower^2}{\internsteps^2\step^2
       } \log\parenth{\frac{\warmparam}{\errorparam}} &\text{if
         $\logisopower = 0 \qquad$
         [isoperimetric~\eqref{eq:assumption_isoperimetric}]}.
    \end{cases}
  \end{align*}
\end{theorem}
\noindent See
Section~\ref{ssub:proof_of_theorem_thm:hmc_mixing_general} for the
proof, where we also provide explicit conditions on $\step$ and
$\internsteps$ in terms of the other parameters (cf.
equation~\eqref{eq:step_condition_thm_proof}).

Theorem~\ref{thm:hmc_mixing_general} covers mixing time bounds for
distributions that satisfy isoperimetric or log-isoperimetric
inequality provided that: (a) both the gradient and Hessian of the
negative log-density are Lipschitz; and (b) there is a convex set that
contains a large mass $(1-\res)$ of the distribution. The mixing time
only depends on two quantities: the log-isoperimetric (or
isoperimetric) constant of the target distribution and the effective
step-size $\internsteps^2 \step^2$.  As shown in the sequel, these
conditions hold for log-concave distributions as well as certain
perturbations of them.  If the distribution satisfies a
log-isoperimetric inequality, then the mixing time dependency on the
initialization warmness parameter $\warmparam$ is relatively weak
$O(\log\log\warmparam)$.  On the other hand, when only an
isoperimetric inequality (but not log-isoperimetric) is available, the
dependency is relatively larger $O(\log \warmparam)$.  In our current
analysis, we can establish the $\errorparam$-mixing time bounds up-to
an error $\errorparam$ such that $\errorparam^2 \geq 3\warmparam\res$.
If mixing time bounds up to an arbitrary accuracy are desired, then
the distribution needs to
satisfy~\eqref{eq:assumption_high_mass_gradbound} for arbitrary small
$\res$. For example, as we later show in
Lemma~\ref{lem:strongly_logconcave_regular_to_general}, arbitary small
$\res$ can be imposed for strongly log-concave densities (i.e., satisfying
Assumption~\ref{itm:assumptionB}).

Let us now derive several corollaries of
Theorem~\ref{thm:hmc_mixing_general}. We begin with non-asymptotic
mixing time bounds for \hmc~chain for strongly-log concave target
distributions. Then we briefly discuss the corollaries for weakly
log-concave target and non-log-concave target distributions and defer
the precise statements to
Appendix~\ref{sec:beyond_strongly_log_concave}. These results also
provide a basis for comparison of our results with prior work.


\subsubsection{Strongly log-concave target}
\label{ssub:hmc_mixing_time_bounds_for_strongly_log_concave_target}

We now state an explicit mixing time bound of HMC for a strongly
log-concave distribution. We consider an $(\smoothness, \hessianLip,
\scparam)$-strongly log-concave distribution
(assumption~\ref{itm:assumptionB}). We use \mbox{$\condition
  =\smoothness/\scparam$} to denote the condition number of the
distribution.  Our result makes use of the following function
\begin{subequations}
\begin{align}
\label{eq:def_radius}
  \radius(\res) & \defn 1 + \max \braces { \left(\frac{\log(1/\res)}{\dims}
    \right)^{1/4}, \; \parenth{\frac{\log(1/\res)}{\dims}}^{1/2} },
\end{align}
for $\res >0$, and involves the step-size choices
\begin{align}
  \label{eq:step_hmc_waram_vs_feasible}
  \stepwarm =
  \sqrt{\frac{1}{c\smoothness\cdot\radius\parenth{\frac{\errorparam^2}{3\warmparam}}
      \dims^{\frac{7}{6}}}} \;, \quad \mbox{and} \quad \stepfeasible =
  \sqrt{\frac{1}{c\smoothness\cdot\radius\parenth{\frac{\errorparam^2}{2\condition^{\dims}}}}
    \min \braces{\frac{1}{\dims\condition^{\frac{1}{2}}},
      \frac{1}{\dims^ {\frac{2}{3}}\condition^{\frac{5}{6}}},
      \frac{1}{\dims^{\frac{1}{2}} \condition^ {\frac{3}{2}}}}}.
\end{align}
\end{subequations}
With these definitions, we have the following:
\begin{corollary}
  \label{cor:hmc_mixing_sc_fixedK}
  Consider an $(\smoothness, \hessianLip, \scparam)$-strongly
  log-concave target distribution $\target$
  (cf. Assumption~\ref{itm:assumptionB}) such that
  \mbox{$\hessianLip^{2/3}=O(\smoothness)$}, and any error tolerance
  $\errorparam \in (0, 1)$.
  \begin{subequations}
  \begin{enumerate}[label=(\alph*)]
  	\item Suppose that $\condition = O(\dims^{\frac{2}{3}})$ and
          $\warmparam =
          O\parenth{\exp\parenth{\dims^{\frac{2}{3}}}}$. Then with any
          $\warmparam$-warm initial distribution $\initial$,
          hyper-parameters $\internsteps=\dims^{\frac{1}{4}}$ and
          $\step = \stepwarm$, the \hmc~chain satisfies
  	\begin{align}
  	\label{eq:hmc_warm_mix}
    \Tmix_2^\taghmc(\errorparam; \initial) \leq c \;
    \dims^{\frac{2}{3}} \; \condition \, \radius
    \parenth{\frac{\errorparam^2} {3\warmparam}} \; \log \parenth{
    \frac{\log \warmparam}{\errorparam} }.
     \end{align}
     \item With the initial distribution $\initialstar =
       \Normal(\xstar, \frac{1}{\smoothness}\Ind_\dims)$,
       hyper-parameters $\internsteps= \condition^{\frac{3}{4}}$ and
       $\step = \stepfeasible$, the \hmc~chain satisfies
      \begin{align}
      \label{eq:hmc_feasible_mix}
    \Tmix_2^\taghmc(\errorparam; \initialstar) \leq c \; \radius
    \parenth{\frac{\errorparam^2}{3\condition^\dims}} \, \max
    \braces{\dims\log{\condition}, \max\brackets{\dims,
        \dims^{\frac{2}{3}}\condition^{\frac{1}{3}},
        \dims^{\frac{1}{2}}\condition }
      \log\parenth{\frac{\dims\log{\condition}}{\errorparam}}}.
  \end{align}
  \end{enumerate}
  \end{subequations}
\end{corollary}
\noindent See
Appendix~\ref{sec:proof_of_corollary_cor:hmc_mixing_sc_fixedk} for the
proof.  In the same appendix, we also provide a more refined mixing
time of the HMC chain for a more general choice of hyper-parameters
(see Corollary~\ref{cor:hmc_mixing_sc_general}).  In fact, as shown in
the proof, the assumption $\hessianLip^{2/3} = O(\smoothness)$ is not
necessary in order to control mixing; rather, we adopted it above to
simplify the statement of our bounds. A more detailed discussion on
the particular choice for step size $\step$ is provided in
Appendix~\ref{sec:discussion_around_corollary_cor:hmc_mixing_sc_fixedK}.\\

\paragraph{MALA vs HMC---Warm start:} 
Corollary~\ref{cor:hmc_mixing_sc_fixedK} provides mixing time bounds
for two cases. The first result~\eqref{eq:hmc_warm_mix} implies that
given a warm start for a well-conditioned
strongly log concave distribution,  i.e., with constant $\warmparam$
and $\condition \ll \dims$, the
$\errorparam$-$\Ell_2$-mixing time\footnote{Note that
  $\radius(\errorparam^2) \leq 6$ for $\errorparam \geq \frac{2}
  {e^{\dims/2}}$ and thus we can treat $\radius$ as a small constant
  for a large range of $\errorparam$. Otherwise, if $\errorparam$
  needs to be extremely small, the results still hold with an extra
  $\log^{\frac{1}{2}}\parenth{\frac{1}{\errorparam}}$ dependency.} of
HMC scales \mbox{$\Ot(d^{\frac{2}{3}} \log (1/\errorparam))$.}  It is
interesting to compare this guarantee with known bounds for the MALA
algorithm. However since each iteration of MALA uses only a single gradient
evaluation, a fair comparison would require us to track
the total number of gradient evaluations required by the \hmc~chain to
mix. For HMC to achieve accuracy $\errorparam$, the total
number of gradient evaluations is given by $\internsteps \cdot
\Tmix_2^\taghmc (\errorparam; \initial)$, which in the above setting,
scales as $\Ot(d^{\frac{11}{12}}\condition \log (1/\errorparam))$.
This rate was also summarized in Table~\ref{tab:mixing_times_all}.
On the other hand, Theorem 1 in~\cite{dwivedi2018log} shows that the corresponding
number of gradient evaluations for MALA  is $\Ot(\dims\condition\log(1/\errorparam))$.
As a result, we conclude that the upper bound for HMC is $\dims^
{\frac{1}{12}}$ better than the known upper bound for MALA with a warm start
for a well-conditioned strongly log concave target distribution.
We summarize these rates in Table~\ref{tab:warm_mix}. Note that MRW is a zeroth order algorithm, which makes use of function evaluations but not gradient information.

\begin{table}[ht]
    \centering
    {\renewcommand{\arraystretch}{.8}
    \begin{tabular}{ccc}
        \toprule
         \thead{ \bf Sampling algorithm} & \thead{\bf Mixing time} & \thead{
         \bf \#Gradient evaluations}
        \\ \midrule
        \thead{ MRW~\cite[Theorem 2]{dwivedi2018log}}
        & $\dims\condition^2 \cdot\log \frac{1}{\errorparam}$
        & \small NA
        \\[4mm]
        \thead{ MALA~\cite[Theorem 1]{dwivedi2018log}}
        & $\dims \condition\cdot\log \frac{1}{\errorparam}$
        & $\dims \condition\cdot\log \frac{1}{\errorparam}$
        \\[4mm]
        \thead{\hmc~[ours, Corollary~\ref{cor:hmc_mixing_sc_fixedK}]}
        & $\dims^{\frac{2}{3}} \condition\cdot\log \frac{1}{\errorparam}$
        & $\dims^{\frac{11}{12}} \condition\cdot\log \frac{1}{\errorparam}$
        \\[2mm]
        \bottomrule
    \end{tabular}
    }
    \caption{Summary of the $\errorparam$-mixing time and the
      corresponding number of gradient evaluations for MRW, MALA and
      HMC from a \emph{warm start} with an $(\smoothness, \hessianLip,
      \scparam)$-strongly-log-concave target. These statements hold
      under the assumption $\hessianLip^{2/3} = O(\smoothness)$,
      $\condition=\frac{\smoothness}{\scparam}\ll\dims$, and omit
      logarithmic terms in dimension.}
    \label{tab:warm_mix}
\end{table}


\paragraph{MALA vs HMC---Feasible start:}
\label{par:mala_vs_hmc_feasible_start_}

In the second result~\eqref{eq:hmc_feasible_mix}, we cover the case
when a warm start is not available.  In particular, we analyze the HMC
chain with the feasible initial distribution $\initialstar =
\Normal(\xstar, \frac{1}{\smoothness}\Ind_\dims)$. Here $\xstar$
denotes the unique mode of the target distribution and can be easily
computed using an optimization scheme like gradient descent.  It is
not hard to show (see Corollary~1 in~\cite{dwivedi2018log}) that for an
$\smoothness$-smooth~\eqref{eq:assumption_smoothness} and $\scparam$
strongly log-concave target
distribution~\eqref{eq:assumption_scparam}, the distribution
$\initialstar$ acts as a $\condition^{\dims/2}$-warm start
distribution.  Once again, it is of interest to determine whether HMC
takes fewer gradient steps when compared to MALA to obtain an
$\errorparam$-accurate sample.  We summarize the results in
Table~\ref{tab:feasible_mix}, with log factors hidden, and note
that HMC with $\internsteps=\condition^{3/4}$ is faster than MALA for
as long as $\condition$ is not too large. From the last column, we
find that when $\condition \ll \dims^{\frac{1}{2}}$, HMC is faster
than MALA by a factor of $\condition^{\frac{1}{4}}$ in terms of number
of gradient evaluations. It is worth noting that for the
feasible start $\initialstar$, the mixing time bounds for MALA and
MRW in Dwivedi et al.~\citet{dwivedi2018log} were loose by a factor
$d$ when compared to the tighter bounds in Theorem~\ref{thm:mala_mrw_mixing}
derived later in this paper.

\begin{table}[ht]
    \centering
    {\renewcommand{\arraystretch}{.8}
    \begin{tabular}{cccc}
        \toprule
         \thead{ \bf Sampling algorithm} & \thead{\bf Mixing time} &
         \multicolumn{2}{c}{\thead{\bf \# Gradient Evaluations}}\\
         &&{\small{general}} $\condition$ &$\condition \ll \dims^{\frac{1}
         {2}}$
        \\ \midrule
        \thead{ MRW~[ours, Theorem~\ref{thm:mala_mrw_mixing}]}
        & $\dims\condition^2 $
        & \small NA & \small NA
        \\[4mm]
        \thead{ MALA~[ours, Theorem~\ref{thm:mala_mrw_mixing}]}
        & $\max\braces{\dims \condition, \dims^{\frac{1}{2}} \condition^{\frac{3}{2}}}$
        & $\max\braces{\dims \condition, \dims^{\frac{1}{2}} \condition^{\frac{3}{2}}}$
        & $\dims\condition$
        \\[4mm]
        \thead{\hmc~[ours, Corollary~\ref{cor:hmc_mixing_sc_fixedK}]}
        &$\max\big\{\dims, \dims^{\frac{2}{3}}\condition^
        {\frac{1}{3}}, \dims^{\frac{1}{2}}\condition\big\}
        $
        & $\max\big\{\dims \condition^{\frac{3}{4}}, \dims^{\frac{2}{3}}\condition^
        {\frac{13}{12}}, \dims^{\frac{1}{2}}\condition^{\frac{7}{4}}\big\}
        $
        & $\dims\condition^{\frac{3}{4}}$
        \\[2mm]
        \bottomrule
    \end{tabular}
    }
    \caption{ Summary of the $\errorparam$-mixing time and the
      corresponding number of gradient evaluations for MRW, MALA and
      HMC from the \emph{feasible start} $\initialstar =
      \Normal(\xstar, \frac{1}{\smoothness}\Ind_\dims)$ for an
      $(\smoothness, \hessianLip, \scparam)$-strongly-log-concave
      target. Here $\xstar$ denotes the unique mode of the target
      distribution. These statements hold uner the assumption
      $\hessianLip = O(\smoothness^{\frac{3}{2}})$, and hide the
      logarithmic factors in $\errorparam,\dims$ and
      $\condition=\smoothness/\scparam$.}
    \label{tab:feasible_mix}
\end{table}

\paragraph{Metropolized HMC vs Unadjusted HMC:} 
There are many recent results on the $1$-Wasserstein distance mixing
of unadjusted versions of HMC (for instance, see the papers~\cite{mangoubi2018dimensionally,lee2018algorithmic}).  For
completeness, we compare our results with them in the
Appendix~\ref{sub:comparison_with_guarantees_for_other_versions_of_hmc};
in particular, see Table~\ref{tab:mixing_times_with_poly} for a
comparative summary.  We remark that comparisons of these different
results is tricky for two reasons: (a) The $1$-Wasserstein distance
and the total variation distance are not strictly comparable, and, (b)
the unadjusted HMC results always have a polynomial dependence on the
error parameter $\errorparam$ while our results for Metropolized HMC
have a superior logarithmic dependence on~$\errorparam$.  Nonetheless,
the second difference between these chains has a deeper consequence,
upon which we elaborate further in
Appendix~\ref{sub:comparison_with_guarantees_for_other_versions_of_hmc}.
On one hand, the unadjusted chains have better mixing time in terms of
scaling with $\dims$, if we fix $\errorparam$ or view it as
independent of $\dims$. On the other hand, when such chains are used
to estimate certain higher-order moments, the polynomial dependence on
$\errorparam$ might become the bottleneck and Metropolis-adjusted
chains would become the method of choice.

\paragraph{Ill-conditioned target distributions:}
\label{par:ill_condi}
In order to keep the statement of Corollary~\ref{cor:hmc_mixing_sc_fixedK}
simple, we stated the mixing time bounds of \hmc-chain only for a particular
choice of $(\internsteps, \step)$. In our analysis, this choice ensures
that HMC is better than
MALA only when condition number $\condition$ is small. For Ill-conditioned
distributions, i.e., when $\condition$ is large, finer tuning of \hmc-chain
is required. In Appendices~\ref{sec:proof_of_corollary_cor:hmc_mixing_sc_fixedk}
and \ref{sec:discussion_around_corollary_cor:hmc_mixing_sc_fixedK} (see
Table~\ref{tab:hmc_cor1_params} for the hyper-parameter choices),
we show that HMC is strictly better than MALA as long as $\condition\leq
\dims$ and as good as MALA when $\condition \geq \dims$.

\paragraph{Beyond strongly-log-concave:}
\label{par:remark_}

The proof of Corollary~\ref{cor:hmc_mixing_sc_fixedK} is based on the
fact that $(\smoothness, \hessianLip, \scparam)$-strongly-log-concave
distribution is in fact an $(\smoothness, \hessianLip, \res,
\isoconst_ {1/2}, \gradbound_\res)$ -regular distribution for any
$\res \in (0, 1)$. Here $\isoconst_{1/2}= {1}/{\sqrt{\scparam}}$ is
fixed and the bound on the gradient $\gradbound_\res= \radius(\res)
\sqrt{{\dims}/{\scparam}}$ depends on the choice of $\res$.  The
result is formally stated in
Lemma~\ref{lem:strongly_logconcave_regular_to_general} in
Appendix~\ref{sec:proof_of_corollary_cor:hmc_mixing_sc_fixedk}.
Moreover, in Appendix~\ref{sec:beyond_strongly_log_concave}, we
discuss the case when the target distribution is weakly log concave
(under a bounded fourth moment or bounded covariance matrix
assumption) or a perturbation of log-concave distribution. See
Corollary~\ref{cor:hmc_mixing_wc} for specific details where we
provide explicit expressions for the rates that appear in third and
fourth columns of Table~\ref{tab:mixing_times_all}.


\subsection{Mixing time bounds via conductance profile}
\label{sub:mixing_time_of_a_markov_chain_with_conductance_profile}

In this section, we discuss the general results that form the basis of
the analysis in this paper.  A standard approach to controlling mixing
times is via worst-case conductance bounds.  This method was
introduced by \cite{jerrum1988conductance} for
discrete space chains and then extended to the continuous space
settings by \cite{lovasz1993random}, and has
been thoroughly studied.  See the survey~\citep{vempala2005geometric}
and the references therein for a detailed discussion of conductance
based methods for continuous space Markov chains.

Somewhat more recent work on discrete state chains has introduced more
refined methods, including those based on the conductance
profile~\citep{lovasz1999faster}, the spectral and conductance
profile~\citep{goel2006mixing}, as well as the evolving set
method~\citep{morris2005evolving}.  Here we extend one of the
conductance profile techniques from the paper by \cite{goel2006mixing}
from discrete state to continuous state chains, albeit with several
appropriate modifications suited for the general setting.

We first introduce some background on the conductance profile.  Given
a Markov chain with transition probability $\transprob: \statespace
\times \borel \parenth{\statespace} \rightarrow \real$, its stationary
\emph{flow} $\flow: \borel(\statespace) \rightarrow \real$ is defined
as
\begin{align}
  \label{eq:def_flow}
  \flow(S) &= \int_{x \in S} \transprob(x, S^c) \targetdensity(x) dx
  \quad \text{for any } S \in \borel(\statespace).
\end{align}
Given a set $\convset \subset \statespace$, the
\emph{$\convset$-restricted conductance profile} is given by
\begin{align}
  \label{eq:def_conductanceprofile}
  \conductanceprofile(v) = \inf_{\target(S \cap \convset)\in(0, v]}
    \frac{\flow(S)}{\target(S \cap \convset)} \qquad \text{for any } v
    \in \big(0, \; \target(\convset)/2 \big].
\end{align}
(The classical conductance constant $\Phi$ is a special case; it can
be expressed as $\Phi = \Phi_{\statespace} (\frac{1}{2})$.)  Moreover,
we define the \emph{truncated extension} $\truncconductanceprofile$ of
the function $\conductanceprofile$ to the positive real line as
\begin{align}
\label{eq:def_truncatedprofile}
  \truncconductanceprofile(v) = \begin{cases} \conductanceprofile(v),
    & v \in \left(0, \frac{\target(\convset)}{2}\right ]
      \\ \conductanceprofile(\target(\convset)/2), & v \in
      \left[\frac{\target(\convset)} {2}, \infty \right).
  \end{cases}
\end{align}
In our proofs we use the conductance profile with a suitably chosen
set $\convset$.


\paragraph{Smooth chain assumption:}
\label{par:smooth_chain_}
We say that the Markov chain satisfies the \emph{smooth chain
  assumption} if its transition probability function
$\transprob: \statespace \times \borel (\statespace) \rightarrow
\real_+$ can be expressed in the form
\begin{align}
  \label{eq:smooth_chain_assumption}
  \transprob(x, dy) = \transkernel(x, y) dy + \alpha_x \dirac_x (dy) \quad
  \text{for all }x,  y \in \statespace,
\end{align}
where $\transkernel$ is the transition kernel satisfying
$\transkernel(x, y) \geq 0 $ for all $x, y \in \statespace$.  Here
$\dirac_x$ denotes the Dirac-delta function at $x$ and consequently,
$\alpha_x$ denotes the one-step probability of the chain to stay at
its current state $x$.  Note that the three algorithms discussed in
this paper (MRW, MALA and HMC) all satisfy the smooth chain
assumption~\eqref{eq:smooth_chain_assumption}.  Throughout the paper,
when dealing with a general Markov chain, we assume that it satisfies
the smooth chain assumption.


\paragraph{Mixing time via conductance profile:}
\label{par:mixing_time_via_conductance_profile_}

We now state our
Lemma~\ref{prop:mixing_bound_using_conductanceprofile} that provides a
control on the mixing time of a Markov chain with continuous-state
space in terms of its restricted conductance profile. We show that
this control (based on conductance profile) allows us to have a better
initialization dependency than the usual conductance based control
(see~\cite{lovasz1990ballwalk,lovasz1993random,dwivedi2018log}).  This
method for sharpening the dependence is known for discrete-state
Markov chains; to the best of our knowledge, the following lemma is
the first statement and proof of an analogous sharpening for
continuous state space chains:
\begin{lemma}
  \label{prop:mixing_bound_using_conductanceprofile}
  Consider a reversible, irreducible, $\lazyparam$-lazy and smooth
  Markov chain~\eqref{eq:smooth_chain_assumption} with stationary
  distribution $\target$. Then for any error tolerance $\errorparam$,
  and a $\warmparam$-warm distribution $\initial$, given a set
  $\convset$ such that $\target(\convset) \geq 1-\frac{\errorparam^2}
  {3\warmparam^2}$, the $\errorparam$-$\Ell_2$ mixing time of the
  chain is bounded as
  \begin{align}
    \label{eq:mixing_bound_using_conductanceprofile}
    \Tmix_2(\errorparam; \initial) \leq
    \int_{4/\warmparam}^{8/\errorparam^2} \frac{8 \: dv}{\lazyparam \cdot
      v{\truncconductanceprofile}^2(v)},
  \end{align}
where $\truncconductanceprofile$ denotes the truncated
$\convset$-restricted conductance
profile~\eqref{eq:def_truncatedprofile}.
\end{lemma}
\noindent See Appendix~\ref{sec:proof_of_lemma_2} for the proof, which is
based on an appropriate generalization of the ideas used by~\cite{goel2006mixing} for discrete state chains.\\

The standard conductance based analysis makes use of the worst-case conductance
bound for the chain. In contrast, Lemma~\ref{prop:mixing_bound_using_conductanceprofile}
relates the mixing time to the conductance profile, which can be seen as point-wise conductance.
We use the $\convset$-restricted conductance profile to state our bounds, because often a
Markov chain has poor conductance only in regions that have very small
probability under the target distribution. Such a behavior is not disastrous
as it does not really affect the mixing of the chain up to a suitable
tolerance.
Given the bound~\eqref{eq:mixing_bound_using_conductanceprofile},
we can derive mixing time bound for a Markov chain readily if we
have a bound on the $\convset$-restricted conductance profile $\conductanceprofile$
for a suitable $\convset$. More precisely, if the $\convset$-restricted conductance profile
$\conductanceprofile$ of the Markov chain is bounded as
\begin{align*}
  \conductanceprofile(v) \geq
  \sqrt{B \log \parenth{\frac{1}{v}}} \quad\text{for } v \in
    \brackets{\frac{4} {\warmparam}, \frac{1}{2}},
\end{align*}
for some $\warmparam > 0$ and $\convset$ such that $\target(\convset)
  \geq 1-\frac{\errorparam^2}{3\warmparam^2}$.
Then with a $\warmparam$-warm start, Lemma~\ref{prop:mixing_bound_using_conductanceprofile}
implies the following useful bound on the mixing time of the $\lazyparam$-lazy
Markov chain:
\begin{align}
\label{eq:simple_mixing_bound}
  \Tmix_2(\errorparam; \initial) \leq
    \frac{64}{\lazyparam B }
    \log \parenth{\frac{\log\warmparam}{2\errorparam}}.
\end{align}
We now relate our result to prior work based on conductance profile.

\paragraph{Prior work:} 
\label{par:prior_work_}
For discrete state chains, a result similar to
Lemma~\ref{prop:mixing_bound_using_conductanceprofile} was already
proposed by Lov{\'a}sz and Kannan (Theorem~2.3 in \cite{lovasz1999faster}). Later on, \cite{morris2005evolving} and \cite{goel2006mixing}
used the notion of evolving sets and spectral profile respectively to
sharpen the mixing time bounds based on average conductance for
discrete-state space chains. In the context of continuous state space
chains, Lov{\'a}sz and Kannan claimed in their original
paper~\citep{lovasz1999faster} that a similar result should hold for
general state space chain as well, although we were unable to find any
proof of such a general result in that or any subsequent work.
Nonetheless, in a later work an average conductance based bound was
used by Kannan~\etal~to derive faster mixing time guarantees for
uniform sampling on bounded convex sets for ball walk (see Section 4.3
in~\cite{kannan2006blocking}).  Their proof technique is not
easily extendable to more general distributions including the general
log-concave distributions in $\real^\dims$. Instead, our proof of
Lemma~\ref{prop:mixing_bound_using_conductanceprofile} for general
state space chains proceeds by an appropriate generalization of the
ideas based on the spectral profile by \cite{goel2006mixing} (for discrete state chains).


\paragraph{Lower bound on conductance profile:}
\label{par:lower_bound_on_conductance_profile_}

Given the bound~\eqref{eq:simple_mixing_bound}, it suffices to derive
a lower bound on the conductance profile $\conductanceprofile$ of the
Markov chain with a suitable choice of the set $\convset$.  We now
state a lower bound for the restricted-conductance profile of a
general state space Markov chain that comes in handy for this task.
We note that a closely related logarithmic-Cheeger inequality was used
for sampling from uniform distribution of a convex
body~\citep{kannan2006blocking} and for sampling from log-concave
distributions~\citep{lee2018stochastic} without explicit constants.
Since we would like to derive a non-asymptotic mixing rate, we
re-derive an explicit form of their result.

Let scalars $s\in(0, 1/2]$, $\omega \in (0, 1)$ and $\Delta > 0$ be
      given and let $\transition_x$ denote the one-step transition
      distribution of the Markov chain at point~$x$.  Suppose that
      that chain satisfies
\begin{align}
  \label{EqnOneStep}
        {\tvdist{\transition_x}{\transition_y} \leq 1 - \omega} \qquad
        \mbox{ whenever $x, y \in \convset$ and $\vecnorm{x-y}{2} \leq
          \Delta$.}
\end{align}

\begin{lemma}
  \label{prop:conductanceprofile_via_overlaps}
For a given target distribution $\target$, let $\convset$ be a convex
measurable set such that the distribution $\target_{\convset}$
satisfies the isoperimetry (or log-isoperimetry)
condition~\eqref{eq:assumption_isoperimetric} with $\logisopower = 0$
(or $\logisopower = \frac{1}{2}$ respectively).  Then for any Markov
chain satisfying the condition~\eqref{EqnOneStep}, we have
  \begin{align}
  \label{eq:conductanceprofile_via_overlaps}
    \conductanceprofile(v) \geq \frac{\tvoverlap}{4} \cdot
    \min\braces{1, \frac{ \Delta }{16 \isoconst_\logisopower} \cdot \log^{\logisopower}
      \parenth{1 + \frac{1}{v}}},
      \quad \text{for any } v \in \brackets{0, \frac{\target(\convset)}{2}}.
  \end{align}
\end{lemma}
\noindent See
Appendix~\ref{sec:proof_of_lemma_lem:conductanceprofile_via_overlaps}
for the proof; the extra logarithmic term comes from the logarithmic
isoperimetric inequality ($\logisopower = \frac{1}{2}$).


\paragraph{Faster mixing time bounds:}
\label{par:faster_mixing_time_bounds_}

For any target distribution satisfying a logarithmic isoperimetric
inequality (including the case of a strongly log-concave
distribution), Lemma~\ref{prop:conductanceprofile_via_overlaps} is a
strict improvement of the conductance bounds derived in previous
works~\citep{lovasz1999hit,dwivedi2018log}.  Given this result, suppose
that we can find a convex set $\convset$ such that $\target(\convset)
\approx 1$ and the conditions of
Lemma~\ref{prop:conductanceprofile_via_overlaps} are met, then with a
$\warmparam$-warm start $\initial$, a direct application of the
bound~\eqref{eq:simple_mixing_bound} along with
Lemma~\ref{prop:conductanceprofile_via_overlaps} implies the following
bound:
\begin{align}
\label{eq:clean_bound}
  \Tmix_2(\errorparam; \initial) \leq O\parenth{\frac{1}{\omega^2\Delta^2}
  \log\frac{\log\warmparam}{\errorparam}}.
\end{align}
Results known from previous work for continuous state Markov chains
scale like $\frac{\log( \warmparam/\errorparam)}{\omega^2\Delta^2}$;
for instance, see Lemma~6 in~\cite{chen2018fast}.  In
contrast, the bound~\eqref{eq:clean_bound} provides an additional
logarithmic factor improvement in the factor $\warmparam$. Such an
improvement also allows us to derive a sharper dependency on dimension
$\dims$ for the mixing time for sampling algorithms other than HMC as
we now illustrate in the next section.

%

\subsection{Improved warmness dependency for MALA and MRW}
\label{sub:improved_warmness_dependency_for_mala_and_mrw}
As discussed earlier, the bound~\eqref{eq:clean_bound} helps derive a
$\frac{\log\log\warmparam}{\log\warmparam}$ factor improvement in the
mixing time bound from a $\warmparam$-warm start in comparison to
earlier conductance based results. In many settings, a suitable choice
of initial distribution has a warmness parameter that scales
exponentially with dimension $\dims$, e.g., $\warmparam=O(e^\dims)$.
For such cases, this improvement implies a gain of $O(\frac{d}{\log
  d})$ in mixing time bounds.  As already noted the distribution
$\initialstar = \mathcal{N} (x^*, \frac{1}{\smoothness}\Ind_\dims)$ is
a feasible starting distribution,  whose warmness scales exponentially with dimension
$\dims$. See Section 3.2 of the paper~\citep{dwivedi2018log},
where the authors show that computing $x^*$ is not expensive and even approximate
estimates of $x^*$ and $\smoothness$ are sufficient to provide a feasible
starting distribution.  We now state sharper mixing time bounds for MALA
and MRW with $\initialstar$ as the starting distribution.  In the result,
we
use $c_1, c_2$ to denote positive universal constants.
\begin{theorem}
  \label{thm:mala_mrw_mixing}
 Assume that the target distribution $\target$ satisfies the
 conditions~\eqref{eq:assumption_smoothness}
 and~\eqref{eq:assumption_scparam} (i.e., the negative log-density
 is $\smoothness$-smooth and $\scparam$-strongly convex).  Then given
 the initial distribution $\initialstar = \Normal (x^*,
 \frac{1}{\smoothness}\Ind_\dims)$, the $\frac{1}{2}$-lazy versions of
 MRW and MALA (Algorithms~\ref{algo:mrw} and~\ref{algo:mala}) with
 step sizes
\begin{align}
  \step_\tagmrw = c_2 \cdot \frac{1}{\smoothness\dims\condition},
  \quad\text{and} \quad \step_\tagmala = c_1 \cdot
  \frac{1}{\smoothness\dims \cdot \max\braces{1,
      \sqrt{\condition/\dims} }}
\end{align}
respectively, satisfy the mixing time bounds
\begin{subequations}
\begin{align}
  \Tmix_2^\tagmrw(\errorparam; \initial) &=
  O\parenth{\dims\condition^2\log{\frac{d}{\errorparam}}},
  \quad\text{and}\quad\\ \Tmix_2^\tagmala(\errorparam; \initial) &=
  O\parenth{\dims\condition\log{\frac{d}{\errorparam}}\cdot
    \max\braces{1, \sqrt{\frac{\condition}{\dims}}}}.
\end{align}
\end{subequations}
\end{theorem}

The proof is omitted as it directly follows from the conductance profile based mixing time bound in Lemma~\ref{prop:mixing_bound_using_conductanceprofile}, Lemma~\ref{prop:conductanceprofile_via_overlaps} and the overlap bounds for MALA and MRW provided in
our prior work~\citep{dwivedi2018log}.
Theorem~\ref{thm:mala_mrw_mixing} states that the mixing time bounds
for MALA and MRW with the feasible distribution $\initialstar$ as the initial distribution
scale as $\Ot(\dims\condition \log\parenth{1/\errorparam})$ and $\Ot(\dims\condition^2\log\parenth{1/\errorparam})$.  Once again,
we note that in light of the
inequality~\eqref{eq:mixing_time_relation} we obtain same bounds for
the number of steps taken by these algorithms to mix within
$\errorparam$ total-variation distance of the target distribution
$\target$.  Consequently, our results improve upon the previously
known mixing time bounds for MALA and MRW~\citep{dwivedi2018log} for
strongly log-concave distributions. With $\initialstar$ as the initial
distribution, the authors had derived bounds of order
$\Ot(\dims^2\condition \log\parenth{1/\errorparam})$ and $\Ot(\dims^2\condition^2 \log\parenth{1/\errorparam})$ for MALA and MRW
respectively (cf. Corollary 1 in~\cite{dwivedi2018log}).  However, the numerical
experiments in that work suggested a better
dependency on the dimension for the mixing time. Indeed the mixing
time bounds from Theorem~\ref{thm:mala_mrw_mixing} are smaller by a
factor of $\frac{\dims}{\log \dims}$, compared to our earlier bounds in
the prior work~\citep{dwivedi2018log} for
both of these chains thereby resolving an open
question. Nonetheless, it is still an open question how to establish a lower bound on the mixing time of these sampling algorithms.


\section{Numerical experiments}
\label{sec:numerical_experiments}

In this section, we numerically compare HMC with MALA and MRW to
verify that our suggested step-size and leapfrog steps lead to faster
convergence for the HMC algorithm.  We adopt the step-size choices for
MALA and MRW given in~\cite{dwivedi2018log}, whereas
the choices for step-size and leapfrog steps for HMC are taken from
Corollary~\ref{cor:hmc_mixing_sc_fixedK} in this paper.
When the Hessian-Lipschitz constant $\hessianLip$ is small, our theoretical results
suggest that HMC can be run with much larger step-size
and much larger number of leapfrog steps (see Appendix~\ref{ssub:faster_mixing_time_bounds}).
Since our experiments make use of multivariate Gaussian target distribution,
the Hessian-Lipschitz constant $\hessianLip$ is always zero. Consequently
we also perform experiments with a more \emph{aggressive} choice of parameters,
i.e., larger step-size and number of leapfrog steps. We denote this
choice by HMCagg.

In this simulation, we check the dimension $\dims$ dependency and
condition number $\condition$ dependency in the multivariate Gaussian
case under our step-size choices. We consider sampling from the
multivariate Gaussian distribution with density
\begin{align}
\label{eq:gaussian_density}
\target(x) \propto e^{-\frac{1}{2} x\tp \Sigma^{-1} x},
\end{align}
for some covariance matrix $\Sigma \in \real^{\dims \times \dims}$.
The log density (disregarding constants) and its deriviatives are
given by
\begin{align*}
  \targetf(x) = \frac{1}{2} x\tp \Sigma^{-1} x,\quad \gradf(x) =
  \Sigma^{-1} x, \quad\text{and}\quad \hessf(x) = \Sigma^{-1}.
\end{align*}
Consequently, the function $\targetf$ is strongly convex with
parameter $\scparam = 1/\lambda_\text{max}(\Sigma)$ and smooth with
parameter $\smoothness = 1/\lambda_\text{min}(\Sigma)$.
Since $\Ell_\lp$-divergence can not be measure with finitely many samples,
we use the error in quantiles along different directions for convergence diagnostics.
Using the exact quantile information for each direction
for Gaussians, we measure the error in the $75\%$ quantile of the relative
sample distribution and the true distribution in the \emph{least
  favorable direction}, i.e., along the eigenvector of $\Sigma$
corresponding to the eigenvalue $\lambda_\text{max}(\Sigma)$.  The
\textit{quantile mixing time} is defined as the smallest iteration when this
relative error falls below a constant $\delta = 0.04$. We use $\initial = \Normal\parenth{0,
\smoothness^{-1}\Ind_\dims}$ as the initial distribution.
To make the comparison with MRW and MALA fair, we compare the number of total function and gradient evaluations instead of number of iterations.
For HMC, the number of gradient evaluations is $\internsteps$ times the number of outer-loop iterations.

\begin{figure}[ht]
\begin{tabular}{cc}
  \includegraphics[width=0.49\textwidth]{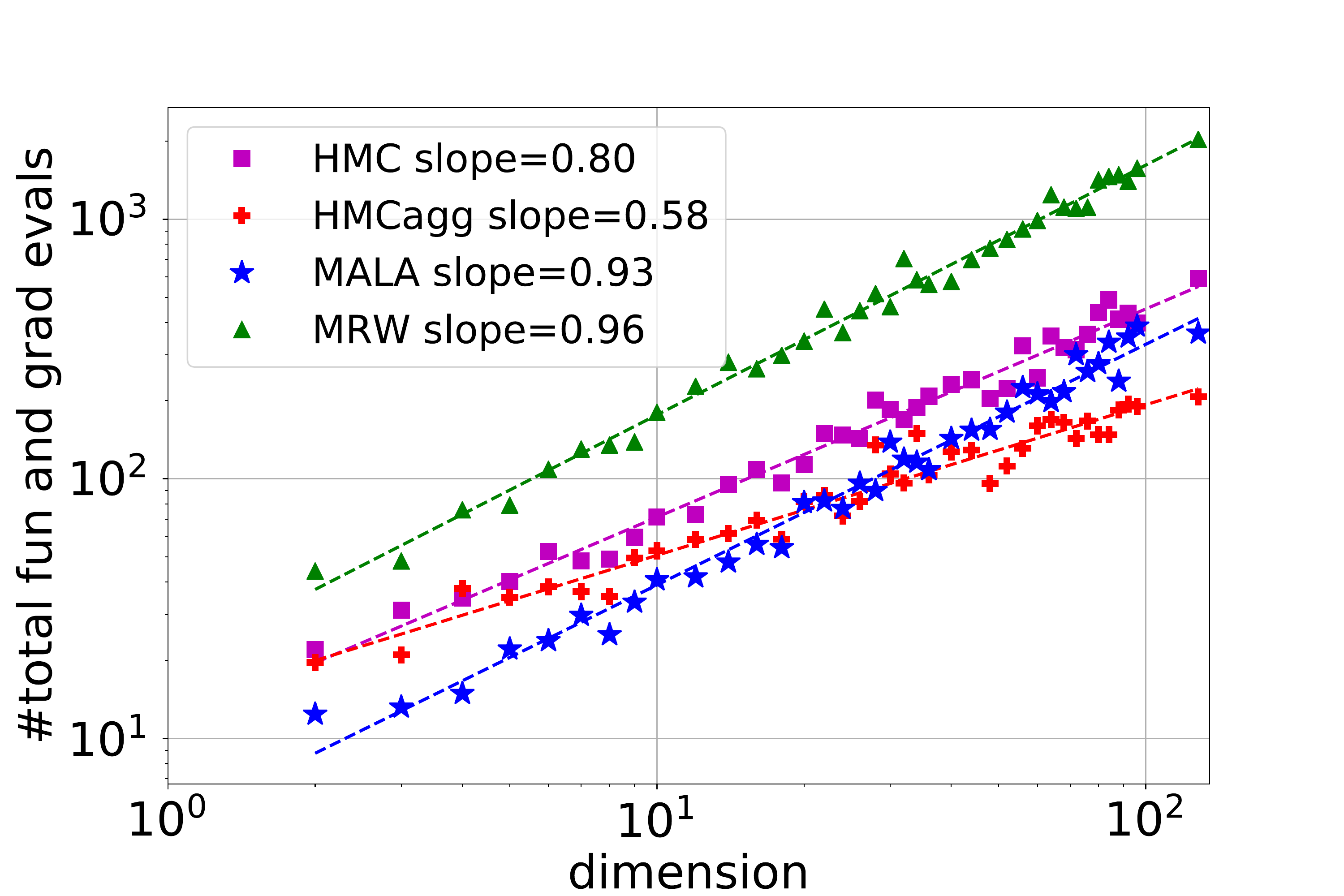}
  &
  \includegraphics[width=0.49\textwidth]{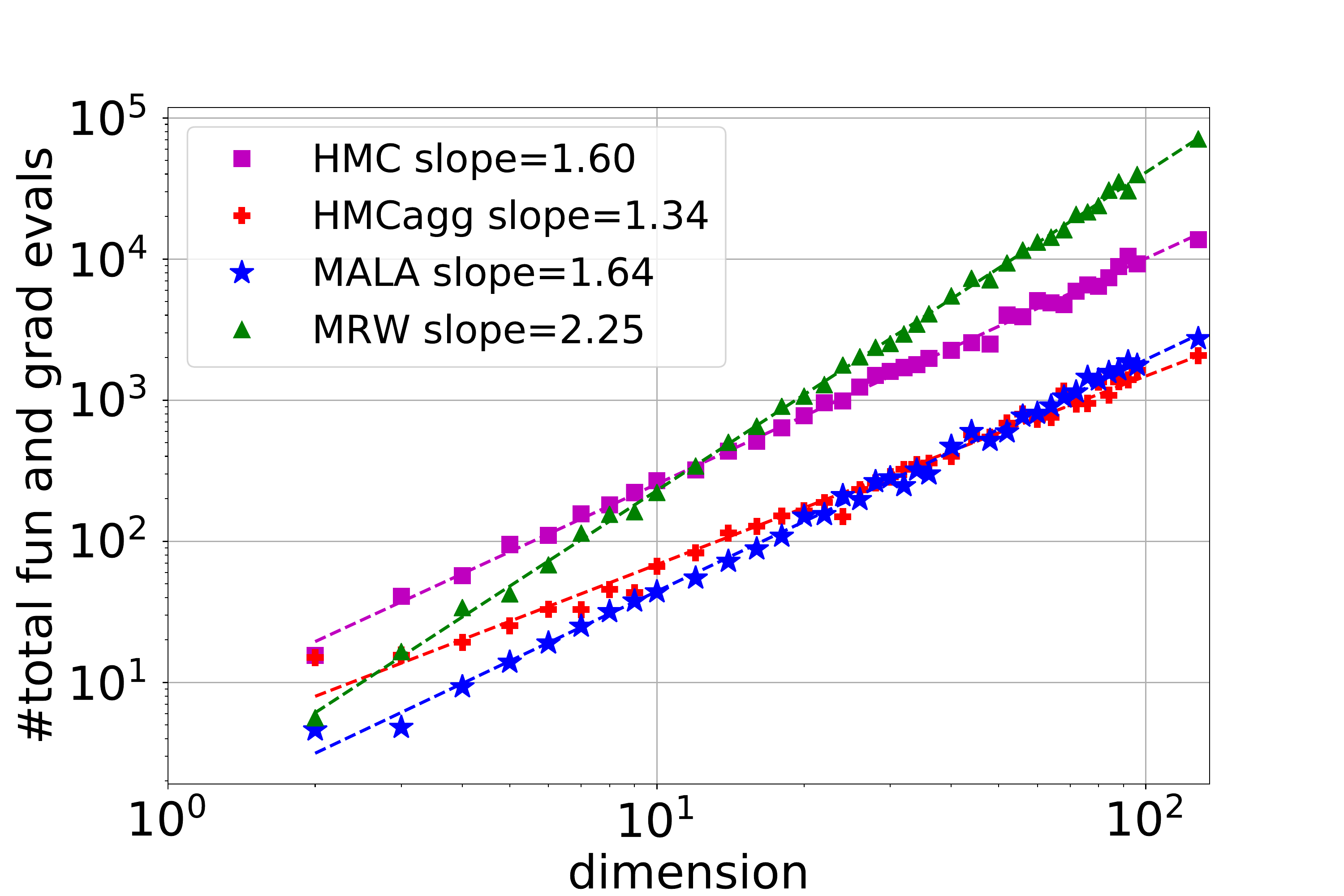}
  \\ (a) $\condition=4$ & (b) $\condition=\dims^{\frac{2}{3}}$
  \end{tabular}
\caption{Average number of total function and gradient evaluations as a function of dimension for four random walks on multivariate Gaussian density~\eqref{eq:gaussian_density} where
  the covariance has a condition number $\condition$ that is (a)
  constant $4$ and (b) scales with dimension $d$. With suggested step-size and leapfrog steps in Corollary~\ref{cor:hmc_mixing_sc_fixedK}, the number of total function and gradient evaluations of HMC has a smaller dimension dependency than that of MALA or MRW. Since the target distributon is Gaussian and the Hessian-Lipschitz constant $\hessianLip$ is zero, larger step-size and larger number of leapfrog steps can be chosen according to Appendix~\ref{ssub:faster_mixing_time_bounds}. The plots does show that HMCagg with larger step-size and larger number of leapfrog steps uses smaller number of total function and gradient evaluations to achieve the same quantile mixing.}
  \label{fig:nonisotropic_gaussian}
\end{figure}

For every case of simulation, the parameters for \hmc~are chosen according to the warm start case in
Corollary~\ref{cor:hmc_mixing_sc_fixedK} with $\internsteps = 4\cdot\dims^{1/4}$, and for MRW and MALA are
chosen according to the paper~\cite{dwivedi2018log}. As alluded to earlier,
we also run the HMC chain a more aggressive choice of parameters, and denote
this chain by  HMCagg. For HMCagg, both the step-size and leapfrog steps
are larger (Appendix~\ref{ssub:faster_mixing_time_bounds}): $\internsteps
= 4 \cdot \dims^{1/8} \condition^{1/4}$ where we take into account that
$\hessianLip$ is zero for Gaussian distribution. We simulate $100$
independent runs of the four chains, MRW, MALA, HMC, HMCagg, and for
each chain at every iteration we compute the quantile error across the $100$
samples from $100$ independent runs of that chain. We compute the minimum number of total function and gradient evaluations
required for the relative quantile error to fall below $\delta=0.04$.
We repeat this computation $10$ times and report the averaged number of
total function and gradient evaluations in Figure~\ref{fig:nonisotropic_gaussian}.
To examine the scaling of the number of evaluations with the dimension $d$,
we vary $\dims \in \braces{2, 4, \ldots, 128}$. For each chain, we also fit
a least squares line for the number of total function and gradient evaluations with
respect to dimension $\dims$ on the log-log scale, and report the slope in the
figure. Note that a slope of $\alpha$ would denote that the number of evaluations
scales with $\dims$ as $\dims^{\alpha}$.


\paragraph{(a) Dimension dependency for fixed $\condition$:}
\label{par:dimension_dependency}
First, we consider the case of fixed condition number.
We fix $\condition = 4$ while we vary the dimensionality $d$ of the target
distribution is varied over $ \braces{2, 4, \ldots, 128}$. The Hessian $\Sigma$ in the multivariate Gaussian
distribution is chosen to be diagonal and the square roots of its eigenvalues
are linearly spaced  between $1.0$ to $2.0$.
Figure~\ref{fig:nonisotropic_gaussian}(a) shows the dependency
of the number of total function and gradient evaluations as a function of dimension $\dims$ for the
four Markov chains on the log-log scale. The least-squares
fits of the slopes for HMC, HMCagg, MALA and MRW are $0.80 (\pm 0.12)$,
$0.58 (\pm 0.15)$, $0.93 (\pm 0.13)$ and $0.96 (\pm 0.10)$, respectively,
where standard errors of the regression coefficient is reported in the parentheses.
These numbers indicate close correspondence to the theoretical slopes (reported
in Table~\ref{tab:warm_mix} and Appendix~\ref{ssub:faster_mixing_time_bounds})
are $0.92, 0.63, 1.0, 1.0$ respectively.


\paragraph{(b) Dimension dependency for $\condition=\dims^{2/3}$:}
\label{par:dimension_dependency_for_condition}

Next, we consider target distributions such that their condition
number varies with $\dims$ as $\condition=\dims^{2/3}$, where
$\dims$ is varied from $2$ to $128$. To ensure such a scaling for $\condition$,
we choose the Hessian $\Sigma$ for the multivariate Gaussian distribution
to be diagonal and set the square roots of its eigenvalues linearly
spaced between $1.0$ to $\dims^{1/3}$. Figure~\ref{fig:nonisotropic_gaussian}(b)
shows the dependency of the number of total function and gradient evaluations as a function of
dimension $\dims$ for the four random walks on the log-log scale.
The least squares fits yield the slopes as $1.60 (\pm 0.09)$, $1.34 (\pm 0.17)$, $1.64 (\pm 0.11)$ and $2.25 (\pm 0.08)$ for HMC, HMCagg, MALA and MRW,
respectively, where standard errors of the regression coefficient are reported
in the parentheses. Recall that the theoretical guarantees
for HMC (Table~\ref{tab:hmc_param_choice_explicit}), HMCagg (Table~\ref{tab:hmc_param_choice_explicit_L3}),
MALA and MRW (Table~\ref{tab:warm_mix}) yield that these exponent should
be close to 1.58, 1.46, 1.67 and 2.33 respectively.
Once again, we observe a good agreement of the numerical results with that
of our theoretical results.

\paragraph{Remark:}
We would like to caution that the aggressive parameter choices for HMCagg
are well-informed only when the Hessian-Lipschitz constant $\hessianLip$
is small---which indeed is the case for the Gaussian target distributions
considered above.
When general log-concave distributions are considered, one may use the
more general choices recommended in Corollary~\ref{cor:hmc_mixing_sc_general}.
See Appendix~\ref{sec:discussion_around_corollary_cor:hmc_mixing_sc_fixedK}
for an in-depth discussion on different scenarios and the optimal parameter
choices derived from our theory.


\section{Proofs}
\label{sec:proof_of_the_main_theorem}

This section is devoted primarily to the proof of
Theorem~\ref{thm:hmc_mixing_general}.  In order to do so, we begin
with the mixing time bound based on the conductance profile from
Lemma~\ref{prop:mixing_bound_using_conductanceprofile}.  We then seek
to apply Lemma~\ref{prop:conductanceprofile_via_overlaps} in order
derive a bound on the conductance profile itself.  However, in order
to do so, we need to derive bound on the overlap between the proposal
distributions of HMC at two nearby points and show that the
Metropolis-Hastings step only modifies the proposal distribution by a
relatively small amount. This control is provided by
Lemma~\ref{lem:transition_closeness}, stated in
Section~\ref{ssub:overlap_bounds_for_hmc}.  We use it to prove
Theorem~\ref{thm:hmc_mixing_general} in
Section~\ref{ssub:proof_of_theorem_thm:hmc_mixing_general}.  Finally,
Section~\ref{sub:lem:transition_closeness} is devoted to the proof of
Lemma~\ref{lem:transition_closeness}.
We provide a sketch-diagram for how various main results of the paper interact
with each other in Figure~\ref{fig:proof_sketch}.


\subsection{Overlap bounds for HMC}
\label{ssub:overlap_bounds_for_hmc}

In this subsection, we derive two important bounds for the
Metropolized HMC chain: (1) first, we quantify the overlap between
proposal distributions of the chain for nearby points, and, (2)
second, we show that the distortion in the proposal distribution
introduced by the Metropolis-Hastings accept-reject step can be
controlled if an appropriate step-size is chosen. Putting the two
pieces together enables us to invoke
Lemma~\ref{prop:conductanceprofile_via_overlaps} to prove
Theorem~\ref{thm:hmc_mixing_general}.

In order to do so, we begin with some notation.  Let $\transition$
denote the transition operator of the HMC chain with leapfrog
integrator taking step-size $\step$ and number of leapfrog updates
$\internsteps$. Let $\proposal_x$ denote the proposal distribution at
$x \in \statespace$ for the chain before the accept-reject step and
the lazy step. Let $\transitionbflazy_x$ denote the corresponding
transition distribution after the proposal and the accept-reject step,
before the lazy step. By definition, we have
\begin{align}
\label{eq:transition_lazy_decomposition}
  \transition_x(A) = \lazyparam \dirac_x(A) + (1-\lazyparam)
  \transitionbflazy_x(A) \qquad \mbox{for any measurable set $A \in
    \borel(\statespace)$.}
\end{align}
Our proofs make use of the Euclidean ball $\truncballres$ defined in
equation~\eqref{eq:def_truncball}. At a high level, the HMC chain has
bounded gradient inside the ball $\truncballres$ for a suitable choice
of $\res$, and the gradient of the log-density gets too large outside
such a ball making the chain unstable in that region.  However, since
the target distribution has low mass in that region, the chain's visit
to the region outside the ball is a rare event and thus we can focus
on the chain's behavior inside the ball to analyze its mixing time.


In the next lemma, we state the overlap bounds for the transition
distributions of the HMC chain.  For a fixed univeral constant $c$, we
require
\begin{subequations}
  \begin{align}
    \label{eq:step_K_condition_thm_proof}
    \internsteps^2\step^2 &\leq \frac{1}{4
      \max\braces{\dims^{\frac{1}{2}}\smoothness,
        \dims^{\frac{2}{3}}\hessianLip^{\frac{2}{3}}}}, \qquad
    \mbox{and} \\
    \label{eq:step_condition_thm_proof}
    \step^2 & \leq \frac{1}{c \smoothness}
    \min\braces{\frac{1}{\internsteps^2},
      \frac{1}{\internsteps\dims^{\frac{1}{2}}},
      \frac{1}{\internsteps^{\frac{2}{3}}\dims^{\frac{1}{3}}\parenth{\frac{\gradbound^2}{\smoothness}}^{\frac{1}{3}}},
      \frac{1}{\internsteps\frac{\gradbound}{\smoothness^{\frac{1}{2}}}},
      \frac{1}{\internsteps^{\frac{2}{3}}\dims}\frac{\smoothness}{\hessianLip^
        {\frac{2}{3}}},
      \frac{1}{\internsteps^{\frac{4}{3}}\frac{\gradbound}{\smoothness^{\frac{1}{2}}}}\parenth{\frac{\smoothness}{\hessianLip^{\frac{2}{3}}}}^{\frac{1}{2}}}.
  \end{align}
\end{subequations}

\begin{lemma}
  \label{lem:transition_closeness}
Consider a $(\smoothness, \hessianLip, \res, \isoconst_{\logisopower},
\gradbound)$-regular target distribution
(cf. Assumption~\ref{itm:assumptionA}) with $\convset$ the convex
measurable set
satisfying~\eqref{eq:assumption_high_mass_gradbound}. Then with the
parameters $\parenth{\internsteps, \step}$ satisfying
$\internsteps\step \leq \frac{1}{4\smoothness}$ and
condition~\eqref{eq:step_K_condition_thm_proof}, the
HMC-$\parenth{\internsteps, \step}$ chain satisfies
\begin{subequations}
  \begin{align}
  \label{eq:proposal_closeness}
    \sup_{\vecnorm{\hmcstate_0 - \hmcstatey_0}{2}\leq
      \frac{\internsteps\step}{4}}\tvdist{\proposal_{\hmcstate_0}}{\proposal_{\hmcstatey_0}}
    \leq \frac{1}{2}.
  \end{align}
If, in addition, condition~\eqref{eq:step_condition_thm_proof} holds,
then we have
\begin{align}
\label{eq:transition_proposal_overlap}
\sup_{x \in \convset} \tvdist{\proposal_x}{\transitionbflazy_x} \leq
\frac{1}{8}.
\end{align}
\end{subequations}
\end{lemma}
\noindent See
Appendix~\ref{sub:lem:transition_closeness} for the
proof.

Lemma~\ref{lem:transition_closeness} is crucial to the analysis of HMC
as it enables us to apply the conductance profile based bounds
discussed in
Section~\ref{sub:mixing_time_of_a_markov_chain_with_conductance_profile}.
It reveals two important properties of the Metropolized HMC. First,
from equation~\eqref{eq:proposal_closeness}, we see that proposal
distributions of HMC at two different points are close if the two
points are close. This is proved by controlling the KL-divergence of
the two proposal distributions of HMC via change of variable formula.
Second, equation~\eqref{eq:transition_proposal_overlap} shows that the
accept-reject step of HMC is well behaved inside $\convset$ provided
the gradient is bounded by $\gradbound$.


\subsection{Proof of Theorem~\ref{thm:hmc_mixing_general}}
\label{ssub:proof_of_theorem_thm:hmc_mixing_general}

We are now equipped to prove our main theorem.  In order to prove
Theorem~\ref{thm:hmc_mixing_general}, we begin by using
Lemma~\ref{prop:conductanceprofile_via_overlaps} and
Lemma~\ref{lem:transition_closeness} to derive an explicit bound for
on the HMC conductance profile. Given the assumptions of
Theorem~\ref{thm:hmc_mixing_general},
conditions~\eqref{eq:step_K_condition_thm_proof}
and~\eqref{eq:step_condition_thm_proof} hold, enabling us to invoke
Lemma~\ref{lem:transition_closeness} in the proof.

Define the function $\MYPSI_\convset: [0, 1] \mapsto \real_+$ as
\begin{align}
\label{eq:psi_define}
  \MYPSI_\convset(v) &= \begin{cases} \displaystyle\frac{1}{32}\cdot
    \min\braces{1, \frac{\internsteps\step}{64\isoconst_\logisopower}
      \log^{\logisopower}\parenth{\frac{1}{v}}} & \mbox{if $v \in
      \left[0, \frac{1-\res}{2} \right]$.}  \\[3mm] \displaystyle
    \frac{\internsteps\step} {2048\isoconst_\logisopower}, & \mbox{if
      $v \in \left(\frac{1-\res}{2}, 1 \right]$.}
  \end{cases}
\end{align}
This function acts as a lower bound on the truncated conductance
profile. Define the Euclidean ball
\begin{align}
  \label{eq:def_truncball}
  \truncballres = \ball\parenth{\xstar,
    \radius(\res)\sqrt{\frac{\dims} {\scparam}}},
\end{align}
and consider a pair $(x, y) \in \truncballres$ such that $\vecnorm{x -
  y}{2} \leq \frac{1}{4}\internsteps\step$.  Invoking the
decomposition~\eqref{eq:transition_lazy_decomposition} and applying
triangle inequality for $\lazyparam$-lazy HMC, we have
\begin{align*}
  \tvdist{\transition_x}{\transition_y} &\leq \lazyparam + \parenth{1
    - \lazyparam}
  \tvdist{\transitionbflazy_x}{\transitionbflazy_y} \\
& \leq \lazyparam + \parenth{1 - \lazyparam}
  \parenth{\tvdist{\transitionbflazy_x}{\proposal_y} +
    \tvdist{\proposal_x}{\proposal_y} +
    \tvdist{\proposal_x}{\transitionbflazy_y}} \\
& \stackrel{(i)}{\leq} \lazyparam + \parenth{1 - \lazyparam} \parenth{
    \frac{1} {4}+\frac{1}{2} + \frac{1}{4}} \\
& = 1- \frac{1 - \lazyparam}{4},
\end{align*}
where step~(i) follows from the bounds~\eqref{eq:proposal_closeness}
and~\eqref{eq:transition_proposal_overlap} from
Lemma~\ref{lem:transition_closeness}.  For $\lazyparam=\frac{1}{2}$,
substituting $\omega = \frac{1}{8}$, $\Delta =
\frac{1}{4}\internsteps\step$ and the convex set $\convset =
\truncballres$ into Lemma~\ref{prop:conductanceprofile_via_overlaps},
we obtain that
\begin{align*}
  \conductanceprofile(v) \geq \frac{1}{32}\cdot \min\braces{1,
    \frac{\internsteps\step}{64\isoconst_\logisopower}
    \log^{\logisopower}\parenth{1 + \frac{1}{v}}},\quad\text{for }
  v\in\brackets{0, \frac{1-\res}{2}}.
\end{align*}
Here $\logisopower$ equals to $\frac{1}{2}$ or $0$, depending on the
assumption~\eqref{eq:assumption_isoperimetric}.  By the definition of
the truncated conductance profile~\eqref{eq:def_truncatedprofile}, we
have that $\truncconductanceprofile(v) \geq
\frac{\internsteps\step}{2048\isoconst_\logisopower}$ for
$v\in\brackets{ \frac{1-\res}{2}, 1}$. As a consequence,
$\MYPSI_\convset$ is effectively a lower bound on the truncated
conductance profile.  Note that the assumption~\ref{itm:assumptionA}
ensures the existence of $\convset$ such that $\target(\convset) \geq
1- \res$ for $\res = \frac{\errorparam^2}{3\warmparam^2}$.  Putting
the pieces together and applying
Lemma~\ref{prop:mixing_bound_using_conductanceprofile} with the convex
set $\convset$ concludes the proof of the theorem.


\subsection{Proof of Lemma~\ref{lem:transition_closeness}}
\label{sub:lem:transition_closeness}

In this subsection, we prove the two main
claims~\eqref{eq:proposal_closeness}
and~\eqref{eq:transition_proposal_overlap} in
Lemma~\ref{lem:transition_closeness}. Before going into the claims, we
first provide several convenient properties about the HMC proposal.


\subsubsection{Properties of the HMC proposal}
\label{sub:properties_of_the_hmc_proposal}

Recall the Hamiltonian Monte Carlo (HMC) with leapfrog
integrator~\eqref{eq:hmc_leapfrog_integrator}.  Using an induction
argument, we find that the final states in one iteration of
$\internsteps$ steps of the HMC chain, denoted by
$\hmcstate_\internsteps$ and $\hmcnoise_\internsteps$ satisfy
\begin{subequations}
\begin{align}
  \label{eq:hmcnoise_integrate_expression_K}
  \hmcnoise_\internsteps &= \hmcnoise_0 -
  \frac{\step}{2}\gradf(\hmcstate_0) -
  \sum_{j=1}^{\internsteps-1}\gradf(\hmcstate_j) -
  \frac{\step}{2}\gradf(\hmcstate_\internsteps),\\
  \label{eq:hmcstate_integrate_expression_K}
  \quad\text{and}\quad\hmcstate_\internsteps &= \hmcstate_0 + \internsteps
  \step \hmcnoise_0 - \frac{\internsteps \step^2}{2} \gradf(\hmcstate_0) - \step^2 \sum_{j=1}^{\internsteps-1}\parenth{\internsteps-j} \gradf(\hmcstate_j).
\end{align}
It is easy to see that for $k \in \brackets{\internsteps}$, $\hmcstate_k$
can be seen as a function of the initial state $\hmcstate_0$ and
$\hmcnoise_0$.
We denote this function as the \emph{forward mapping} $\mapF$,
\begin{align}
  \label{eq:def_forward_mapping_F}
  \hmcstate_k \rdefn \mapFk(\hmcnoise_0, \hmcstate_0)
  \quad\text{and}\quad
  \hmcstate_\internsteps \rdefn \mapF_{\internsteps}(\hmcnoise_0, \hmcstate_0)
  \rdefn \mapF \parenth{\hmcnoise_0,\hmcstate_0}
\end{align}
where we introduced the simpler notation $\mapF\defn \mapF_{\internsteps}$ for
the final iterate.
The forward mappings $\mapFk$ and $\mapF$ are deterministic functions that
only depends on the gradient $\gradf$, the number of leapfrog updates
$\internsteps$ and the step size $\step$.

Denote $\jacob_x \mapF$ as the Jacobian matrix of the forward mapping
$\mapF$ with respect to the first variable. By definition, it
satisfies
\begin{align}
\label{eq:jacobian_f}
  \brackets{\jacob_x \mapF(x, q_0)}_{ij} = \frac{\partial}{\partial
    x_j} \brackets{\mapF(x, q_0)}_i, \quad \text{for all} \quad i, j
  \in \brackets{\dims}.
\end{align}
\end{subequations}
Similarly, denote $\jacob_y \mapF$ as the Jacobian matrix of the
forward mapping $\mapF$ with respect to the second variable. The
following lemma characterizes the eigenvalues of the Jacobian
$\jacob_x \mapF$.
\begin{lemma}
  \label{lem:jacob_F_x}
  Suppose the log density $\targetf$ is $\smoothness$-smooth. For the
  number of leapfrog steps and step-size satisfying
  $\internsteps^2\step^2 \leq \frac{1}{4\smoothness}$, we have
  \begin{align*}
  \matsnorm{\internsteps \step \Ind_d - \jacob_x\mapF(x, y)}{2} \leq
  \frac{1}{8} \internsteps \step, \quad \text{for all}\quad x, y
  \in \statespace \text{ and } i \in \brackets{\dims}.
  \end{align*}
  Also all eigenvalues of $\jacob_x \mapF(x, y)$ have absolute value
  greater or equal to $\frac{7}{8} \internsteps \step$.
\end{lemma}
\noindent See Appendix~\ref{ssub:proof_of_lemma_lem:jacob_f_x} for the
proof.\\

Since the Jacobian is invertible for $\internsteps^2 \step^2 \leq
\frac{1} {4\smoothness}$, we can define the inverse function of
$\mapF$ with respect to the first variable as the backward mapping
$\mapG$. We have
\begin{align}
  \label{eq:def_backward_mapping_G}
  \mapF\parenth{\mapG(x, y), y} = x, \quad \text{for all}\quad x, y \in \statespace.
\end{align}
Moreover as a direct consequence of Lemma~\ref{lem:jacob_F_x}, we
obtain that the magnitude of the eigenvalues of the Jacobian matrix
$\jacob_x \mapG(x, y)$ lies in the interval
$\brackets{\frac{8}{9\internsteps\step}, \frac{8}{7\internsteps\step}}$.  In
the next lemma, we state another set of bounds on different Jacobian
matrices:
\begin{lemma}
  \label{lem:jacob_G_y}
  Suppose the log density $\targetf$ is $\smoothness$-smooth. For the
  number of leapfrog steps and step-size satisfying
  $\internsteps^2\step^2 \leq \frac{1}{4\smoothness}$, we have
  \begin{subequations}
  \begin{align}
    \matsnorm{\jacob_y \mapG(x, y)}{2} &\leq
    \frac{4}{3\internsteps\step}, \quad \text{for all}\quad x, y
    \in \statespace, \quad\text{and}\quad
    \label{eq:jacob_g_y_bound}\\
    \matsnorm{\frac{\partial\mapFk(\mapG(x, y), y)}{\partial y}}{2} &\leq
    3, \quad\quad\text{for all}\quad k \in \brackets{\internsteps}.
    \label{eq:fk_y_bound}
  \end{align}
  \end{subequations}
\end{lemma}
\noindent See Appendix~\ref{ssub:proof_of_lemma_lem:jacob_g_y} for the
proof.\\

Next, we would like to obtain a bound on the quantity $\frac{\partial
  \log\det\jacob_xG(x, \hmcstate_0)}{\partial y}$. Applying the chain
rule, we find that
\begin{align}
  \label{eq:grady_log_det_jacob_G_x_trace_form}
  \frac{\partial \log\det\jacob_xG(x, \hmcstate_0)}{\partial y}
  = \begin{bmatrix}\trace\parenth{[\jacob_x \mapG(x,
        \hmcstate_0)]^{-1} \jacob_{xy_1} \mapG(x, \hmcstate_0)}
    \\ \vdots \\ \trace\parenth{[\jacob_x \mapG(x, q_0)]^{-1}
      \jacob_{xy_\dims} \mapG(x, \hmcstate_0)}
    \end{bmatrix}.
\end{align}
Here $\jacob_{xy}\mapG(x, \hmcstate_0)$ is a third order tensor and we
use $\jacob_{xy_l} \mapG(x, \hmcstate_0)$ to denote the matrix
corresponding to the $l$-th slice of the tensor which satisfies
\begin{align*}
  \brackets{\jacob_{xy_l} \mapG(x, \hmcstate_0)}_{ij} =
  \frac{\partial\partial}{\partial x_j y_l} \brackets{\mapF(x,
    \hmcstate_0)}_i, \ \quad \text{for all} \quad i, j, l \in \brackets{\dims}.
\end{align*}
\begin{lemma}
  \label{lem:grady_log_det_jacob_G_x}
  Suppose the log density $\targetf$ is $\smoothness$-smooth and
  $\hessianLip$-Hessian Lipschitz. For the number of leapfrog steps
  and step-size satisfying $\internsteps^2\step^2 \leq
  \frac{1}{4\smoothness}$, we have
  \begin{align*}
    \vecnorm{\frac{\partial \log\det\jacob_xG(x,
        \hmcstate_0)}{\partial y}}{2} =
    \vecnorm{\begin{bmatrix}\trace\parenth{[\jacob_x \mapG(x,
            \hmcstate_0)]^{-1} \jacob_{xy_1} \mapG(x, \hmcstate_0)}
        \\ \vdots \\ \trace\parenth{[\jacob_x \mapG(x, q_0)]^{-1}
          \jacob_{xy_\dims} \mapG(x, \hmcstate_0)}
    \end{bmatrix}}{2}
   \leq 2\dims \internsteps^2 \step^2 \hessianLip.
  \end{align*}
\end{lemma}
\noindent See
Appendix~\ref{ssub:proof_of_lemma_lem:grady_log_det_jacob_g_x} for the
proof.\\

As a direct consequence of the equation~\eqref{eq:hmcstate_integrate_expression_K} at $k$-th step of leapfrog updates, we obtain the following two bounds for the difference between successive $\mapFk$ terms that come in handy later in our proofs.
\begin{lemma}
  \label{lem:q_j_recursion_bounds}
  Suppose that the log density $\targetf$ is $\smoothness$-smooth. For
  the number of leapfrog steps and step-size satisfying
  $\internsteps^2\step^2 \leq \frac{1}{4\smoothness}$, we have
  \begin{subequations}
    \begin{align}
        \label{eq:q_j_recursion_bound}
        \vecnorm{\mapFn{k}(\hmcnoise_0,\hmcstate_0)- \hmcstate_0}{2}
        &\leq 2 k \step \vecnorm{\hmcnoise_0}{2} + 2 k^2 \step^2
        \vecnorm{\gradf (\hmcstate_0)}{2}\ \ \text{for } k \in
        \brackets{\internsteps},
        \ \text{and}\\ \vecnorm{\mapFn{k+1}(\hmcnoise_0,\hmcstate_0) -
          \mapFn{k}(\hmcnoise_0,\hmcstate_0)} {2} &\leq 2 \step
        \vecnorm{\hmcnoise_0}{2} + 2(k+1) \step^2 \vecnorm{\gradf
          (\hmcstate_0)}{2}\ \text{for } k \in \brackets{\internsteps-1}.
        \label{eq:q_j_diff_recursion_bound}
    \end{align}
\end{subequations}
\end{lemma}
\noindent See Appendix~\ref{ssub:proof_of_lemma_lem:q_j_recursion_bounds}
for the proof.\\

We now turn to the proof the two claims in
Lemma~\ref{lem:transition_closeness}.  Note that the
claim~\eqref{eq:proposal_closeness} states that the proposal
distributions at two close points are close; the
claim~\eqref{eq:transition_proposal_overlap} states that the proposal
distribution and the transition distribution are close.


\subsubsection{Proof of claim~\texorpdfstring{\eqref{eq:proposal_closeness}}{proposal-proposal-overlap} in Lemma~\ref{lem:transition_closeness}}
\label{sub:proof_of_claim_eq:proposal_closeness}

In order to bound the distance between proposal distributions of
nearby points, we prove the following stronger claim: For a
$\smoothness$-smooth $\hessianLip$-Hessian-Lipschitz target
distribution, the proposal distribution of the HMC algorithm with step
size $\step$ and leapfrog steps $\internsteps$ such that
$\internsteps\step \leq \frac{1}{4\smoothness}$ satisfies
  \begin{align}
    \label{eq:proposal_closeness_stronger}
    \tvdist{\proposal_{\hmcstate_0}}{\proposal_{\hmcstatey_0}} \leq
    \parenth{\frac{2\vecnorm{\hmcstate_0 -
          \hmcstatey_0}{2}^2}{\internsteps^2\step^2} + 3 \sqrt{\dims}
      \internsteps\step \smoothness \vecnorm{\hmcstate_0-\hmcstatey_0}
                        {2} + 4 \dims \internsteps^2\step^2
                        \hessianLip \vecnorm{\hmcstate_0 -
                          \hmcstatey_0}{2} }^{1/2},
  \end{align}
  for all $\hmcstate_0, \hmcstatey_0 \in \real^\dims$.  Then for any
  two points $\hmcstate_0, \hmcstatey_0$ such that
  $\vecnorm{\hmcstate_0 - \hmcstatey_0}{2}\leq
  \frac{1}{4}\internsteps\step$, under the
  condition~\eqref{eq:step_K_condition_thm_proof}, i.e.,
  $\internsteps^2\step^2 \leq \frac{1}{4\max
    \braces{\dims^{\frac{1}{2}}\smoothness,
      \dims^{\frac{2}{3}}\hessianLip^{\frac{2}{3}}}}$, we have
\begin{align*}
  \tvdist{\proposal_{\hmcstate_0}}{\proposal_{\hmcstatey_0}} \leq
  \parenth{\frac{1}{8} + \frac{3}{64} + \frac{1}{64}}^{1/2} \leq
  \frac{1}{2},
\end{align*}
and the claim~\eqref{eq:proposal_closeness} follows.

The proof of claim~\eqref{eq:proposal_closeness_stronger} involves the
following steps: (1) we make use of the update
rules~\eqref{eq:hmcstate_integrate_expression_K} and change of
variable formula to obtain an expression for the density of
$\hmcstate_{\nbsteps}$ in terms of $\hmcstate_0$, (2) then we use
Pinsker's inequality and derive expressions for the KL-divergence
between the two proposal distributions, and (3) finally, we upper
bound the KL-divergence between the two distributions using different
properties of the forward mapping $\mapF$ from
Appendix~\ref{sub:properties_of_the_hmc_proposal}.

According to the update
rule~\eqref{eq:hmcstate_integrate_expression_K}, the proposals from
two initial points $\hmcstate_0$ and $\hmcstatey_0$ satisfy
respectively
\begin{align*}
  \hmcstate_\internsteps = \mapF(\hmcnoise_0, \hmcstate_0),\quad\text{
    and }\quad \hmcstatey_\internsteps = \mapF(\hmcnoisey_0,
  \hmcstatey_0),
\end{align*}
where $\hmcnoise_0$ and $\hmcnoisey_0$ are independent random variable
from Gaussian distribution $\Normal(0, \Ind_\dims)$.

Denote $\density_{\hmcstate_0}$ as the density function of the
proposal distribution $\proposal_{\hmcstate_0}$. For two different
initial points $\hmcstate_0$ and $\hmcstatey_0$, the goal is to bound
the total variation distance between the two proposal distribution,
which is by definition
\begin{align}
  \label{eq:proposal_hmc_overlap_TV_distance}
  \tvdist{\proposal_{\hmcstate_0}}{\proposal_{\hmcstatey_0}} =
  \frac{1}{2} \int_{x \in \statespace} \abss{\density_{\hmcstate_0}(x)
    - \density_{\hmcstatey_0}(x)} dx.
\end{align}
Given $\hmcstate_0$ fixed, the random variable
$\hmcstate_\internsteps$ can be seen as a transformation of the
Gaussian random variable $\hmcnoise_0$ through the function
$\mapF(\cdot, \hmcstate_0)$. When $\mapF$ is invertible, we can use
the change of variable formula to obtain an explicit expression of the
density $\density_{\hmcstate_0}$:
\begin{align}
  \label{eq:density_hmc_change_of_variable}
  \density_{\hmcstate_0} (x) = \densityNormal\parenth{\mapG(x,
    \hmcstate_0)} \det\parenth{\jacob_x \mapG(x, \hmcstate_0)},
\end{align}
where $\densityNormal$ is the density of the standard Gaussian
distribution $\Normal(0, \Ind_\dims)$. Note that even though explicit,
directly bounding the total variation
distance~\eqref{eq:proposal_hmc_overlap_TV_distance} using the
complicated density
expression~\eqref{eq:density_hmc_change_of_variable} is difficult.  We
first use Pinsker's inequality \citep{Cover} to give an upper bound of
the total variance distance in terms of KL-divergence
\begin{align}
  \label{eq:Pinsker_inequality}
  \tvdist{\proposal_{\hmcstate_0}}{\proposal_{\hmcstatey_0}} \leq
  \sqrt{2 \kldiv{\proposal_{\hmcstate_0}}{\proposal_{\hmcstatey_0}}},
\end{align}
and then upper bound the KL-divergence.  Plugging the
density~\eqref{eq:density_hmc_change_of_variable} into the
KL-divergence formula, we obtain that
\begin{align}
  \label{eq:KL_divergence_explicit_form}
  \kldiv{\proposal_{\hmcstate_0}}{\proposal_{\hmcstatey_0}} &=
  \int_{\real^d} \density_{\hmcstate_0}(x)
  \log\parenth{\frac{\density_
      {\hmcstate_0}(x)}{\density_{\hmcstatey_0}(x)}} dx \notag \\
  & = \int_{\real^d} \density_{\hmcstate_0}(x)
  \brackets{\log\parenth{\frac{\densityNormal\parenth{\mapG(x,
          \hmcstate_0)}}{\densityNormal\parenth{\mapG(x,
          \hmcstatey_0)}}} + \log\det\jacob_x\mapG(x, \hmcstate_0) -
    \log\det\jacob_x\mapG(x, \hmcstatey_0)} dx \notag \\ &=
  \underbrace{\int_{\real^d} \density_{\hmcstate_0}(x) \brackets{
      \frac{1}{2}\parenth{-\vecnorm{\mapG(x, \hmcstate_0)}{2}^2 +
        \vecnorm{\mapG (x, \hmcstatey_0)}{2}^2}} dx}_{T_1} \notag\\
  & \quad\quad\quad\quad + \underbrace{\int_{\real^d}
    \density_{\hmcstate_0} (x) \brackets{\log\det\jacob_x\mapG(x,
      \hmcstate_0) - \log\det\jacob_x\mapG(x, \hmcstatey_0)} dx}_{T_2}
\end{align}
We claim the following bounds on the terms $T_1$ and $T_2$:
\begin{subequations}
  \begin{align}
  \label{eq:bound_on_T1}
  \abss{T_1} & \leq \frac{8}{9}\frac{\vecnorm{\hmcstate_0 -
      \hmcstatey_0} {2}^2}{\internsteps^2\step^2} +
  \frac{3}{2}\sqrt{\dims} \internsteps \step \smoothness
  \vecnorm{\hmcstate_0 - \hmcstatey_0}{2}, \quad\text{and} \\
\label{eq:bound_on_T2}
  \abss{T_2} & \leq 2\dims \internsteps^2\step^2
  \hessianLip\vecnorm{\hmcstate_0 - \hmcstatey_0} {2},
  \end{align}
\end{subequations}
where the bound on $T_2$ follows readily from
Lemma~\ref{lem:grady_log_det_jacob_G_x}:
\begin{align}
  \label{eq:T2_final_bound}
  \abss{T_2} & = \abss{\int \density_{\hmcstate_0}(x)
    \brackets{\log\det\jacob_x\mapG(x, \hmcstate_0) -
      \log\det\jacob_x\mapG(x, \hmcstatey_0)} dx} \notag \\ &\leq
  \vecnorm{\frac{\partial \log\det\jacob_xG(x, \hmcstate_0)}{\partial
      y}}{2} \vecnorm{\hmcstate_0 - \hmcstatey_0}{2} \notag \\ &\leq
  2\dims \internsteps^2\step^2 \hessianLip\vecnorm{\hmcstate_0 -
    \hmcstatey_0}{2}.
\end{align}
Putting together the inequalities~\eqref{eq:Pinsker_inequality},
\eqref{eq:KL_divergence_explicit_form}, \eqref{eq:bound_on_T1}
and~\eqref{eq:bound_on_T2} yields the
claim~\eqref{eq:proposal_closeness_stronger}.\\

\noindent It remains to prove the bound~\eqref{eq:bound_on_T1} on
$T_1$.


\paragraph{Proof of claim~\eqref{eq:bound_on_T1}:}
\label{par:proof_of_claim_eq:bound_on_t1_}

For the term $T_1$, we observe that
\begin{align*}
  \frac{1}{2}\parenth{\vecnorm{\mapG(x,
      \hmcstatey_0)}{2}^2-\vecnorm{\mapG (x, \hmcstate_0)}{2}^2} =
  \frac{1}{2}\vecnorm{\mapG(x, \hmcstate_0)-\mapG(x,
    \hmcstatey_0)}{2}^2 - \parenth{\mapG(x, \hmcstate_0) - \mapG(x,
    \hmcstatey_0)}\tp \mapG(x, \hmcstate_0).
\end{align*}
The first term on the RHS can be bounded via the Jacobian of $\mapG$
with respect to the second variable.  Applying the
bound~\eqref{eq:jacob_g_y_bound} from Lemma~\ref{lem:jacob_G_y}, we
find that
\begin{align}
  \label{eq:T1_first_part}
  \vecnorm{\mapG(x, \hmcstate_0)-\mapG(x, \hmcstatey_0)}{2} \leq
  \matsnorm{\jacob_y\mapG(x, y)}{2} \vecnorm{\hmcstate_0 -
    \hmcstatey_0)}{2} \leq \frac{4}{3\internsteps \step}
  \vecnorm{\hmcstate_0 - \hmcstatey_0)}{2}.
\end{align}
For the second part, we claim that there exists a deterministic
function $C$ of $\hmcstate_0$ and $\hmcstatey_0$ and independent of
$x$, such that
\begin{align}
  \label{eq:T1_diff_G_residual_bound}
  \vecnorm{\mapG(x, \hmcstate_0) - \mapG(x, \hmcstatey_0) -
    C(\hmcstate_0, \hmcstatey_0)}{2} \leq \frac{3}{2}\internsteps
  \step\smoothness \vecnorm{\hmcstate_0 - \hmcstatey_0}{2}.
\end{align}
Assuming the claim~\eqref{eq:T1_diff_G_residual_bound} as given at the
moment, we can further decompose the second part of $T_1$ into two
parts:
\begin{align}
  \label{eq:T1_second_part_decompose}
  \parenth{\mapG(x, \hmcstate_0) - \mapG(x, \hmcstatey_0)}\tp\mapG(x,
  \hmcstate_0) =\parenth{\mapG(x, \hmcstate_0) - \mapG(x,
    \hmcstatey_0) - C(\hmcstate_0, \hmcstatey_0)}\tp\mapG(x,
  \hmcstate_0) + C(\hmcstate_0, \hmcstatey_0)\tp\mapG(x, \hmcstate_0)
\end{align}
Applying change of variables along with
equation~\eqref{eq:density_hmc_change_of_variable}, we find that
\begin{align*}
  \int \density_{\hmcstate_0}(x) \mapG(x, \hmcstate_0) dx = \int
  \densityNormal(x) xdx = 0.
\end{align*}
Furthermore, we also have
\begin{align*}
  \int_{x \in \statespace} \density_{\hmcstate_0}(x) \vecnorm{\mapG(x,
    \hmcstate_0)}{2} dx &= \int_{x \in \statespace} \densityNormal(x)
  \vecnorm{x}{2}dx \\ &\stackrel{(i)}{\leq} \brackets{\parenth{\int_{x
        \in \statespace} \densityNormal (x) \vecnorm{x}
      {2}^2dx}{\parenth{\int_{x \in \statespace} \densityNormal(x)
        dx}}}^{1/2} = \sqrt{\dims},
\end{align*}
where step~(i) follows from Cauchy-Schwarz's inequality.  Combining
the inequalities~\eqref{eq:T1_first_part},
\eqref{eq:T1_diff_G_residual_bound}~and~\eqref{eq:T1_second_part_decompose}
together, we obtain the following bound on term $T_1$:
\begin{align}
  \label{eq:T1_final_bound}
  \abss{T_1} &= \abss{\int \density_{\hmcstate_0}(x) \brackets{-
      \frac{1}{2}\vecnorm{\mapG(x, \hmcstate_0)}{2}^2 +\frac{1}{2}
      \vecnorm{\mapG(x, \hmcstatey_0)}{2}^2} dx} \notag \\ & \leq
  \frac{1}{2}\abss{\int \density_{\hmcstate_0} (x) \vecnorm{\mapG(x,
      \hmcstate_0) - \mapG(x, \hmcstatey_0)}{2}^2 dx}
  \notag\\ &\qquad\qquad\qquad+ \abss{\int \density_{\hmcstate_0} (x)
    \vecnorm{\mapG (x, \hmcstate_0) - \mapG(x, \hmcstatey_0) -
      C(\hmcstate_0, \hmcstatey_0)}{2}\vecnorm{\mapG(x,
      \hmcstate_0)}{2} dx}\notag \\ &\leq
  \frac{8}{9}\frac{\vecnorm{\hmcstate_0 -
      \hmcstatey_0}{2}^2}{\internsteps^2\step^2} +
  \frac{3}{2}\sqrt{\dims} \internsteps \step \vecnorm{\hmcstate_0 -
    \hmcstatey_0}{2},
\end{align}
which yields the claimed bound on $T_1$.\\

We now prove our earlier claim~\eqref{eq:T1_diff_G_residual_bound}.

\paragraph{Proof of claim~\eqref{eq:T1_diff_G_residual_bound}:}

\label{par:proof_of_claim_eq:t1_diff_g_residual_bound_}
For any pair of states $\hmcstate_0$ and $\hmcstatey_0$, invoking the
definition~\eqref{eq:def_backward_mapping_G} of the map $\mapG(x,
\cdot)$, we obtain the following implicit equations:
\begin{align*}
  x &= \hmcstate_0 + \internsteps\step \mapG(x, \hmcstate_0) -
  \internsteps \frac{\step^2}{2} \gradf(\hmcstate_0) - \step^2
  \sum_{j=1}^{\internsteps-1} (\internsteps-j)
  \gradf(\mapFn{j}(\mapG(x, \hmcstate_0), \hmcstate_0)),
  \quad\text{and}\\ x &= \hmcstatey_0 + \internsteps\step \mapG(x,
  \hmcstatey_0) - \internsteps\frac{\step^2}{2} \gradf(\hmcstatey_0) -
  \step^2 \sum_{j=1}^{\internsteps-1} (\internsteps-j)
  \gradf(\mapFn{j}(\mapG(x, \hmcstatey_0), \hmcstatey_0)).
\end{align*}
Taking the difference between the two equations above, we obtain
\begin{align*}
  \mapG(x, \hmcstate_0) - \mapG(x, \hmcstatey_0) &- \frac{\hmcstate_0
    - \hmcstatey_0}{\internsteps\step}
  -\frac{\step}{2}\parenth{\gradf(\hmcstate_0) -
    \gradf(\hmcstatey_0)}\\ &= \frac{\step^2}{\internsteps\step}
  \sum_{k=1}^{\internsteps-1} (\internsteps-j)
  \parenth{\gradf(\mapFn{k}(\mapG(x, \hmcstate_0), \hmcstate_0)) -
    \gradf(\mapFn{k}(\mapG(x, \hmcstatey_0), \hmcstatey_0))}.
\end{align*}
Applying $\smoothness$-smoothness of $f$ along with the
bound~\eqref{eq:fk_y_bound} from Lemma~\ref{lem:jacob_G_y}, we find
that
\begin{align*}
  \vecnorm{\gradf(\mapFn{k}(\mapG(x, \hmcstate_0), \hmcstate_0)) -
    \gradf(\mapFn{k}(\mapG(x, \hmcstatey_0), \hmcstatey_0))}{2} &\leq
  \smoothness\matsnorm{\frac{\partial\mapFk(\mapG(x, y), y)}{\partial
      y}} {2} \vecnorm{\hmcstate_0-\hmcstatey_0}{2} \\ &\leq
  3\smoothness \vecnorm{\hmcstate_0-\hmcstatey_0}{2}.
\end{align*}
Putting the pieces together, we find that
\begin{align*}
  \vecnorm{\mapG(x, \hmcstate_0) - \mapG(x, \hmcstatey_0) -
    \frac{\hmcstate_0 - \hmcstatey_0}{\internsteps\step}
    -\frac{1}{2}\parenth{\gradf(\hmcstate_0) -
      \gradf(\hmcstatey_0)}}{2} \leq
  \frac{3\internsteps\step\smoothness}{2}
  \vecnorm{\hmcstate_0-\hmcstatey_0} {2},
\end{align*}
which yields the claim~\eqref{eq:T1_diff_G_residual_bound}.


\subsubsection{Proof of claim~\texorpdfstring{\eqref{eq:transition_proposal_overlap}}{transition-proposal-overlap} in Lemma~\ref{lem:transition_closeness}}
\label{sub:proof_of_claim_eq:transition_proposal_overlap}

We now bound the distance between the one-step proposal distribution
$\proposal_x$ at point $x$ and the one-step transition distribution
$\transitionbflazy_x$ at $x$ obtained after performing the
accept-reject step (and no lazy step).  Using
equation~\eqref{eq:hmcnoise_integrate_expression_K}, we define the
forward mapping $\mapFP$ for the variable $\hmcnoise_\internsteps$ as
follows
\begin{align}
  \notag \hmcnoise_\internsteps = \mapFP(\hmcnoise_0, \hmcstate_0)
  \defn \hmcnoise_0 - \frac{\step}{2}\gradf(\hmcstate_0) -
  \step\sum_{j=1}^{\internsteps-1}\gradf(\hmcstate_j) -
  \frac{\step}{2}\gradf(\hmcstate_\internsteps).
\end{align}
Consequently, the probability of staying at $x$ is given by
\begin{align}
\notag
  \transitionbflazy_{x}(\braces{x}) = 1 - \int_{\statespace} \min\braces{1,
  \frac{\exp(-\hamiltonian(\mapFP(z, x), \mapF(z, x)))}{\exp(-\hamiltonian(z, x))}} \densityNormal_{x}(z)dz,
\end{align}
where the Hamiltonian~$\hamiltonian(q, p) = \targetf(q) + \frac{1}{2}
\vecnorm{p}{2}^2$ was defined in equation~\eqref{eq:hamiltonian}.  As
a result, the TV-distance between the proposal and transition
distribution is given by
\begin{align}
  \label{eq:transition_proposal_overlap_explicit_form}
  \tvdist{\proposal_x}{\transitionbflazy_x} &= 1 - \int_{\statespace}
  \min\braces{1, \frac{\exp(-\hamiltonian(\mapFP(z, x), \mapF(z,
      x)))}{\exp(-\hamiltonian(z, x))}} \densityNormal_{x}(z)dz \notag
  \\ & = 1 - \Exs_{z \sim \Normal(0, \Ind_\dims)}
  \brackets{\min\braces{1, \frac{\exp(-\hamiltonian(\mapFP(z, x),
        \mapF(z, x)))}{\exp(-\hamiltonian(z, x))}}}.
\end{align}
An application of Markov's inequality yields that
\begin{align}
  &\Exs_{z \sim \Normal(0, \Ind_\dims)} \brackets{\min\braces{1,
      \frac{\exp (-\hamiltonian(\mapFP(z, x), \mapF(z,
        x)))}{\exp(-\hamiltonian(z, x))}}} \notag\\ &\qquad\geq \alpha
  \Prob_{z \sim \Normal(0, \Ind_\dims)} \brackets{\frac{\exp
      (-\hamiltonian(\mapFP(z, x), \mapF(z,
      x)))}{\exp(-\hamiltonian(z, x))} \geq
    \alpha},\label{eq:Markov_ineq_in_accept_reject}
\end{align}
for any $\alpha\in (0, 1]$.  Thus, to bound the distance
  $\tvdist{\proposal_x}{\transitionbflazy_x}$, it suffices to derive a
  high probability lower bound on the ratio
  ${\exp(-\hamiltonian(\mapFP(z, x), \mapF(z,
    x)))}/{\exp(-\hamiltonian (z, x))}$ when $z \sim \Normal(0,
  \Ind_\dims)$.

We now derive a lower bound on the following quantity:
\begin{align*}
   \exp\parenth{-\targetf(\mapF(\hmcnoise_0, \hmcstate_0)) +
     \targetf(\hmcstate_0) - \frac{1}{2}\vecnorm{\mapFP(\hmcnoise_0,
       \hmcstate_0)}{2}^2 + \frac{1}{2}\vecnorm{\hmcnoise_0}{2}^2},
   \quad\mbox{when $\hmcnoise_0 \sim \Normal(0, \Ind_\dims)$.}
\end{align*}
We derive the bounds on the two terms $-\targetf(\mapF(\hmcnoise_0,
\hmcstate_0)) + \targetf(\hmcstate_0) $ and
$\vecnorm{\mapFP(\hmcnoise_0, \hmcstate_0)}{2}^2$ separately.

Observe that
\begin{align*}
  \targetf(\mapF(\hmcnoise_0, \hmcstate_0)) - \targetf(\hmcstate_0) =
  \sum_{j=0}^{\internsteps-1}
  \brackets{\targetf(\mapFn{j+1}(\hmcnoise_0, \hmcstate_0)) -
    \targetf(\mapFn{j}(\hmcnoise_0, \hmcstate_0))}.
\end{align*}
The intuition is that it is better to apply Taylor expansion on closer
points.  Applying the third order Taylor expansion and using the
smoothness assumptions~\eqref{eq:assumption_smoothness} and
\eqref{eq:assumption_hessianLip} for the function $\targetf$, we
obtain
\begin{align*}
  \targetf(x) - \targetf(y) \leq \frac{(x - y)\tp}{2}
  \parenth{\gradf(x) + \gradf(y)} + \hessianLip\vecnorm{x - y}{2}^3.
\end{align*}
For the indices $j \in \braces{0, \ldots, \internsteps - 1}$,
using $\mapFn{j}$  as the shorthand for $\mapFn{j}(\hmcnoise_0, \hmcstate_0)$,
we find that
\begin{align}
  \label{eq:f_qj_diff_bound}
  \targetf(\mapFn{j+1}) - \targetf(\mapFn{j}) &\leq \frac{(\mapFn{j+1}
    - \mapFn{j})\tp}{2} \parenth{\gradf(\mapFn{j+1}) +
    \gradf(\mapFn{j})} + \hessianLip\vecnorm{\mapFn{j+1} -
    \mapFn{j}}{2}^3 \notag \\ & = \frac{1}{2}\step
  \hmcnoise_0\tp\parenth{\gradf(\mapFn{j+1}) + \gradf (\mapFn{j})}
  \notag\\ &\quad-\frac{\step^2}{2}
  \bigg[{\frac{1}{2}\gradf(\hmcnoise_0) + \sum_{k=1}^j
      \gradf(\mapFn{k})}\bigg]\tp \parenth{\gradf(\mapFn{j+1}) +
    \gradf(\mapFn{j})} + \hessianLip\vecnorm{\mapFn{j+1} -
    \mapFn{j}}{2}^3,
\end{align}
where the last equality follows by definition~\eqref{eq:def_forward_mapping_F}
of the operator $\mapFn{j}$.

Now to bound the term $\mapFP(\hmcnoise_0, \hmcstate_0)$, we observe that
\begin{align}
  \frac{\vecnorm{\mapFP(\hmcnoise_0, \hmcstate_0)}{2}^2}{2} &=
  \frac{\vecnorm{\hmcnoise_0 - \frac{\step}{2}\gradf(\hmcstate_0) -
      \step\sum_{j=1}^{\internsteps-1}\gradf(\mapFn{j}) -
      \frac{\step}{2}\gradf(\mapFn{\internsteps})}{2}^2}{2} \notag
  \\ &= \frac{\vecnorm{\hmcnoise_0}{2}^2}{2} - \step \hmcnoise_0\tp
  \bigg({ \frac{1}{2}\gradf(\hmcstate_0) +
    \sum_{j=1}^{\internsteps-1}\gradf(\mapFn {j}) +
    \frac{1}{2}\gradf(\mapFn{\internsteps})}\bigg)
  \notag\\ &\qquad\qquad\qquad+ \frac{\step^2}{2}
  \big\Vert{\frac{1}{2}\gradf(\hmcstate_0) +
    \sum_{j=1}^{\internsteps-1}\gradf(\mapFn{j}) +
    \frac{1}{2}\gradf(\mapFn {\internsteps})}\big\Vert_2^2.
  \label{eq:pk_expansion}
\end{align}

Putting the equations~\eqref{eq:f_qj_diff_bound} and~\eqref{eq:pk_expansion}
together leads to cancellation of many gradient terms and we
obtain
\begin{align}
  \label{eq:accept_rate_lower_bound}
  &-\targetf(\mapF(\hmcnoise_0, \hmcstate_0)) + \targetf(\hmcstate_0)
  - \frac{1}{2}\vecnorm{\mapFP(\hmcnoise_0, \hmcstate_0)}{2}^2 +
  \frac{1}{2}\vecnorm{\hmcnoise_0}{2}^2 \notag \\ &\quad \geq
  \frac{\step^2}{8}\parenth{\gradf(\hmcstate_0) - \gradf(\mapFn
    {\internsteps})}\tp \parenth{\gradf(\hmcstate_0) +
    \gradf(\mapFn{\internsteps})} -
  \hessianLip\sum_{j=0}^{\internsteps-1}\vecnorm{\mapFn{j+1} -
    \mapFn{j}}{2}^3 \notag \\ &\quad \geq
  -\frac{\step^2\smoothness}{4}
  \vecnorm{\hmcstate_0-\mapF(\hmcnoise_0, \hmcstate_0)}{2}
  \vecnorm{\gradf(\hmcstate_0)}{2} - \frac{\step^2\smoothness^2}{2}
  \vecnorm{\hmcstate_0-\mapF(\hmcnoise_0, \hmcstate_0)}{2}^2 -
  \hessianLip \sum_{j=0}^{\internsteps - 1}\vecnorm{\mapFn{j+1} -
    \mapFn{j}}{2}^3
\end{align}
The last inequality uses the smoothness condition~\eqref{eq:assumption_smoothness}
for the function $\targetf$.
Plugging the bounds~\eqref{eq:q_j_recursion_bound} and \eqref{eq:q_j_diff_recursion_bound}
in equation~\eqref{eq:accept_rate_lower_bound}, we obtain a lower bound
that only depends on $\vecnorm{\hmcnoise_0}{2}$ and $\vecnorm{\gradf(\hmcstate_0)}{2}$:
\begin{align}
  \label{eq:accept_rate_lower_bound_only_p_0}
  \text{RHS of }\eqref{eq:accept_rate_lower_bound} \geq -
  2\internsteps^2\step^4\smoothness^2 \vecnorm{\hmcnoise_0}{2}^2 -
  2\internsteps\step^3\smoothness
  \vecnorm{\hmcnoise_0}{2}\vecnorm{\gradf(\hmcstate_0)}{2} - 2
  \internsteps^2 \step^4 \smoothness\vecnorm{\gradf(\hmcstate_0)}{2}^2
  \notag\\ - \hessianLip\parenth{32\internsteps\step^3
    \vecnorm{\hmcnoise_0}{2}^3 + 8
    \internsteps^{4}\step^6\vecnorm{\gradf(\hmcstate_0)}{2}^3}.
\end{align}
According to assumption~\ref{itm:assumptionA}, we have bounded
gradient in the convex set $\convset$. For any $x \in \convset$, we
have $\vecnorm{\gradf(x)}{2} \leq \gradbound$.  Standard Chi-squared
tail bounds imply that
\begin{align}
\label{eq:chi_square_p0}
    \Prob\brackets{\vecnorm{\hmcnoise_0}{2}^2 \leq \dims \alpha_1} \geq 1 - \frac{1}{16},
    \quad\mbox{for $\alpha_1 = 1+ 2 \sqrt{\log(16)} + 2\log(16)$}.
\end{align}
Plugging the gradient bound and the bound~\eqref{eq:chi_square_p0}
into equation~\eqref{eq:accept_rate_lower_bound_only_p_0}, we conclude
that there exists an absolute constant $c \leq 2000$ such that for
$\step^2$ satisfying equation~\eqref{eq:step_condition_thm_proof},
namely
\begin{align*}
    \step^2 & \leq \frac{1}{c \smoothness}
    \min\braces{\frac{1}{\internsteps^2},
      \frac{1}{\internsteps\dims^{\frac{1}{2}}},
      \frac{1}{\internsteps^{\frac{2}{3}}\dims^{\frac{1}{3}}\parenth{\frac{\gradbound^2}{\smoothness}}^{\frac{1}{3}}},
      \frac{1}{\internsteps\frac{\gradbound}{\smoothness^{\frac{1}{2}}}},
      \frac{1}{\internsteps^{\frac{2}{3}}\dims}\frac{\smoothness}{\hessianLip^
        {\frac{2}{3}}},
      \frac{1}{\internsteps^{\frac{4}{3}}\frac{\gradbound}{\smoothness^{\frac{1}{2}}}}\parenth{\frac{\smoothness}{\hessianLip^{\frac{2}{3}}}}^{\frac{1}{2}}},
\end{align*}
we have
\begin{align*}
  \Prob\brackets{-\targetf(\mapF(\hmcnoise_0, \hmcstate_0)) +
    \targetf(\hmcstate_0) - \frac{1}{2}\vecnorm{\mapFP(\hmcnoise_0,
      \hmcstate_0)}{2}^2 + \frac{1}{2}\vecnorm{\hmcnoise_0}{2}^2 \geq
    - 1/16} \geq 1-\frac{1}{16}.
\end{align*}
Plugging this bound in the
inequality~\eqref{eq:Markov_ineq_in_accept_reject} yields that
\begin{align*}
  \Exs_{z \sim \Normal(0, \Ind_\dims)} \brackets{\min\braces{1, \frac{\exp
  (-\hamiltonian(\mapFP(z, x), \mapF(z, x)))}{\exp(-\hamiltonian(z, x))}}} \geq 1 - \frac{1}{8},
\end{align*}
which when plugged in
equation~\eqref{eq:transition_proposal_overlap_explicit_form} implies
that $\tvdist{\proposal_x}{\transitionbflazy_x} \leq 1/8$ for any $x
\in \truncballres$, as claimed. The proof is now complete.


\section{Discussion}
\label{sec:discussion}

In this paper, we derived non-asymptotic bounds on mixing time of
Metropolized Hamiltonian Monte Carlo for log-concave
distributions. By choosing appropriate step-size
and number of leapfrog steps, we obtain mixing-time bounds for HMC
that are smaller than the best known mixing-time bounds for MALA. This
improvement can be seen as the benefit of using multi-step gradients
in HMC. An interesting open problem is to determine whether
our HMC mixing-time bounds are tight for log-concave sampling under the
assumptions made in the paper.

Even though, we focused on the problem of sampling only from strongly
and weakly log-concave distribution, our
Theorem~\ref{thm:hmc_mixing_general} can be applied to general distributions
including nearly log-concave distributions as mentioned in
Appendix~\ref{ssub:hmc_mixing_time_bounds_for_nearly_logconcave_target}.
It would be interesting to determine the explicit expressions for mixing-time
of HMC for more general target distributions. The other main contribution
of our paper is to
improve the warmness dependency in mixing rates of Metropolized
algorithms that are proved previously such as MRW and
MALA~\citep{dwivedi2018log}. Our techniques are inspired by those
used to improve warmness dependency in the literature of
discrete-state Markov chains. It is an interesting future direction
to determine if this warmness dependency can be further improved to prove
a convergence sub-linear in $\dims$ for HMC with generic initialization even for small condition number
$\condition$.

\subsection*{Acknowledgements}
We would like to thank Wenlong Mou for his insights and helpful discussions and Pierre Monmarch\'e for pointing out an error in the previous revision.
We would also like to thank the anonymous reviewers for their helpful comments.
This work was supported by Office of Naval Research grant
DOD ONR-N00014-18-1-2640, and National Science Foundation Grant
NSF-DMS-1612948 to MJW and by ARO W911NF1710005, NSF-DMS-1613002, NSF-IIS-174134 and the Center for Science of Information (CSoI), US NSF Science and Technology Center, under grant agreement CCF-0939370 to BY.

\appendix


\input{appendix}


\bibliography{hmc_bib}

\end{document}

%% file: macros.tex
\DeclareSymbolFont{extraup}{U}{zavm}{m}{n}
\DeclareMathSymbol{\vardiamond}{\mathalpha}{extraup}{87}

\newcommand{\statespace}{\ensuremath{\mathcal{X}}}
\newcommand{\density}{\rho}
\newcommand{\proposal}{\mathcal{P}}
\newcommand{\transition}{\mathcal{T}}
\newcommand{\transitionbflazy}{\mathcal{T}^{\text{before-lazy}}}
\newcommand{\transkernel}{\theta}
\newcommand{\transprob}{\Theta}
\newcommand{\target}{\ensuremath{\Pi^*}}
\newcommand{\targetdensity}{\ensuremath{\pi^*}}
\newcommand{\initial}{\ensuremath{\mu_0}}
\newcommand{\initialstar}{\ensuremath{\mu_\dagger}}
\newcommand{\Ltpi}{\ensuremath{L_2(\targetdensity)}}

\newcommand{\targetf}{f}
\newcommand{\gradf}{{\nabla \targetf}}
\newcommand{\hessf}{{\nabla^2 \targetf}}
\newcommand{\thirdf}{{\nabla^3 \targetf}}
\newcommand{\hamiltonian}{\mathcal{H}}
\newcommand{\xstar}{\ensuremath{x^\star}}

\newcommand{\dirac}{\mathbf{\delta}}

\newcommand{\internsteps}{K}
\newcommand{\step}{\eta}
\newcommand{\nbsteps}{n}
\newcommand{\warmparam}{\beta}
\newcommand{\errorparam}{\ensuremath{\epsilon}}
\newcommand{\lazyparam}{\zeta}

\newcommand{\scparam}{m}
\newcommand{\smoothness}{L}
\newcommand{\condition}{\kappa}
\newcommand{\hessianLip}{L_{\text{H}}}
\newcommand{\gradbound}{M}
\newcommand{\isoconst}{\psi}
\newcommand{\logisopower}{\mathfrak{a}}
\newcommand{\hessianLipRatio}{\vartheta}

\newcommand{\lp}{\mathfrak{p}}

\newcommand{\conductanceprofile}{\Phi_\convset}
\newcommand{\truncconductanceprofile}{\tilde{\Phi}_\convset}

\newcommand{\flow}{\phi}

\newcommand{\convset}{\ensuremath{\Omega}}
\newcommand{\res}{s}
\newcommand{\truncballres}{\mathcal{R}_\res}
\newcommand{\radius}{r}
\newcommand{\dirichlet}{\mathcal{E}}

\newcommand{\respectralgap}{\lambda}
\newcommand{\spectralprofile}{\Lambda}
\newcommand{\denratio}{h}

\newcommand{\varJ}{\ensuremath{J}}
\newcommand{\varJt}{\tilde{J}}

\newcommand{\Exspi}{\Exs_{\targetdensity}}

\newcommand{\Varpi}{\Var_{\targetdensity}}

\newcommand{\indMaxInters}{\mathcal{I}}

\newcommand{\tvoverlap}{\omega}

\newcommand{\discreteD}{\ensuremath{D}}
\newcommand{\indicator}{\mathbf{1}}

\newcommand{\locnewdensity}{\rho}

\newcommand{\rdefn}{\ensuremath{=:}}
\newcommand{\defn}{\ensuremath{: \, =}}
\newcommand{\TMIX}{\ensuremath{\tau}}
\newcommand{\Tmix}{\TMIX}
\newcommand{\smalltv}{\ensuremath{\operatorname{TV}}}

\newcommand{\MYPSI}{\ensuremath{\Psi}}

\newcommand{\hmc}{\ensuremath{\text{HMC-}(\internsteps,\step)}}

\newcommand{\opnorm}[1]{\ensuremath{\matsnorm{#1}{\mbox{\tiny{op}}}}}

\newcommand{\stepwarm}{\ensuremath{\step_{\mbox{\tiny{warm}}}}}
\newcommand{\stepfeasible}{\ensuremath{\step_{\mbox{\tiny{feas}}}}}

\newcommand{\median}{\nu}

\newcommand{\etal}{{et al.}}
\newcommand{\tagmala}{\text{\tiny MALA}}
\newcommand{\taghmc}{\text{\tiny HMC}}
\newcommand{\tagmrw}{\text{\tiny MRW}}

\newcommand{\hmcstate}{q}
\newcommand{\hmcstatey}{{\tilde{\hmcstate}}}
\newcommand{\hmcnoise}{p}
\newcommand{\hmcnoisey}{{\tilde{\hmcnoise}}}

\newcommand{\tensor}{\mathcal{T}}

\newcommand{\mapF}{F}
\newcommand{\mapFn}[1]{\mapF_{#1}}
\newcommand{\mapFk}{\mapFn{k}}
\newcommand{\mapG}{G}
\newcommand{\mapFP}{E}

\newcommand{\hmcmass}{M}
\newcommand{\targetg}{h}
\newcommand{\jacob}{\mathbf{J}}

\long\def\comment#1{}
\definecolor{battleshipgrey}{rgb}{0.52, 0.52, 0.51}
\definecolor{darkgray}{rgb}{0.66, 0.66, 0.66}
\definecolor{darkgreen}{rgb}{0.0, 0.2, 0.13}
\definecolor{darkspringgreen}{rgb}{0.09, 0.45, 0.27}
\definecolor{dukeblue}{rgb}{0.0, 0.0, 0.61}
\definecolor{olivedrab7}{rgb}{0.24, 0.2, 0.12}
\definecolor{darkblue}{rgb}{0.0, 0.0, 0.55}
\definecolor{darkscarlet}{rgb}{0.34, 0.01, 0.1}
\definecolor{candyapplered}{rgb}{1.0, 0.03, 0.0}
\definecolor{ao(english)}{rgb}{0.0, 0.5, 0.0}
\definecolor{applegreen}{rgb}{0.55, 0.71, 0.0}
\definecolor{orange}{rgb}{1.0, 0.65, 0.0}

\newcommand{\red}[1]{\textcolor{red}{#1}}

\newcommand{\orange}[1]{\textcolor{orange}{#1}}

\newcommand{\dims}{\ensuremath{d}}
\newcommand{\real}{\ensuremath{\mathbb{R}}}
\newcommand{\naturalnum}{\ensuremath{\mathbb{N}}}
\newcommand{\Ind}{\ensuremath{\mathbb{I}}}
\newcommand{\borel}{\ensuremath{\mathcal{B}}}

\newcommand{\ball}{\ensuremath{\mathbb{B}}}
\newcommand{\Exs}{\ensuremath{{\mathbb{E}}}}
\newcommand{\Prob}{\ensuremath{{\mathbb{P}}}}

\newcommand{\Normal}{\ensuremath{\mathcal{N}}}

\DeclareMathOperator{\Var}{Var}

\newcommand{\densityNormal}{\ensuremath{\varphi}}

\DeclarePairedDelimiterX{\infdivx}[2]{(}{)}{%
  #1\;\delimsize\|\;#2%
}
\newcommand{\kldiv}{\text{KL}\infdivx}
\newcommand{\tvdist}[2]{\ensuremath{ d_{\text{\tiny{TV}}}\parenth{#1, #2} }}
\newcommand{\Ell}{\mathcal{L}}

\newcommand{\Ot}{\widetilde{O}}

\DeclareMathOperator{\trace}{\text{trace}}

\newcommand{\brackets}[1]{\left[ #1 \right]}
\newcommand{\parenth}[1]{\left( #1 \right)}

\newcommand{\braces}[1]{\left\{ #1 \right \}}
\newcommand{\abss}[1]{\left| #1 \right |}

\newcommand{\ceils}[1]{\left\lceil #1 \right \rceil}

\newcommand{\tp}{^\top}

\newcommand{\matsnorm}[2]{\left|\!\left|\!\left| #1 \right|\!\right|\!\right|_{{#2}}}
\newcommand{\vecnorm}[2]{\left\| #1\right\|_{#2}}

%% file: appendix.tex

\etoctocstyle{1}{Appendix}
\etocdepthtag.toc{mtappendix}
\etocsettagdepth{mtchapter}{none}
\etocsettagdepth{mtappendix}{subsection}
\etocsettagdepth{mtappendix}{subsubsection}
\tableofcontents


\section{Proof of Lemmas~\ref{prop:mixing_bound_using_conductanceprofile},
\ref{prop:conductanceprofile_via_overlaps} and \ref{lem:transition_closeness}}
\label{sec:proof_of_lemmas_1_2}

In this appendix, we collect the proofs of
Lemmas~\ref{prop:mixing_bound_using_conductanceprofile},
and~\ref{prop:conductanceprofile_via_overlaps}, as previously stated
in
Section~\ref{sub:mixing_time_of_a_markov_chain_with_conductance_profile},
that are used in proving Theorem~\ref{thm:hmc_mixing_general}.
Moreover, we provide the proof of auxiliary results related to HMC
proposal that were used in the proof of
Lemma~\ref{lem:transition_closeness}.

\subsection{Proof of Lemma~\ref{prop:mixing_bound_using_conductanceprofile}}
\label{sec:proof_of_lemma_2}

In order to prove
Lemma~\ref{prop:mixing_bound_using_conductanceprofile}, we begin by
adapting the spectral profile technique~\citep{goel2006mixing} to the
continuous state setting, and next we relate conductance profile with
the spectral profile.

First, we briefly recall the notation from
Section~\ref{sub:markov_chain_terminology}.  Let
$\transprob: \statespace \times
\borel(\statespace) \rightarrow \real_+$ denote the transition probability
function for the Markov chain and let $\transition$ be the
corresponding transition operator, which maps a probability measure to
another according to the transition probability $\transprob$.  Note
that for a Markov chain satisfying the smooth chain
assumption~\eqref{eq:smooth_chain_assumption}, if the distribution
$\mu$ admits a density then the distribution $\transition(\mu)$ would
also admits a density. We use $\transition_x$ as the shorthand for
$\transition(\dirac_x)$, the transition distribution of the Markov
chain at $x$.

Let $\Ltpi$ be the space of square integrable functions under function
$\targetdensity$.  The
\emph{Dirichlet form}
$\dirichlet: \Ltpi \times \Ltpi \rightarrow \real$ associated with the
transition probability $\transprob$ is given by
\begin{align}
  \label{eq:def_dirichlet_form} \dirichlet(g, h)
  = \frac{1}{2} \int_{(x, y) \in \statespace^2} \parenth{g (x) -
  h(y)}^2 \transprob(x, dy) \targetdensity(x) dx.
\end{align}
The expectation $\Exspi: \Ltpi \rightarrow \real$ and the variance
$\Varpi: \Ltpi \rightarrow \real$ with respect to the density
$\targetdensity$ are given by
\begin{subequations}
\begin{align}
  \label{eq:def_Exs_Var_target} \Exspi(g) = \int_{x \in \statespace}
  g(x) \targetdensity(x) dx \quad\text{and}\quad \Varpi(g)
  = \int_{x\in \statespace} \parenth{g(x)
  - \Exspi(g)}^2 \targetdensity(x) dx.
\end{align}
\end{subequations}
Furthermore, for a pair of measurable sets
  $(S, \convset) \subset \statespace^2$, the
  $\convset$-\emph{restricted spectral gap for the set $S$} is defined
  as
\begin{align}
  \label{eq:def_respectralgap} \respectralgap_{\convset}(S)
  \defn \inf_{g \in c_{0, \convset}^+(S)} \frac{\dirichlet (g,
  g)}{\Varpi(g \cdot \indicator_\convset)},
\end{align}
where
\begin{align}
  \label{eq:define_c0}
  c_{0, \convset}^+(S)
  \defn \braces{g \in \Ltpi \mid \text{supp}(g) \subset S,
  g \geq 0, g \cdot \indicator_\convset \neq \text{constant}, \Exs[g^2 \cdot \indicator_{\convset^c}] \leq \Exs[g^2 \cdot \indicator_\convset]}.
\end{align}
Finally, the $\convset$-\emph{restricted spectral profile}
$ \spectralprofile_{\convset}$ is defined as
\begin{align}
  \label{eq:def_spectralprofile} \spectralprofile_{\convset}(v)
  \defn \inf_{\target(S \cap \convset) \in [0,
  v]} \respectralgap_{\convset}(S), \quad\text{ for all
  } v \in \big[0, \infty).
\end{align}
Note that we restrict the spectral profile to the set $\convset$.
Taking $\convset$ to be $\statespace$, our definition agrees with the
standard definition definitions of the restricted spectral gap and
spectral profile in the paper~\citep{goel2006mixing} for finite state
space Markov chains to continuous state space Markov chains.

\noindent We are now ready to state a mixing time bound using spectral profile.
\begin{lemma}
  \label{lem:mixing_bound_using_spectralprofile} Consider a reversible
  irreducible $\lazyparam$-lazy Markov chain with stationary
  distribution $\target$ satifying the smooth chain
  assumption~\eqref{eq:smooth_chain_assumption}. Given a
  $\warmparam$-warm start $\initial$, an error tolerance
  $\errorparam \in (0, 1)$ and a set $\convset \subset \statespace$
  with $\target(\convset) \geq 1-\frac{\errorparam^2}{3\warmparam^2}$,
  the $L_2$-mixing time is bounded as
\begin{align}
\label{eq:mixing_bound_using_spectralprofile}
\TMIX_{2}(\errorparam; \initial) \leq \ceils{\int_{4/\warmparam}^{8/\errorparam^2} \frac{dv}{\lazyparam \cdot
  v\spectralprofile_{\convset} (v)}},
\end{align}
where $\spectralprofile_{\convset}$ denotes the $\convset$-restricted
spectral profile~\eqref{eq:def_spectralprofile} of the chain.
\end{lemma}
\noindent See Appendix~\ref{sub:proof_of_lemma_lem:mixing_bound_using_spectralprofile}
for the proof.\\

In the next lemma, we state the relationship between the
$\convset$-restricted spectral profile~\eqref{eq:def_spectralprofile}
of the Markov chain to its $\convset$-restricted conductance
profile~\eqref{eq:def_conductanceprofile}.
\begin{lemma}
\label{lem:relate_spectralprofile_and_conductanceprofile}
For a Markov chain with state space $\statespace$ and stationary
distribution $\target$, given any measurable set
$\convset\subset\statespace$, its $\convset$-restricted spectral
profile~ \eqref{eq:def_spectralprofile} and $\convset$-restricted
conductance profile~\eqref{eq:def_conductanceprofile} are related
as
\begin{align}
\label{eq:relate_spectralprofile_and_conductanceprofile}
\spectralprofile_{\convset}(v) \geq \begin{cases}\displaystyle \frac{\conductanceprofile^2(v)}{4} \quad
&\text{for all } v \in \brackets{0, \frac{\target
(\convset)}{2}}\\ \displaystyle \frac{\conductanceprofile^2(\target(\convset)/2)}{8} \quad
&\text{for all }
v \in \big(\frac{\target(\convset)}{2}, \infty).  \end{cases} \end{align}
\end{lemma}
\noindent See Appendix~\ref{sub:proof_of_lemma_lem:relate_spectralprofile_and_conductanceprofile}
for the proof.\\

\noindent Lemma~\ref{prop:mixing_bound_using_conductanceprofile} now follows
from Lemmas~\ref{lem:mixing_bound_using_spectralprofile}
and~\ref{lem:relate_spectralprofile_and_conductanceprofile} as well as
the definition~\eqref{eq:def_truncatedprofile} of
$\truncconductanceprofile$.


\subsubsection{Proof of Lemma~\ref{lem:mixing_bound_using_spectralprofile}}
\label{sub:proof_of_lemma_lem:mixing_bound_using_spectralprofile}

We need the following lemma, proved in for the case of finite state
Markov chains in~\cite{goel2006mixing}, which lower bounds
the Dirichlet form in terms of the spectral profile.
\begin{lemma}
  \label{lem:lower_bound_Dirichlet_w_spectralprofile}
For any measurable set $\convset\subset\statespace$, any function $g: \statespace \rightarrow \real_+$ such that $g \in \Ltpi$, $g \cdot \indicator_\convset$ is not constant and $\Exs[g^2 \cdot \indicator_{\convset}] \geq 2 \Exs[g^2 \cdot \indicator_{\convset^c}] $, we have
\begin{align}
\label{eq:lower_bound_Dirichlet_w_spectralprofile}
\frac{\dirichlet(g,
  g)}{\Varpi(g \cdot \indicator_\convset)} \geq \frac{1}{2} \spectralprofile_
  {\convset}\parenth{\frac{4\parenth{\Exspi(g \cdot \indicator_\convset)}^2}{\Varpi(g \cdot \indicator_\convset)}}.
\end{align}
\end{lemma}
\noindent The proof of Lemma~\ref{lem:lower_bound_Dirichlet_w_spectralprofile}
is a straightforward extension of Lemma 2.1 from \cite{goel2006mixing}, which deals with finite state spaces, to
the continuous state Markov chain. See the end of
Section~\ref{par:proof_of_lemma_lem:lower_bound_dirichlet_w_spectralprofile}
for the proof.


\noindent We are now equipped to prove
Lemma~\ref{lem:mixing_bound_using_spectralprofile}.

\paragraph{Proof of Lemma~\ref{lem:mixing_bound_using_spectralprofile}:}

We begin by introducing some notations.  Recall that for any Markov
chain satisfying the smooth chain
assumption~\eqref{eq:smooth_chain_assumption}, given an initial
distribution $\initial$ that admits a density, the distribution of the
chain at any step $\nbsteps$ also admits a density.  As a result, we
can define the ratio of the density of the Markov chain at the
$\nbsteps$-th iteration $\denratio_{\initial, \nbsteps}: \statespace
\rightarrow \real$ with respect to the target density $\targetdensity$ via
the following recursion
\begin{align*}
\denratio_{\initial, 0}(x)
  = \frac{\initial(x)}{\targetdensity(x)} \quad\text{and} \quad \denratio_{\initial, \nbsteps+1}(x)
  = \frac{\transition\parenth{\targetdensity\cdot \denratio_{\initial, \nbsteps}}(x)}
  {\targetdensity(x)},
\end{align*}
where we have used the notation $\transition(\mu)(x)$ to denote the
density of the distribution $\transition(\mu)$ at $x$.  Note that
\begin{align}
\label{eq:density_ratio_expectation}
  \Exspi(\denratio_{\initial, \nbsteps}) =
  1 \quad\text{and}\quad \Exspi(\denratio_{\initial, \nbsteps}\cdot \indicator_\convset) \leq
  1 \quad \text{for all } \nbsteps \geq 0,
\end{align}
where $\convset\subset\statespace$ is a measurable set.

We also define the quantity $\varJ(\nbsteps) \defn \Varpi(\denratio_{\initial,
\nbsteps})$ (we prove the existence of this variance below in Step (1)) and also $\varJt(\nbsteps) \defn \Varpi(\denratio_{\initial,
\nbsteps} \cdot \indicator_\convset)$.
Note that the $L_2$-distance between the distribution of the chain at
step $\nbsteps$ and the target distribution is given by
\begin{align*}
d_{2, \targetdensity}(\transition^\nbsteps(\initial), \target)
= \parenth{\int_{x \in \real^\dims} \parenth{\denratio_{\initial, \nbsteps}(x)
- 1}^2 \targetdensity(x) dx}^{1/2}
= \Varpi(\denratio_{\initial, \nbsteps}).
\end{align*}
Consequently, to prove the $\errorparam$-$L_2$ mixing time
bound~\eqref{eq:mixing_bound_using_spectralprofile}, it suffices to
show that for any measurable set $\convset \subset \statespace$, with
$\target(\convset) \geq 1-\frac{\errorparam^2}{3\warmparam^2}$, we
have
\begin{align}
\label{eq:mixing_time_in_terms_of_var}
  \varJ(\nbsteps) \leq \errorparam^2 \quad\text{ for
  } \nbsteps \geq \ceils{\int_{4/\warmparam}^{8/\errorparam^2} \frac{dv}{\lazyparam \cdot
  v\spectralprofile_{\convset}(v)}}
\end{align}
We now establish the claim~\eqref{eq:mixing_time_in_terms_of_var} via
a three-step argument: (1) we prove the existence of the variance
$\varJ(\nbsteps)$ for all $\nbsteps \in \naturalnum$ and relate $\varJ(\nbsteps)$ with $\varJt(\nbsteps)$. (2) then we
derive a recurrence relation for the difference
$\varJ(\nbsteps+1)-\varJ(\nbsteps)$ in terms of Dirichlet forms that
shows the $\varJ$ is a decreasing function, and (3) finally, using an
extension of the variance $\varJ$ from natural indices to real
numbers, we derive an explicit upper bound on the number of steps
taken by the chain until $\varJ$ lies below the required threshold.

\paragraph{Step (1):}
Using the reversibility~\eqref{eq:def_reversible} of the chain, we
find that
\begin{align}
\denratio_{\initial, \nbsteps+1}(x) dx
  = \frac{\int_{y \in \statespace} \transprob(y,
  dx) \denratio_{\initial, \nbsteps}(y)\targetdensity(y)dy}{\targetdensity(x)}
  & = \frac{\int_{y \in \statespace} \transprob(x,
  dy) \denratio_{\initial, \nbsteps}(y)\targetdensity(x)dx}{\targetdensity(x)} \notag\\
  & = \int_{y \in \statespace} \transprob(x,
  dy) \denratio_{\initial, \nbsteps} (y)
  dx\label{eq:equality_of_density_ratio}
\end{align}
Applying an induction argument along with the
relationship~\eqref{eq:equality_of_density_ratio} and the initial
condition $\denratio_{\initial, 0}(x) \leq \warmparam$, we obtain that
\begin{align}
  \label{eq:density_ratio_always_bounded} \denratio_{\initial, \nbsteps}(x) \leq \warmparam, \quad \text{for
  all } \nbsteps \geq 0.
\end{align}
As a result, the variances of the functions $\denratio_{\initial, 0}$
and $\denratio_{\initial, \nbsteps} \cdot \indicator_\convset$ under
the target density $\targetdensity$ are well-defined and
\begin{align}
\label{eq:var_at_n}
  \varJ(\nbsteps)
  = \int_{\statespace} \denratio^2_{\initial,\nbsteps}(x) \targetdensity(x)
  dx - 1.
\end{align}
Then we show that $\varJ$ can be upper bounded via $\varJt$ as follows
\begin{align}
\label{eq:partial_variance_lower_bound}
\varJ(\nbsteps) - \varJt(\nbsteps)
  &=\Varpi(\denratio_{\initial, \nbsteps})
  - \Varpi(\denratio_{\initial, \nbsteps} \cdot \indicator_\convset) \notag \\
  &= \int_{x \in \statespace \setminus \convset
  } \denratio^2_{\initial, \nbsteps}(x) \targetdensity(x) dx
  - \parenth{\int_{x \in \statespace} \denratio_{\initial, \nbsteps}(x) \targetdensity(x)
  dx}^2 \notag\\
  &\quad\quad\quad+ \parenth{\int_
  {x \in \convset} \denratio_{\initial, \nbsteps}(x) \targetdensity(x)
  dx}^2 \notag \\
  &\leq \warmparam^2 \parenth{1
  - \target(\convset)} \leq \frac{\errorparam^2}{3} \rdefn B,
\end{align}
where the last inequality follows from the fact that $\convset$
satisfies $\target(\convset) \geq 1-\errorparam^2/(3\warmparam^2)$. Similarly, we have
\begin{align}
  \label{eq:partial_Esquare_lower_bound}
  \Exspi (\denratio_{\initial, \nbsteps}^2) - \Exspi(\denratio_{\initial, \nbsteps}^2 \cdot \indicator_\convset) \leq B
\end{align}

\paragraph{Step (2):}
\label{par:step}
First, note the following bound on $\varJ(0)$:
\begin{align}
\label{eq:bound_on_var0}
  \varJ(0)
  = \int_{x\in\statespace}\frac{\initial(x)^2}{\targetdensity(x)} dx -
  1 \leq \warmparam \int_{x\in\statespace}\initial(x)dx -
  1 \leq \warmparam - 1.
\end{align}

Define the two step transition kernel $\transprob \circ \transprob$ as
\begin{align*}
  \transprob \circ \transprob(y, dz)
  = \int_{x \in \statespace} \transprob(y, dx) \transprob(x, dz).
\end{align*}
We have
\begin{align*}
  \varJ(\nbsteps+1)\defn\Varpi(\denratio_{\initial, \nbsteps+1})
  &= \int_
  {x \in \statespace} \denratio^2_{\initial, \nbsteps+1}(x) \targetdensity(x)
  dx - 1 \\
  &\stackrel{(i)}{=} \int_{x \in \statespace} \int_{y \in \statespace} \transprob
  (y,
  dx) \denratio_{\initial, \nbsteps}(y)\targetdensity(y)dy \int_{z \in \statespace} \transprob(x,
  dz) \denratio_{\initial, \nbsteps}(z) - 1\\ & = \int_{y,
  z \in \statespace^2} \transprob \circ \transprob(y,
  dz) \denratio_{\initial, \nbsteps}(y) \denratio_{\initial, \nbsteps}(z) \targetdensity(y)
  dy - 1,
\end{align*}
where step~(i) follows from the
relation~\eqref{eq:equality_of_density_ratio}.  Using the above
expression for $\varJ(\nbsteps+1)$ and the expression from
equation~\eqref{eq:var_at_n} for $\varJ(\nbsteps)$, we find that
\begin{align}
  \varJ(\nbsteps+1) - \varJ(\nbsteps)
  &= \int_{\statespace^2} \transprob \circ \transprob(y,
  dz) \denratio_{\initial, \nbsteps}(y) \denratio_{\initial, \nbsteps}(z) \targetdensity(y)
  dy
  -\int_{\statespace} \denratio^2_{\initial,\nbsteps}(x) \targetdensity(x)
  dx, \notag\\ &\stackrel{(a)}{=}
  - \dirichlet_{\transprob \circ \transprob}(\denratio_
  {\initial, \nbsteps}, \denratio_{\initial, \nbsteps}), \label{eq:var_h_diff}
\end{align}
where $\dirichlet_{\transprob \circ \transprob}$ is the Dirichlet form~
\eqref{eq:def_dirichlet_form} with transition probability $\transprob$ being
replaced by $\transprob \circ \transprob$. We come back to the proof
of equality~(a) at the end of this paragraph.  Assuming it as given at
the moment, we proceed further.  Since the Markov chain is
$\lazyparam$-lazy, we can relate the two Dirichlet forms
$\dirichlet_{\transprob \circ \transprob}$ and
$\dirichlet_{\transprob}$ as follows: For any $y, z \in \statespace$
such that $y \neq z$, we have
\begin{align}
  \label{eq:dirichlet_two_step_lazy_bound} \transprob \circ \transprob(y,
  dz) = \int_{x \in \statespace} \transprob(y, dx)\transprob(x, dz)
  &\geq \transprob(y, dy)\transprob(y, dz) + \transprob(y,
  dz)\transprob (z, dz)\notag\\ &\geq 2 \lazyparam \transprob(y, dz).
\end{align}
If $\Exspi (\denratio_{\initial, \nbsteps}^2 \cdot \indicator_\convset) < 2\Exspi(\denratio_{\initial, \nbsteps}^2 \cdot \indicator_{\convset^c})$, then according to Equation~\eqref{eq:partial_Esquare_lower_bound}, we have
\begin{align}
  \label{eq:varJ_without_recursion}
  \varJ(\nbsteps) \leq \Exs(\denratio_{\initial, \nbsteps}^2) \leq \Exs(\denratio_{\initial, \nbsteps}^2 \cdot \indicator_\convset) + B\leq 3 B \leq \errorparam^2,
\end{align}
and we are done. If $\denratio_{\initial, \nbsteps}^2 \cdot \indicator_\convset$ is constant, then $\Varpi(\denratio_{\initial, \nbsteps}^2 \cdot \indicator_\convset) = 0$ and
\begin{align*}
  \varJ(\nbsteps) \leq \Varpi(\denratio_{\initial, \nbsteps}^2) \leq 0 + B \leq \errorparam^2.
\end{align*}
Otherwise, we meet the assumptions of Lemma~\ref{lem:lower_bound_Dirichlet_w_spectralprofile} and we have
\begin{align}
  \label{eq:varJ_recursion} \varJ(\nbsteps + 1) - \varJ(\nbsteps) =
  -\dirichlet_{\transprob \circ \transprob}(\denratio_{\initial, \nbsteps}, \denratio_{\initial, \nbsteps})
  &\stackrel{(i)}{\leq}
  -2\lazyparam \dirichlet_{\transprob}(\denratio_{\initial, \nbsteps}, \denratio_{\initial, \nbsteps}) \notag \\
  &\stackrel{(ii)}{\leq}
  -\lazyparam \Varpi(\denratio_{\initial, \nbsteps} \cdot \indicator_\convset
  ) \spectralprofile_{\convset} \parenth{\frac{4 \brackets{\Exspi(\denratio_{\initial, \nbsteps}\cdot \indicator_\convset)}^2}
  {\Varpi(\denratio_
  {\initial, \nbsteps}\cdot \indicator_\convset)}} \notag\\
  &\stackrel{(iii)}{\leq}
  - \lazyparam \cdot \parenth{\varJ(\nbsteps)-B} \spectralprofile_{\convset}\parenth{\frac{4}{\varJ(\nbsteps)
  - B}}.
\end{align}
where step~(i) follows from
inequality~\eqref{eq:dirichlet_two_step_lazy_bound},
step~(ii) follows from
Lemma~\ref{lem:lower_bound_Dirichlet_w_spectralprofile}, and finally
step~(iii) follows from
inequality~\eqref{eq:partial_variance_lower_bound} which implies that
$\Varpi(\denratio_{\initial, \nbsteps} \cdot \indicator_\convset
) \geq \varJ(\nbsteps)-B$, and the fact that the spectral profile
$\spectralprofile_\convset$ is a non-increasing function.

\paragraph{Proof of equality~(a) in equation~\eqref{eq:var_h_diff}:}
\label{par:proof_of_equality_a}

Since the distribution $\target$ is stationary with respect to the
kernel $\transprob$, it is also stationary with respect to the two
step kernel $\transprob \circ \transprob$.  We now prove a more
general claim: For any transition kernel $K$ which has stationary
distribution $\target$ and any measurable function $h$, the Dirichlet
form $\dirichlet_{K}$, defined by replacing $\transprob$ with $K$ in
equation~\eqref{eq:def_dirichlet_form}, we have
\begin{align}
\label{eq:more_general_claim_dirichlet}
  \dirichlet_{K}(h, h) = \int_{\statespace} h^2(x)\targetdensity(x)dx
  - \int_{\statespace}\int_{\statespace} h(x) h(y) K(x,
  dy) \targetdensity (x) dx.
\end{align}
Note that invoking this claim with $K = \transprob \circ \transprob$
and $h = \denratio_{\initial, \nbsteps}$ implies step~(a) in
equation~\eqref{eq:var_h_diff}.  We now establish the
claim~\eqref{eq:more_general_claim_dirichlet}.  Expanding the square
in the definition~\eqref{eq:def_dirichlet_form}, we obtain that
\begin{align*}
  \dirichlet_{K}(h, h)
  &= \frac{1}{2} \int_{\statespace} \int_{\statespace} h^2(x) K(x,
  dy) \targetdensity (x) dx
  + \frac{1}{2} \int_{\statespace} \int_{\statespace} h^2(y) K(x,
  dy) \targetdensity (x) dx\\
  &\quad\quad\quad\quad- \int_{\statespace} \int_{\statespace} h(x)
  h(y) K(x, dy)\targetdensity(x) dx\\
  &\stackrel{(i)}{=} \frac{1}{2} \int_{\statespace}
  h^2(x) \targetdensity(x) dx +\frac{1}{2} \int_{\statespace}
  h^2(x) \targetdensity(x) dx - \int_{\statespace} \int_{\statespace}
  h(x) h(y) K(x, dy)\targetdensity(x) dx,
\end{align*}
where equality~(i) follows from the following facts: For the first
term, we use the fact that $\int_\statespace K(x, dy) = 1$ since $K$
is a transition kernel, and, for the second term we use the fact that
$\int_\statespace K(x, dy)\targetdensity(x)dx = \targetdensity(y) dy$,
since $\target$ is the stationary distribution for the kernel $K$.
The claim now follows.

\paragraph{Step (3):}

Consider the domain extension of the function $\varJ$ from
$\naturalnum$ to the set of non-negative real numbers $\real_+$ by
piecewise linear interpolation. We abuse notation and denote this
extension also by $\varJ$. The extended function $\varJ$ is continuous
and is differentiable on the set $\real_+ \backslash \naturalnum$.
Let $\nbsteps^* \in \real_+ \cup \braces{\infty}$ denote the first index
such that $\varJ(\nbsteps^*) \leq 3B$.  Since $\spectralprofile_\convset$
is non-increasing and $\varJ$ is non-increasing, we have
\begin{align}
\label{eq:varJ_differential_eq}
\varJ'(t) \leq - \lazyparam \cdot \parenth{\varJ(t) -
  B}\spectralprofile_\convset \parenth{\frac{4}{\varJ(t)-
  B}} \quad \text{for all } t \in \real_+\backslash\naturalnum \text{
  such that } t \leq \nbsteps^*.
\end{align}
Moving the $\varJ$ terms on one side and integrating for
$t \leq \nbsteps^*$, we obtain
\begin{align*}
  \int_{\varJ(0)}^{\varJ(t)} \frac{d\varJ}{\parenth{\varJ-B} \cdot \spectralprofile_\convset\parenth{\frac{4}{\varJ-B}}} \leq
  - \lazyparam t.
\end{align*}
Using the change of variable $v = 4/\parenth{\varJ-B}$, we obtain
\begin{align}
  \label{eq:varJ_solved_bound} \lazyparam
  t \leq \int_{4/\parenth{\varJ(0)-B}}^{4/\parenth{\varJ(t)-B}} \frac{dv}{v \spectralprofile_\convset(v)}
\end{align}
Furthermore, equation~\eqref{eq:varJ_solved_bound} implies that for $T
\geq
\frac{1}{\lazyparam} \int_{4/\warmparam}^{8/\errorparam^2}
\frac{dv}{v\spectralprofile_\convset(v)}$, we have
\begin{align*}
 \int_{4/\warmparam}^{8/\errorparam^2} \frac{dv}{v\spectralprofile_\convset(v)} \leq \int_{4/\parenth{\varJ(0)-B}}^{4/\parenth{\varJ(T)-B}} \frac{dv}{v\spectralprofile_\convset(v)}.
\end{align*}
The bound~\eqref{eq:bound_on_var0} and the fact that $B
= \errorparam^2/3$ imply that $4/(\varJ(0) -B) > 4/\warmparam$.  Using
this observation, the fact that
$0 \leq \spectralprofile_\convset(v) < \infty$ for $v \geq
4/\warmparam$ and combining with the case in Equation~\eqref{eq:varJ_without_recursion}, we conclude that $\varJ$ satisfies
\begin{align*}
  \varJ(T) \leq 3B = \errorparam^2 \text{ or
  } \frac{4}{\varJ(T) - B} \geq \frac{8}{\errorparam^2} \quad\text{for
  }T \geq \frac{1}{\lazyparam} \int_{4/\warmparam}^{8/\errorparam^2}
\frac{dv}{v\spectralprofile(v)},
\end{align*}
which implies the claimed
bound~\eqref{eq:mixing_time_in_terms_of_var}.\\

Finally, we turn to the proof of
Lemma~\ref{lem:lower_bound_Dirichlet_w_spectralprofile}.

\paragraph{Proof of Lemma~\ref{lem:lower_bound_Dirichlet_w_spectralprofile}:}
\label{par:proof_of_lemma_lem:lower_bound_dirichlet_w_spectralprofile}

Fix a function $g: \statespace \rightarrow \real_+$ such
  that $g \in \Ltpi$ and $g \cdot \indicator_\convset$ is not constant and  $\Exs[g^2 \cdot \indicator_\convset] \geq 2 \Exs[g^2 \cdot \indicator_{\convset^c}]$. Note that for any constant $\gamma > 0$, we have
\begin{align*}
  \dirichlet(g, g) &= \frac{1}{2} \int_{(x,
  y) \in \statespace^2} \parenth{g(x) - g(y)}^2 \transprob(x,
  dy) \target(x) dx \\
  &= \frac{1}{2} \int_{(x,
  y) \in \statespace^2} \parenth{(g(x) - \gamma) - (g(y) - \gamma)}^2 \transprob(x, dy) \target(x) dx \\
  &= \dirichlet\parenth{(g - \gamma), (g - \gamma)}.
\end{align*}
Let $\gamma = \Varpi(g \cdot \indicator_\convset) / 4 \Exspi\parenth{g \cdot \indicator_\convset}$. We have
\begin{align}
  \label{eq:lower_bound_dirichlet_1} \dirichlet(g, g)
  = \dirichlet\parenth{(g - \gamma), (g - \gamma)} & \stackrel{(i)}{\geq} \dirichlet\parenth{(g - \gamma)_+, (g - \gamma)_+} \notag \\
  & \stackrel{(ii)}{\geq} \Varpi\parenth{(g - \gamma)_+ \cdot \indicator_\convset } \inf_{f\in c_{0, \convset}^+(\braces{ g > \gamma})} \frac{\dirichlet(f, f)}{\Varpi\parenth{f \cdot \indicator_\convset}} \notag \\
  & \stackrel{(iii)}{\geq} \Varpi\parenth{(g - \gamma)_+ \cdot \indicator_\convset } \cdot \spectralprofile_\convset
  (\target(\braces{g > \gamma} \cap \Omega)).
\end{align}
Here $(x)_+ = \max\braces{0, x}$ denotes the positive part of
$x$. Inequality (i) follows from Lemma 2.3 in~\cite{goel2006mixing}. Inequality (iii)
follows from the definition~\eqref{eq:def_spectralprofile} of
$\convset$-restricted spectral profile.
Inequality (ii) follows from the definition of infimum and we need to verify that $\Exspi[(g-c)_+^2 \indicator_{\convset^c}] \leq \Exspi[(g-c)_+^2 \indicator_\convset]$. It follow because
\begin{align*}
  \Exspi[(g-c)_+^2 \indicator_\convset] &\stackrel{(iv)}{\geq} \Exspi[\parenth{g \cdot \indicator_\convset}^2] - 2\gamma \cdot \Exspi \parenth{g \cdot \indicator_\convset} \\
  &= \frac{\Exspi[\parenth{g \cdot \indicator_\convset}^2] + \parenth{\Exspi[g \cdot \indicator_\convset]}^2}{2} \\
  &\stackrel{(v)}{\geq} \Exspi[\parenth{g \cdot \indicator_{\convset^c}}^2] \\
  &\stackrel{(vi)}{\geq}\Exspi[\parenth{(g-c)_+ \cdot \indicator_{\convset^c}}^2],
\end{align*}
where inequality~(iv) and (vi) follows from the fact that
\begin{align}
  \label{eq:a_b_nonegative_inequality}
  (a - b)_+^2 \geq a^2 - 2 ab \quad\text{and}\quad (a - b)_+ \leq
  a, \quad\mbox{for scalars $a, b \geq 0$},
\end{align}
inequality (v) follows from the assumption in Lemma~\ref{lem:lower_bound_Dirichlet_w_spectralprofile} that $\Exspi[g^2 \cdot \indicator_\convset] \geq 2 \Exspi[g^2 \cdot \indicator_{\convset^c}]$.

Additionally, we have
\begin{align}
  \label{eq:lower_bound_dirichlet_variance_c} \Varpi\parenth{(g - \gamma)_+ \cdot \indicator_\convset }
  &= \Exspi\parenth{(g - \gamma)_+ \cdot \indicator_\convset}^2 - \brackets{\Exspi\parenth{(g - \gamma)_+ \cdot \indicator_\convset}}^2 \notag \\
  &\stackrel{(i)}{\geq} \Exspi \parenth{g \cdot \indicator_\convset}^2 -
  2\gamma \cdot \Exspi \parenth{g \cdot \indicator_\convset}
  - \brackets{\Exspi\parenth{g\cdot \indicator_\convset}}^2 \notag \\
  &\geq \Varpi\parenth{g \cdot \indicator_\convset} - 2 \gamma \cdot \Exspi \parenth{g \cdot \indicator_\convset},
\end{align}
where inequality~(i) follows from the facts in Equation~\eqref{eq:a_b_nonegative_inequality}. Together the choice of $\gamma$, we obtain from
equation~\eqref{eq:lower_bound_dirichlet_variance_c} that
\begin{align}
  \label{eq:lower_bound_dirichlet_2} \Varpi\parenth{(g -
  \gamma)_+ \indicator_\convset } \geq \frac{1}{2}\Varpi\parenth{g \cdot \indicator_\convset}
\end{align}
Furthermore applying Markov's inequality for the non-negative
function $g \cdot \indicator_\convset$, we also have $\target(\braces{g>\gamma} \cap \convset) = \target(\braces{g \cdot \indicator_\convset >\gamma} ) \leq \brackets{\Exspi\parenth{g \cdot \indicator_\convset}} / \gamma$.
Combing equation~\eqref{eq:lower_bound_dirichlet_1}
and~\eqref{eq:lower_bound_dirichlet_2}, together with the fact that
$\spectralprofile_\convset$ is non-increasing, we obtain
\begin{align*}
  \dirichlet(g,
   g) \geq \frac{1}{2}\Varpi\parenth{g \cdot \indicator_\convset} \cdot \spectralprofile_\convset \parenth{\frac{4 \parenth{\Exspi
   (g \cdot \indicator_\convset) }^2}{\Varpi \parenth{g \cdot \indicator_\convset}}},
\end{align*}
as claimed in the lemma.


\subsubsection{Proof of Lemma~\ref{lem:relate_spectralprofile_and_conductanceprofile}}

\label{sub:proof_of_lemma_lem:relate_spectralprofile_and_conductanceprofile}

The proof of the
Lemma~\ref{lem:relate_spectralprofile_and_conductanceprofile} follows
along the lines of Lemma 2.4 in~\cite{goel2006mixing},
except that we have to deal with continuous-state transition
probability. This technical challenge is the main reason for
introducing the restricted conductance profile.  At a high level, our
argument is based on reducing the problem on general functions to a
problem on indicator functions, and then using the definition of the
conductance.  Similar ideas have appeared in the proof of the
Cheeger's inequality~\citep{cheeger1969lower} and the modified
log-Sobolev constants~\citep{houdre2001mixed}.\\

We split the proof of
Lemma~\ref{lem:relate_spectralprofile_and_conductanceprofile} in two
cases based on whether $v \in [\frac{4}{\warmparam}, \frac{\target
(\convset)}{2}]$, referred to as Case 1, or $v \geq \frac{\target
(\convset)}{2}$, referred to as Case 2.


\paragraph{Case 1:}

First we consider the case when $v \in
[\frac{4}{\warmparam}, \frac{\target (\convset)}{2}]$. Define $\discreteD^+: \Ltpi \rightarrow \Ltpi$ as
\begin{align*}
  \discreteD^+(h)(x) = \int_{y \in \statespace} \parenth{h(x) -
  h(y)}_+ \transprob(x, dy) \, \text{and}\, \discreteD^-(h)(x)
  = \int_{y \in \statespace} \parenth{h(x) - h(y)}_- \transprob(x,
  dy),
\end{align*}
where $(x)_+ = \max\braces{0, x}$ and (resp. $(\cdot)_-$) denote the
positive and negative part of $x$ respectively.  We note that
$\discreteD^+$ and $\discreteD^-$ satisfy the following co-area
formula:
\begin{subequations}
\begin{align}
  \label{eq:discrete_coarea_formular} \Exspi \discreteD^+(h)
  = \int_{-\infty}^{+\infty} \Exspi \discreteD^+ \parenth{\indicator_{h > t}}
  dt.
\end{align}
See Lemma 1 in \cite{houdre2001mixed} or Lemma 2.4 in \cite{goel2006mixing} for a proof of the
equality~\eqref{eq:discrete_coarea_formular}.  Moreover, given any
measurable set $A \subset \statespace$, scalar $t$, and function
$h \in c_{0, \convset}^+(A)$, we note that the term
$\Exspi \discreteD^+(\indicator_{h > t})(x)$ is equal to the flow
$\flow$ (defined in equation~\eqref{eq:def_flow}) of the level set
$H_t = \braces{x \in \statespace \mid h(x) > t}$:
\begin{align}
  \Exspi \discreteD^+ (\indicator_{h > t}) = \int_{x \in
  H_t}\transprob(x, H_t^c) \targetdensity(x) dx = \flow(H_t).
\end{align}
By the definition of infimum, we have
\begin{align}
  \label{eq:flow_lower_bound_by_conductance} \flow(H_t) \geq \target(H_t \cap \convset) \cdot \inf_{0 \leq \target(S \cap \convset) \leq \target(A \cap \convset)} \frac{\flow(S)}{\target(S \cap \convset)}.
\end{align}
\end{subequations}
Combining the previous three equations, we obtain\footnote{Note
that this step demonstrates that the continuous state-space treatment
is different from the discrete state-space one in Lemma 2.4 of \cite{goel2006mixing}.}
\begin{align*}
  \Exspi \discreteD^+(h)
  = \int_{-\infty}^{+\infty} \Exspi \discreteD^+ \parenth{\indicator_{h > t}} dt
  &\geq \int_{-\infty}^{+\infty} \target(H_t \cap \convset)
  dt \cdot \inf_{\substack{S \subset \statespace \\ 0 \leq \target(S \cap \convset) \leq \target(A \cap \convset)}} \frac{\flow(S)}{\target(S \cap \convset)} \\
  & = \Exspi(h \cdot \indicator_\convset) \cdot \conductanceprofile(\target(A \cap \convset)),
\end{align*}
where the last equality follows from that $h \geq 0$ and the definition of the restricted conductance.
In a similar fashion, using the fact that $\flow(H_t) = \flow(H_t^c)$, we obtain that
\begin{align*}
  \Exspi \discreteD^-(h) \geq \Exspi(h \cdot \indicator_\convset) \cdot \conductanceprofile(\target(A \cap \convset)).
\end{align*}
Combining the bounds on $\Exspi \discreteD^+(h)$ and $\Exspi \discreteD^-(h)$, we obtain
\begin{align*}
  \int_{\statespace}\int_{\statespace} \abss{h(x) -
  h(y)} \transprob(x, dy) \targetdensity(x) dx
  = \Exspi \discreteD^+(h) + \Exspi \discreteD^-(h) \geq 2 \Exspi(h  \cdot \indicator_\convset) \cdot \conductanceprofile(\target(A \cap \convset)).
\end{align*}
Given any function $g \in c_{0, \convset}^+(A)$, applying the above inequality by replacing $h$ with $g^2$, we have
\begin{align*}
  &2 \Exspi(g^2 \cdot \indicator_\convset) \cdot \conductanceprofile(\target(A \cap \convset)) \\
  & \leq\int_{\statespace}\int_{\statespace} \abss{g^2(x) - g^2(y)} \transprob(x, dy) \targetdensity(x) dx \\
  & \stackrel{(i)}{\leq} \parenth{ \int_{\statespace}\int_{\statespace}\abss{g(x) - g(y)}^2 \transprob(x, dy) \targetdensity(x) dx}^{1/2} \cdot \parenth{\int_{\convset}\int_{\convset} \abss{g(x) + g(y)}^2 \transprob(x, dy) \targetdensity(x) dx}^{1/2}  \\
  & \stackrel{(ii)}{\leq} \parenth{2 \dirichlet(g,
  g)}^{1/2} \cdot \parenth{ \int_{\statespace}\int_{\statespace}
  2\parenth{g(x)^2 + g(y)^2} \transprob(x, dy) \targetdensity(x)
  dx}^{1/2} \\
  & = \parenth{2 \dirichlet(g, g)}^{1/2} \parenth{4 \Exspi(g^2)}^{1/2} \\
  &\stackrel{(iii)}{\leq} \parenth{2 \dirichlet(g, g)}^{1/2} \parenth{8 \Exspi(g^2 \cdot \indicator_\convset)}^{1/2}.
\end{align*}
Rearranging the last equation, we obtain
\begin{align}
  \label{eq:dirichlet_ratio_lower_bound_conductanceprofile} \frac{\dirichlet(g,
  g)}{\Exspi(g^2 \cdot \indicator_\convset)} \geq \frac{\conductanceprofile^2(\target(A \cap \convset))}{4}.
\end{align}
In the above sequence of steps, inequality~(i) follows from the
Cauchy-Schwarz inequality, and inequality~(ii) from the
definition~\eqref{eq:def_dirichlet_form} and the fact that
$(a+b)^2\leq 2(a^2+b^2)$. Inequality~(iii) follows from the assumption in the definition of the spectral profile~\eqref{eq:def_respectralgap} that $\Exspi[g^2 \cdot \indicator_{\convset^c}] \leq \Exspi[g^2 \cdot \indicator_\convset]$. Taking infimum over $g \in
c_{0, \convset}^+(A)$ in
equation~\eqref{eq:dirichlet_ratio_lower_bound_conductanceprofile}, we
obtain
\begin{align*}
  \respectralgap_{\convset}(A) = \inf_{g \in
  c_{0, \convset}^+(A)} \frac{\dirichlet (g,
  g)}{\Varpi(g \cdot \indicator_\convset)} \geq \inf_{g \in
  c_{0, \convset}^+(A)} \frac{\dirichlet (g,
  g)}{\Exspi(g^2 \cdot \indicator_\convset)} \geq \frac{ \conductanceprofile^2(\target(A \cap \convset))
  }{4},
\end{align*}
where the first inequality follows from the fact that
$\Exspi(g^2 \cdot \indicator_\convset) \geq \Varpi(g \cdot \indicator_\convset)$.  Given \mbox{$v \in
[0, \frac{\target(\convset)} {2}]$}, taking infimum over
$\target(A \cap \convset) \leq v$ on both sides, we conclude the
claimed bound for this case:
\begin{align*}
  \spectralprofile_{\convset}(v) = \inf_{\target(A \cap \convset) \in
  [0,
  v]} \respectralgap_{\convset}(A) \geq \inf_{\target(A \cap \convset) \in
  [0, v]} \frac{ \conductanceprofile^2(\target(A \cap \convset)) }{4}
  = \frac{ \conductanceprofile^2(v)}{4},
\end{align*}
where the last equality follows from the fact that the conductance
profile $\conductanceprofile$ defined in
equation~\eqref{eq:def_conductanceprofile} is non-increasing over its
domain $[0, \frac{\target (\convset)}{2}]$.


\paragraph{Case 2:}
Next, we consider the case when $v \geq \frac{\target(\convset)}{2}$.
We claim that
\begin{align}
\label{eq:relate_spectralprofile_and_conductanceprofile_for_large_v_proof}
\spectralprofile_{\convset}(v) \stackrel{(i)}{\geq} \spectralprofile_{\convset}(\target(\convset))
\stackrel{(ii)}{\geq} \frac{\spectralprofile_{\convset}(\target(\convset)/2)}{2} \stackrel{(iii)}{\geq} \frac{\conductanceprofile(\target(\convset)/2)^2}{8},
\end{align}
where step~(i) follows from the fact that the spectral profile
$\spectralprofile$ is a non-increasing function, and step~(iii) from
the result of Case 1.  Note that the bound from
Lemma~\ref{lem:relate_spectralprofile_and_conductanceprofile} for this
case follows from the bound above.  It remains to establish
inequality~(ii), which we now prove.

Note that given the definition~\eqref{eq:def_spectralprofile}, it
suffices to establish that
\begin{align}
\label{eq:case_2_reduced_case}
  \frac{\dirichlet\parenth{g, g
   }}{\Varpi(g\cdot \indicator_\convset)} \geq \frac{\spectralprofile_{\convset}(\target(\convset)/2)}{2} \quad\mbox{for
   all functions $g \in \Ltpi$.}
\end{align}
Consider any $g \in \Ltpi$ and let $\median \in \real$ be such that
\begin{align*}
  \target (\braces{g > \median} \cap \convset) = \target(\braces{g
< \median} \cap
\convset) = \frac{\target(\convset)}{2}.
\end{align*}
Using the same argument as in the proof of
Lemma~\ref{lem:lower_bound_Dirichlet_w_spectralprofile} and Lemma 2.3 in~\cite{goel2006mixing}, we have
\begin{align}
  \label{eq:dirichlet_lower_bound_median_1} \dirichlet\parenth{g, g}
  &= \dirichlet\parenth{(g - \median),
  (g-\median)} \notag\\
  &\geq \dirichlet\parenth{(g - \median)_+ ,
  (g-\median)_+ } + \dirichlet\parenth{(g
  - \median)_- , (g
  - \median)_- }.
\end{align}
For the two terms above, we have
\begin{align}
  \label{eq:dirichlet_lower_bound_median_2p} \dirichlet\parenth{(g
  - \median)_+ , (g
  - \median)_+ } \geq \Exspi\parenth{(g
  - \median)^2_+ \cdot \indicator_\convset} \cdot \inf_{f \in
  c_{0, \convset}^+(\braces{g
  > \median})} \frac{\dirichlet\parenth{f,
  f}}{\Exspi\parenth{f \cdot \indicator_\convset}^2},
\end{align}
and similarly
\begin{align}
  \label{eq:dirichlet_lower_bound_median_2m} \dirichlet\parenth{(g
  - \median)_- , (g
  - \median)_- } \geq \Exspi\parenth{(g
  - \median)^2_- \cdot \indicator_\convset} \cdot \inf_{f \in
  c_{0, \convset}^+(\braces{g
  < \median})} \frac{\dirichlet\parenth{f,
  f}}{\Exspi\parenth{f \cdot \indicator_\convset}^2}.
\end{align}
For $f \in c_{0, \convset}^+(\braces{g > \median})$, we have $f \cdot \indicator_\convset \in c_{0, \convset}^+(\braces{g > \median} \cap \convset)$. Using
Cauchy-Schwarz inequality, we have
\begin{align*}
  \Exspi \parenth{f \cdot \indicator_\convset}^2 = \int_{x \in \braces{g > \median} \cap \convset }
  f(x)^2 \target(x) dx \geq \frac{\parenth{\int_{x \in \braces{g
  > \median} \cap \convset } \abss{f(x)} \target(x)
  dx}^2}{\target(\braces{g > \median} \cap \convset)} \geq \frac{(\Exspi f \cdot \indicator_\convset)^2}{\target(\braces{g > \median} \cap \convset)}
\end{align*}
Using this bound and noting the $\median$ is chosen such that
$\target(\braces{g > \median} \cap \convset)=\target(\convset)/2$,
for $f \in c_{0, \convset}^+(\braces{g > \median})$, we have
\begin{align}
  \label{eq:dirichlet_lower_bound_median_3_var_exs} \Varpi (f \cdot \indicator_\convset)
  = \Exspi \parenth{f \cdot \indicator_\convset}^2 - \parenth{\Exspi f \cdot \indicator_\convset}^2 \geq \Exspi \parenth{f \cdot \indicator_\convset}^2 \cdot \parenth{1 - \frac{\target(\convset)}{2}}.
\end{align}
Putting the
equations~\eqref{eq:dirichlet_lower_bound_median_1},~\eqref{eq:dirichlet_lower_bound_median_2p},~\eqref{eq:dirichlet_lower_bound_median_2m}
and~\eqref{eq:dirichlet_lower_bound_median_3_var_exs} together, we
obtain
\begin{align*}
  \dirichlet\parenth{g, g}
  & \geq \Exspi\parenth{(g-\median)^2 \cdot \indicator_\convset} \cdot \parenth{1 - \frac{\target(\convset)}{2}} \cdot \inf_{\target(S) \in
  [0,\frac{\target (\convset)}{2}]}\inf_{f \in c_{0, \convset}^+
  (S)} \frac{\dirichlet(f, f)}{\Varpi(f \cdot \indicator_\convset)} \\
  &\geq \Varpi(g \cdot \indicator_\convset) \cdot \frac{1}{2} \cdot \spectralprofile_\convset(\target(\convset)/2).
\end{align*}
which implies the claim~\eqref{eq:case_2_reduced_case} and we conclude Case 2 of Lemma~\ref{lem:relate_spectralprofile_and_conductanceprofile}.


\subsection{Proof of Lemma~\ref{prop:conductanceprofile_via_overlaps}}
\label{sec:proof_of_lemma_lem:conductanceprofile_via_overlaps}

The proof of this lemma is similar to the conductance based proof for
continuous Markov chains (see, e.g., Lemma 2 in our past
work~\cite{dwivedi2018log}). In addition to it, we have to deal with
the case when target distribution satisfies the logarithmic
isoperimetric inequality.

For any set $A_1$ such that $\target(A_1 \cap \convset) \leq \frac{\target(\convset)}
{2}$, with its complement denoted by
\mbox{$A_2 = \statespace\backslash A_1$}, we have $\target(A_2\cap\convset)
\geq \frac{\target(\convset)}{2} \geq \target
(A_1\cap\convset)$, since $\target (A_1\cap\convset) + \target(A_2\cap\convset) = \target(\convset)$.
We claim that
\begin{align}
\label{eq:simple_conductance_bound}
  \int_{x \in A_1} \transprob(x, A_2) \targetdensity(x) dx
  \geq \target(A_1\cap\convset)\cdot \frac{\tvoverlap }{4}\cdot \min\braces{1,
  \frac{ \Delta} {16 \isoconst_\logisopower} \cdot \log^{\logisopower}\parenth{1 + \frac{1}
  {\target(A_1\cap\convset)}}}.
\end{align}
Note that the claim~\eqref{eq:conductanceprofile_via_overlaps} of Lemma~
\ref{prop:conductanceprofile_via_overlaps} can be directly obtained
from the claim~\eqref{eq:simple_conductance_bound},
by dividing both sides by $\target(A_1\cap\convset)$, taking infimum
with respect to $A_1$ such $\target(A_1\cap\convset) \in (0, v]$ and noting
that $\inf_{t\in(0, v]}\log^{\frac{1}{2}}(1+1/t) =\log^{\frac{1}{2}}(1+1/v)$.\\

We now prove the claim~\eqref{eq:simple_conductance_bound}.\\

Define the following sets,
\begin{align}
  \label{eq:def_low_conductance_sets}
  A_1' \defn \braces{x \in A_1 \cap \convset \mid \transprob(x, A_2) < \frac{\tvoverlap}{2}},\quad
  A_2'\defn \braces{x \in A_2 \cap \convset \mid \transprob(x, A_1) < \frac{\tvoverlap}
  {2}},
\end{align}
along with the complement
$A_3'\defn\convset \setminus \parenth{A_1' \cup A_2'}$.
Note that $A_i' \subset\convset$ for $i=1, 2, 3$.
We split the proof into two distinct cases:
\begin{itemize}
  \item Case 1: $\target(A_1') \leq \target(A_1 \cap \convset)/2$ or $\target(A_2') \leq \target(A_2 \cap \convset)/2$.
  \item Case 2: $\target(A_1') > \target(A_1 \cap \convset)/2$ and $\target(A_2') > \target(A_2 \cap \convset)/2$.
\end{itemize}
Note that these cases are mutually exclusive and exhaustive.
We now consider these cases one by one.

\paragraph{Case 1:}
If we have $\target(A_1') \leq \target(A_1 \cap \convset)/2$,
then
\begin{align}
  \label{eq:case1_assumption_conduct_set_large}
  \target(A_1 \cap \convset \setminus A_1') \geq \target(A_1 \cap \convset)/2.
\end{align}
We have
\begin{align*}
  \int_{x \in A_1} \transprob(x, A_2) \targetdensity(x) dx
  \geq  \int_{x \in A_1 \cap \convset \setminus A_1'} \transprob(x, A_2) \targetdensity(x) dx
  &\stackrel{(i)}{\geq} \frac{\tvoverlap}{2} \int_{x \in A_1 \cap \convset \setminus A_1'} \targetdensity(x) dx\\
  &\stackrel{(ii)}{\geq} \frac{\tvoverlap}{4} \target(A_1 \cap \convset),
\end{align*}
where inequality~(i) follows from the definition of the set $A_1'$
in equation~\eqref{eq:def_low_conductance_sets} and
inequality~(ii) follows from equation~\eqref{eq:case1_assumption_conduct_set_large}.
For the case $\target(A_2') \leq \target(A_2 \cap \convset)/2$, we use a similar argument with the role of $A_1$ and $A_2$ exchanged to obtain
\begin{align*}
  \int_{x \in A_1} \transprob(x, A_2) \targetdensity(x) dx = \int_{x \in A_2} \transprob(x, A_1) \targetdensity(x) dx \geq \frac{\tvoverlap}{4} \target(A_2 \cap \convset).
\end{align*}
Putting the pieces together for this case, we have established
that
\begin{align}
  \label{eq:case1_conductance_lower_bound}
  \int_{x \in A_1} \transprob(x, A_2) \targetdensity(x) dx \geq \frac{\tvoverlap}{4} \min\braces{\target(A_1 \cap \convset), \target(A_2 \cap \convset)}
  =\frac{\tvoverlap}{4} \target(A_1 \cap \convset).
\end{align}

\paragraph{Case 2:}
We have $\target(A_1') > \target(A_1 \cap \convset)/2$
and $\target(A_2') > \target(A_2 \cap \convset)/2$.
We first show that in this case the sets $A_1'$ and $A_2'$ are far away,
and then we invoke the logarithmic isoperimetry inequality from Lemma~\ref{lem:logarithmic_isoperimetric_ineq}.

For any two vectors $u \in A_1'$ and $v \in A_2'$, we have
\begin{align*}
  \tvdist{\transition_u}{\transition_v} \geq \transprob(u, A_1) - \transprob(v, A_1) = 1 - \transprob(u, A_2) - \transprob(v, A_1) > 1 - \tvoverlap.
\end{align*}
Consequently, the assumption of the lemma implies that
\begin{align}
  \label{eq:distance_A1_A2_prime}
  d(A_1', A_2') \geq \Delta.
\end{align}
Using the fact that under the stationary distribution, the flow from $A_1$
to $A_2$ is equal to that from $A_2$ to $A_1$, we obtain
\begin{align}
  \label{eq:flow_larger_than_middle_area}
  \int_{x \in A_1} \transprob(x, A_2) \targetdensity(x) dx &= \frac{1}{2}\parenth{\int_{x \in A_1} \transprob(x, A_2) \targetdensity(x) dx + \int_{x \in A_2} \transprob(x, A_1) \targetdensity(x) dx} \notag \\
  &\geq \frac{1}{4} \parenth{\int_{x \in A_1\cap \convset \setminus A_1'} \transprob(x, A_2) \targetdensity(x) dx + \int_{x \in x \in A_2\cap \convset \setminus A_2'} \transprob(x, A_1) \targetdensity(x) dx} \notag \\
  &\geq \frac{\tvoverlap}{8} \target(\convset\setminus (A_1' \cup A_2')),
\end{align}
where the last inequality follows from the definition of the set $A_1'$
in equation~\eqref{eq:def_low_conductance_sets}. Note that the sets $A_1'$,
$A_2'$ and $\statespace\setminus (A_1' \cup A_2')$ partition $\statespace$. Using the condition~\eqref{eq:assumption_isoperimetric} with the $\convset$-restricted distribution $\target_\convset$ with density $\targetdensity_\convset$ defined as
\begin{align*}
  \targetdensity_\convset(x) = \frac{\targetdensity(x) \indicator_\convset
  (x)}{\target(\convset)},
\end{align*}
we obtain
\begin{align}
  \label{eq:logarithmic_isoperimetric_ineq_applied_to_middle_area}
  &\target(\convset \setminus (A_1' \cap A_2')) \notag \\
  &= \target(\convset) \cdot
  \target_\convset(\statespace \setminus (A_1' \cap A_2')) \notag \\
  &\stackrel{(i)}{\geq} \target(\convset) \cdot \frac{d(A_1', A_2')}{2\isoconst_\logisopower} \cdot \min\braces{\target_\convset(A_1'), \target_\convset(A_2')} \cdot \log^{\logisopower}\parenth{1 + \frac{1}{\min\braces{\target_\convset(A_1'), \target_\convset\parenth{A_2'}}}} \notag\\
  &\stackrel{(ii)}{\geq}\target(\convset) \cdot \frac{\Delta }
  {4\isoconst_\logisopower} \min\braces{\target(A_1 \cap \convset), \target(A_2 \cap \convset)} \cdot \log^{\logisopower}\parenth{1 + \frac{2}{\min\braces{\target(A_1 \cap \convset), \target(A_2 \cap \convset)}}} \notag \\
  &\geq \frac{1}{2} \cdot \frac{\Delta }{4\isoconst_\logisopower} \cdot \target
  (A_1 \cap \convset) \cdot \log^{\logisopower}\parenth{1 + \frac{1}{\target(A_1 \cap \convset)}},
\end{align}
where step~(i) follows from the assumption~\eqref{eq:assumption_isoperimetric},
step~(ii) from the bound~\eqref{eq:distance_A1_A2_prime} and
the facts that $\target_\convset(A_i')\geq \target(A_i') \geq \frac{1}{2}\target
(A_i\cap\convset)$ and that the map $x \mapsto x\log^{\logisopower}(1+1/x)$ is an increasing function for either $\logisopower = \frac{1}{2}$ or $\logisopower = 0$.
Putting the pieces~\eqref{eq:flow_larger_than_middle_area} and~\eqref{eq:logarithmic_isoperimetric_ineq_applied_to_middle_area} together, we conclude that
\begin{align}
\label{eq:case2_conductance_lower_bound}
  \int_{x \in A_1} \transprob(x, A_2) \targetdensity(x) dx
  \geq  \frac{\tvoverlap}{16} \cdot \frac{\Delta}{4\isoconst_\logisopower} \cdot
  \target
  (A_1 \cap \convset) \cdot \log^{\logisopower}\parenth{1 + \frac{1}{\target(A_1
  \cap \convset)}}.
\end{align}
Finally, the claim~\eqref{eq:simple_conductance_bound} follows from combining the two bounds~\eqref{eq:case1_conductance_lower_bound} and \eqref{eq:case2_conductance_lower_bound}
from the two separate cases.


\subsection{Proofs related to Lemma~\ref{lem:transition_closeness}}
\label{sub:proofs_of_properties_of_the_hmc_proposal}

We now present the proof of the intermediate results related to the HMC
chain that were used in the proof of Lemma~\ref{lem:transition_closeness},
namely, Lemmas~\ref{lem:jacob_F_x}, \ref{lem:jacob_G_y}, \ref{lem:grady_log_det_jacob_G_x}
and \ref{lem:q_j_recursion_bounds}.
For simplicity, we
adopt following the tensor notation.
\paragraph{Notations for tensor: } 
\label{par:Notations} Let $\tensor \in \real^{\dims \times \dims \times
\dims}$ be a third order tensor. Let $U \in \real^{\dims \times \dims_1}$,
\mbox{$V \in \real^{\dims \times \dims_2}$}, and
$W \in \real^{\dims \times \dims_3}$ be three matrices. Then the multi-linear
form applied on $(U, V, W)$ is a tensor in $\real^{\dims_1 \times \dims_2 \times \dims_3}$:
\begin{align*}
  \brackets{\tensor(U, V, W)}_{p, q, r} = \sum_{i, j, k \in [\dims]} \tensor_{ijk}U_{ip}V_{jq}W_{kr}.
\end{align*}
In particular, for the vectors $u, v, w \in \real^\dims$, the quantity $\tensor
(u, v, w)$ is a real number that depends linearly on $u, v, w$ (tensor analogue
of the quantity $u\tp M v$ in the context of matrices and vector). Moreover,
the term $\tensor(u, v, \Ind_\dims)$ denotes a vector in $\real^\dims$
(tensor analogue of the quantity $M v$ in the context of matrices and
vector). Finally, the term $\tensor(u, \Ind_\dims, \Ind_\dims)$ represents
a matrix in $\real^{\dims \times \dims}$.

\subsubsection{Proof of Lemma~\ref{lem:jacob_F_x}} 
\label{ssub:proof_of_lemma_lem:jacob_f_x}
We will prove an equivalent statement: for $\internsteps^2\step^2 \leq \frac{1}{4\smoothness}$, there is a matrix $Q(x, y) \in \real^{\dims \times \dims}$ with $\matsnorm{Q}{2} \leq \frac{1}{8}$ such that
  \begin{align}
    \label{eq:jacob_F_x_expression_Q}
    \jacob_x\mapF(x, y) = \internsteps\step\parenth{\Ind_\dims - Q(x, y)},
    \quad \text{for all } x, y \in \statespace.
  \end{align}
Recall from equation~\eqref{eq:hmcstate_integrate_expression_K} that the intermediate iterate $\hmcstate_k$ is defined recursively as
\begin{align*}
  \hmcstate_k = \mapFk(\hmcnoise_0, \hmcstate_0) = \hmcstate_0 + k \step
  \hmcnoise_0 - \frac{k \step^2}{2} \gradf(\hmcstate_0) - \step^2 \sum_
  {j=1}^{k-1}\parenth{k-j} \gradf(\hmcstate_j) \quad\text{for}\quad 1 \leq k \leq \internsteps.
\end{align*}
Taking partial derivative with respective to the first variable, we obtain
\begin{align}
  \label{eq:jacob_Fk_x_recursion}
  \frac{\partial}{\partial \hmcnoise_0} \hmcstate_k = \jacob_{\hmcnoise_0}\mapFk(\hmcnoise_0, \hmcstate_0) = k\step \Ind_\dims - \step^2 \sum_{j=1}^{k-1} (k-j) \hessf_{\hmcstate_j} \jacob_{\hmcnoise_0}\mapFn{j}(\hmcnoise_0, \hmcstate_0),
\end{align}
where $\hessf_{\hmcstate_j}$ is the Hessian of $\targetf$ at $\hmcstate_j$.
We claim that for $1 \leq k \leq \internsteps$,
there is a matrix $Q_k \in \real^{\dims \times \dims}$ with $\matsnorm{Q_k}
{2} \leq \frac{1}{8}$ such that
\begin{align}
  \label{eq:jacob_Fk_x_expression_Qk}
  \jacob_{\hmcnoise_0}\mapFk(\hmcnoise_0, \hmcstate_0) = k\step\parenth{\Ind_\dims - Q_k}.
\end{align}
Note that substituting $k = \internsteps$ in this claim yields the
result of the lemma. We now prove the
claim~\eqref{eq:jacob_Fk_x_expression_Qk} using strong induction.

\paragraph{Base case ($k = 1, 2$):}
For the base case $k=1, 2$, using
equation~\eqref{eq:jacob_Fk_x_recursion}, we have
\begin{align*}
  \jacob_{\hmcnoise_0}\mapFn{1}(\hmcnoise_0, \hmcstate_0)
  &= \step \Ind_\dims, \quad\text{and}\\ \jacob_{\hmcnoise_0}\mapFn{2}(\hmcnoise_0, \hmcstate_0)
  &= 2\step \Ind_\dims
  - \step^2 \hessf_{\hmcstate_1} \jacob_{\hmcnoise_0}\mapFn{1}(\hmcnoise_0, \hmcstate_0)
  = 2\step \parenth{\Ind_\dims
  - \frac{\step^2}{2}\hessf_{\hmcstate_1}}.
\end{align*}
Combining the inequality
$\matsnorm{\hessf_{\hmcstate_1}}{2} \leq \smoothness$ from smoothness
assumption and the assumed stepsize bound
$\step^2 \leq \frac{1}{4\smoothness}$ yields
\begin{align*}
  \matsnorm{\frac{\step^2}{2}\hessf_{\hmcstate_1}}{2} \leq \frac{1}{8}.
\end{align*}
The statement in equation~\eqref{eq:jacob_Fk_x_expression_Qk} is verified for $k = 1, 2$.

\paragraph{Inductive step:}
Assuming that the hypothesis holds for all iterations up to $k$, we now
establish it for iteration $k+1$. We have
\begin{align*}
  \jacob_{\hmcnoise_0}\mapFn{k+1}(\hmcnoise_0, \hmcstate_0)
  &= (k+1) \step \Ind_\dims - \step^2 \sum_{j=1}^k (k+1-j) \hessf_{\hmcstate_j} \jacob_{\hmcnoise_0}\mapFn{j}(\hmcnoise_0, \hmcstate_0) \\
  &\stackrel{(i)}{=} (k+1) \step \Ind_\dims - \step^2 \sum_{j=1}^k (k+1-j) \hessf_{\hmcstate_j} \cdot j \step \parenth{\Ind_\dims - Q_j} \\
  & =(k+1) \step (\Ind_\dims - Q_{k+1}),
\end{align*}
where $Q_{k+1}
= \frac{\step^2}{k+1} \sum_{j=1}^k(k+1-j)j\hessf_{\hmcstate_j}(\Ind_\dims
- Q_j)$. Equality (i) follows from the hypothesis of the
induction. Finally, we verify that the spectral norm of $Q_{k+1}$ is
bounded by $\frac{1}{8}$,
\begin{align*}
  \matsnorm{Q_{k+1}}{2}
  &\leq \frac{1}{k+1} \sum_{j=1}^k \matsnorm{\step^2
  (k+1-j)j\hessf_{\hmcstate_j}}{2}\matsnorm{\Ind_\dims - Q_j}{2} \\
  &\stackrel{(i)}{\leq} \frac{1}{k+1} \sum_{j=1}^k \matsnorm{\step^2 \frac{\internsteps^2}{4} \hessf_{\hmcstate_j}}{2}\matsnorm{\Ind_\dims
  - Q_j}{2} \\
  & \stackrel{(ii)}{\leq} \frac{1}{k+1} \sum_{j=1}^k \frac{1}{16} \parenth{1
  + \frac{1}{8}} \\
& \leq \frac{1}{8}.
\end{align*}
Inequality (i) follows from the inequality
$(k+1-j)j \leq \parenth{\frac{k+1-j+j}{2}}^2 \leq \frac{\internsteps^2}{4}$. Inequalilty
$(ii)$ follows from the assumption
$\internsteps^2 \step^2 \leq \frac{1}{4\smoothness}$ and the
hypothesis $\matsnorm{Q_j}{2} \leq \frac{1}{8}$. This completes the
induction.


\subsubsection{Proof of Lemma~\ref{lem:jacob_G_y}}
\label{ssub:proof_of_lemma_lem:jacob_g_y}

Recall that the backward mapping $\mapG$ is defined implicitly as
\begin{align}
  \label{eq:mapG_implicit_expression_in_proof_jacob_g_y}
  x = y + \internsteps \step \mapG(x, y) - \frac{\internsteps\step^2}{2} \gradf(y) - \step^2 \sum_{k=1}^{\internsteps-1} (\internsteps - k) \gradf\parenth{\mapFk(\mapG(x, y), y)}.
\end{align}
First we check the derivatives of $\mapFk(\mapG(x, y), y)$. Since $\mapFk(\mapG(x, y), y)$ satisfies
\begin{align*}
  \mapFk(\mapG(x, y), y) = y + k\step \mapG(x, y) - \frac{k\step^2}{2} \gradf(y) - \step^2 \sum_{j=1}^{k-1} (k - j) \gradf(\mapFn{j}(\mapG(x, y), y)),
\end{align*}
taking derivative with respect to $y$, we obtain
\begin{align}
  \label{eq:mapFk_G_partial_y_in_proof_jacob_g_y}
  \frac{\partial}{\partial y} \mapFk(\mapG(x, y), y)
  &= \Ind_\dims + k\step \jacob_y\mapG(x, y) - \frac{k\step^2}{2} \hessf(y)
  \notag\\
  &\quad- \step^2 \sum_{j=1}^{k-1} (k-j)\hessf(\mapFn{j}(\mapG(x, y), y))
  \frac{\partial}{\partial y} \mapFn{j}(\mapG(x, y), y).
\end{align}
Using the same proof idea as in the previous lemma, we show by induction that for $1 \leq k \leq \internsteps$, there exists matrices $A_k, B_k \in \real^{\dims \times \dims}$ with $\matsnorm{A_k}{2} \leq \frac{1}{6}$ and $\matsnorm{B_k}{2} \leq \frac{1}{8}$ such that
\begin{align}
  \label{eq:mapFk_G_partial_y_ABform_in_proof_jacob_g_y}
  \frac{\partial}{\partial y} \mapFk(\mapG(x, y), y)  = (\Ind_\dims -A_k) + k\step \parenth{\Ind_\dims - B_k} \jacob_y \mapG(x, y).
\end{align}
\paragraph{Case $k=1$:} The case $k=1$ can be easily checked according to equation~\eqref{eq:mapFk_G_partial_y_in_proof_jacob_g_y}, we have
\begin{align*}
  \frac{\partial}{\partial y} \mapFn{1}\parenth{\mapG(x, y), y} = \Ind_\dims - \frac{\step^2}{2} \hessf(y)  + \step \jacob_y\mapG(x, y)
\end{align*}
It is sufficient to set $A_1 = \frac{\step^2}{2}\hessf(y)$ and $B_1 = 0$.
\paragraph{Case $k$ to $k+1$:} Assume the statement is verified until $k \geq 1$. For $k+1 \leq K$, according to equation~\eqref{eq:mapFk_G_partial_y_in_proof_jacob_g_y}, we have
\begin{align*}
  &\frac{\partial}{\partial y} \mapFn{k+1}(\mapG(x, y), y)\\
  &= \Ind_\dims + (k+1)\step \jacob_y\mapG(x, y) - \frac{(k+1)\step^2}{2} \hessf(y) - \step^2 \sum_{j=1}^{k} (k+1-j)\hessf(\mapFn{j}(\mapG(x, y), y)) \frac{\partial}{\partial y} \mapFn{j}(\mapG(x, y), y) \\
  &= \Ind_\dims - \frac{(k+1)\step^2}{2} \hessf(y)  + (k+1)\step \jacob_y\mapG(x, y) \\
  &\quad- \step^2 \sum_{j=1}^{k} (k+1-j)\hessf(\mapFn{j}(\mapG(x, y), y))
  \parenth{(\Ind_\dims -A_j) + j\step \parenth{\Ind_\dims - B_j} \jacob_y \mapG(x, y)} \\
  &= \Ind_\dims - \frac{(k+1)\step^2}{2} \hessf(y) - \step^2 \sum_{j=1}^{k} (k+1-j)\hessf(\mapFn{j}(\mapG(x, y), y)) (\Ind_\dims -A_j) \\
  &\quad+ (k+1)\step \jacob_y\mapG(x, y) - \step^2 \sum_{j=1}^{k} (k+1-j)\hessf
  (\mapFn{j}(\mapG(x, y), y)) \parenth{j\step \parenth{\Ind_\dims - B_j} \jacob_y \mapG(x, y)}
\end{align*}
To conclude, it suffices to note the following
values of $A_{k+1}$ and $B_{k+1}$:
\begin{align*}
  A_{k+1} &= \frac{(k+1)\step^2}{2} \hessf(y) + \step^2 \sum_{j=1}^{k} (k+1-j)\hessf
  (\mapFn{j}(\mapG(x, y), y)) (\Ind_\dims - A_j),\quad\text{and}\\
  B_{k+1} &= \frac{1}{k+1}\step^2 \sum_{j=1}^{k} (k+1-j)j \hessf(\mapFn
  {j}(\mapG(x, y), y)) \parenth{\Ind_\dims - B_j}.
\end{align*}
We now have the following operator norm bounds:
\begin{align*}
  \matsnorm{A_{k+1}}{2} &\leq \frac{k+1}{2} \step^2 \smoothness + \step^2
  \sum_{j=1}^{k}(k+1-j) \smoothness (1 + \frac{1}{6}) \leq \frac{7}{12}
  (k+1)^2 \step^2 \smoothness \leq \frac{1}{6}, \quad\text{and}\\
  \matsnorm{B_{k+1}}{2} &\leq \frac{1}{k+1} \step^2 (1 + \frac{1}{8}) \smoothness \sum_{j=1}^k (k+1-j) j = \frac{9}{8 \cdot 6} k (k-1) \step^2 \smoothness \leq \frac{1}{8}.
\end{align*}
This concludes the proof of equation~\eqref{eq:mapFk_G_partial_y_ABform_in_proof_jacob_g_y}.
As a particular case, for $k = \internsteps$, we observe that
\begin{align*}
  \mapFn{\internsteps}\parenth{\mapG(x, y), y} = x.
\end{align*}
Plugging it into equation~\eqref{eq:mapFk_G_partial_y_ABform_in_proof_jacob_g_y},
we obtain that
\begin{align*}
  \mathbf{J}_yG(x, y) = \frac{1}{\internsteps \step} \parenth{\Ind_\dims - B_\internsteps}^{-1}\parenth{\Ind_\dims - A_\internsteps}
  \quad\Longrightarrow\quad
  \matsnorm{\mathbf{J}_yG(x, y)}{2} \leq \frac{4}{3\internsteps\step}.
\end{align*}
Plugging the bound on $\matsnorm{\mathbf{J}_y G(x, y)}{2}$ back to equation~\eqref{eq:mapFk_G_partial_y_ABform_in_proof_jacob_g_y} for other $k$, we obtain
\begin{align*}
  \matsnorm{\frac{\partial}{\partial y} \mapFk(\mapG(x, y), y)}{2} \leq 3.
\end{align*}
This concludes the proof of Lemma~\ref{lem:jacob_G_y}.


\subsubsection{Proof of Lemma~\ref{lem:grady_log_det_jacob_G_x}} 
\label{ssub:proof_of_lemma_lem:grady_log_det_jacob_g_x}
Recall that the backward mapping $\mapG$ is defined implicitly as
\begin{align}
  \label{eq:mapG_implicit_expression_in_proof_log_det}
  x = y + \internsteps \step \mapG(x, y) - \frac{\internsteps\step^2}{2} \gradf(y) - \step^2 \sum_{k=1}^{\internsteps-1} (\internsteps - k) \gradf\parenth{\mapFk(\mapG(x, y), y)}.
\end{align}
First we check the derivatives of $\mapFk(\mapG(x, y), y)$. Since $\mapFk(\mapG(x, y), y)$ satisfies
\begin{align*}
  \mapFk(\mapG(x, y), y) = y + k\step \mapG(x, y) - \frac{k\step^2}{2} \gradf(y) - \step^2 \sum_{j=1}^{k-1} (k - j) \gradf(\mapFn{j}(\mapG(x, y), y)),
\end{align*}
we have
\begin{align}
  \label{eq:mapFk_G_partial_x_in_proof_log_det}
  \frac{\partial}{\partial x} \mapFk(\mapG(x, y), y) = k\step \jacob_x\mapG(x, y) - \step^2 \sum_{j=1}^{k-1} (k-j)\hessf(\mapFn{j}(\mapG(x, y), y)) \frac{\partial}{\partial x} \mapFn{j}(\mapG(x, y), y).
\end{align}
Similar to the proof of equation~\eqref{eq:jacob_Fk_x_expression_Qk}, we
show by induction (proof omitted) that for \mbox{$1 \leq k \leq \internsteps$},
there exists matrices $\tilde{Q}_k \in \real^{\dims \times \dims}$ with $\matsnorm{\tilde{Q}_k}{2} \leq \frac{1}{2}$ such that
\begin{align}
    \label{eq:mapFk_G_partial_x_Qform_in_proof_log_det}
    \frac{\partial}{\partial x} \mapFk(\mapG(x, y), y)  = k\step \parenth{\Ind_\dims - \tilde{Q}_k} \mathbf{J}_xG(x, y).
\end{align}
Then, by taking another derivative with respect to $y_i$ in equation~\eqref{eq:mapFk_G_partial_x_in_proof_log_det}, we obtain
\begin{align}
  \label{eq:mapFk_G_partial_x_partial_yi_in_proof_log_det}
  \frac{\partial\partial}{\partial x \partial y_i} \mapFk(\mapG(x, y), y) &= k\step \jacob_{xy_i}\mapG(x, y) && \notag \\
  &- \step^2 \sum_{j=1}^{k-1} (k-j) \Bigg\{ && \thirdf_{\mapFn{j}(\mapG(x, y), y)}\parenth{\frac{\partial \mapFn{j}(\mapG(x, y), y)}{\partial y_i}, \Ind_\dims, \Ind_\dims} \frac{\partial}{\partial x} \mapFn{j}(\mapG(x, y), y) \notag \\
  &  &&+ \hessf_{\mapFn{j}(\mapG(x, y), y)} \frac{\partial\partial}{\partial x\partial y_i} \mapFn{j}(\mapG(x, y), y)\Bigg\}
\end{align}
Now we show by induction that for $1 \leq k \leq \internsteps$, for any $\alpha \in \real^\dims$, we have
\begin{align}
  \label{eq:trace_mapFk_G_partial_x_partial_yi_bound_in_proof_log_det}
  \vecnorm{\sum_{i=1}^\dims \alpha_i \parenth{\frac{\partial\partial}{\partial x \partial y_i} \mapFk(\mapG(x, y), y) \jacob_x\mapG(x, y)^{-1}}}{2}
  &\leq 2k\step \vecnorm{\sum_{i=1}^\dims \alpha_i \parenth{ \jacob_{xy_i}\mapG
  (x, y)\jacob_x\mapG(x, y)^{-1}}}{2} \notag\\
  &\quad+ 2\vecnorm{\alpha}{2} k^3 \step^3 \hessianLip.
\end{align}
\paragraph{Case $k=1$:} We first examine the case $k=1$. According to equation~\eqref{eq:mapFk_G_partial_x_partial_yi_in_proof_log_det}, we have
\begin{align*}
  \sum_{i=1}^\dims \alpha_i \parenth{\frac{\partial\partial}{\partial x \partial y_i} \mapFn{1}(\mapG(x, y), y)\jacob_x\mapG(x, y)^{-1}} = \step \sum_{i=1}^\dims \alpha_i\parenth{\jacob_{xy_i}\mapG(x, y)\jacob_x\mapG(x, y)^{-1}}.
\end{align*}
The statement in equation~\eqref{eq:trace_mapFk_G_partial_x_partial_yi_bound_in_proof_log_det} is easily verified for $k=1$.
\paragraph{Case $k$ to $k+1$:}
Assume the statement~\eqref{eq:trace_mapFk_G_partial_x_partial_yi_bound_in_proof_log_det} is verified until $k$. For $k+1 \leq \internsteps$, according to equation~\eqref{eq:mapFk_G_partial_x_partial_yi_in_proof_log_det}, we have
\begin{align*}
  &\sum_{i=1}^\dims \alpha_i \parenth{\frac{\partial\partial}{\partial x \partial y_i} \mapFn{k+1}(\mapG(x, y), y)\jacob_x\mapG(x, y)^{-1}} \\
  &= (k+1)\step \sum_{i=1}^\dims \alpha_i \parenth{\jacob_{xy_i}\mapG(x, y)\jacob_x\mapG(x, y)^{-1}} \\
  &\quad- \step^2 \sum_{j=1}^k\parenth{k+1-j} \braces{\thirdf_{\mapFn{j}
  (\mapG(x, y), y)}\parenth{\sum_{i=1}^\dims \alpha_i\frac{\partial \mapFn{j}(\mapG(x, y), y)}{\partial y_i}, \Ind_\dims, \Ind_\dims} \frac{\partial}{\partial x} \mapFn{j}(\mapG(x, y), y)\jacob_x\mapG(x, y)^{-1}} \\
  &\quad- \step^2 \sum_{j=1}^k\parenth{k+1-j} \hessf_{\mapFn{j}(\mapG(x,
  y), y)} \sum_{i=1}^\dims \alpha_i \parenth{\frac{\partial\partial}{\partial x \partial y_i} \mapFn{j}(\mapG(x, y), y)\jacob_x\mapG(x, y)^{-1}}.
\end{align*}
In the last equality, we have used the fact that $\thirdf_{\mapFn{j}(\mapG(x, y), y)}$ is a multilinear form to enter the coefficients $\alpha_i$ in the tensor. Let
\begin{align*}
  M_\alpha =  \vecnorm{\sum_{i=1}^\dims \alpha_i \parenth{\jacob_{xy_i}\mapG(x, y)\jacob_x\mapG(x, y)^{-1}}}{2}.
\end{align*}
Applying the hypothesis of the induction, we obtain
\begin{align*}
  & \vecnorm{\sum_{i=1}^\dims \alpha_i \parenth{\frac{\partial\partial}{\partial x \partial y_i} \mapFn{k+1}(\mapG(x, y), y)\jacob_x\mapG(x, y)^{-1}}}{2} \\
  &\stackrel{(i)}{\leq} (k+1)\step  M_\alpha
  + \step^2  \sum_{j=1}^k4(k+1-j)j \hessianLip \vecnorm{\alpha}{2}
  + \step^2 \sum_{j=1}^k(k+1-j) \smoothness\parenth{2j\step M + 2 \vecnorm{\alpha}{2} j^3 \step^3 \hessianLip}\\
  &\leq 2(k+1)\step M_\alpha + 2 \vecnorm{\alpha}{2} (k+1)^3 \step^3 \hessianLip.
\end{align*}
The first inequality (i) used the second part of Lemma~\ref{lem:jacob_G_y}
to bound $\matsnorm{\frac{\partial}{\partial} \mapFk(\mapG(x, y), y)}{2}$.
This completes the induction.
As a particular case for $k = \internsteps$, we note that
\begin{align*}
  \mapFn{\internsteps}(\mapG(x, y), y) = \mapF(\mapG(x, y), y) = x,
\end{align*}
and equation~\eqref{eq:mapFk_G_partial_x_partial_yi_in_proof_log_det} for $k = \internsteps$ gives
\begin{align*}
  0 &= \internsteps \step \jacob_{xy_i}\mapG(x, y) \notag \\
  &- \step^2 \sum_{j=1}^{\internsteps-1} (\internsteps-j)\Bigg\{\thirdf_{\mapFn{j}(\mapG(x, y), y)}\parenth{\frac{\partial \mapFn{j}(\mapG(x, y), y)}{\partial y_i}, \Ind_\dims, \Ind_\dims} \frac{\partial}{\partial x} \mapFn{j}(\mapG(x, y), y) \\
  &+ \hessf_{\mapFn{j}(\mapG(x, y), y)} \frac{\partial\partial}{\partial x\partial y_i} \mapFn{j}(\mapG(x, y), y)\Bigg\}.
\end{align*}
Using the bound in equation~\eqref{eq:trace_mapFk_G_partial_x_partial_yi_bound_in_proof_log_det}, we have
\begin{align*}
  \internsteps \step \vecnorm{\sum_{i=1}^\dims \alpha_i \jacob_{xy_i}\mapG(x, y) \jacob_x\mapG(x, y)^{-1}}{2} \leq \vecnorm{\alpha}{2} \internsteps^3 \step^3 \hessianLip + \frac{1}{2}\internsteps \step \vecnorm{\sum_{i=1}^\dims \alpha_i \jacob_{xy_i}\mapG(x, y) \jacob_x\mapG(x, y)^{-1}}{2}.
\end{align*}
Hence, we obtain
\begin{align*}
  \trace\parenth{{\sum_{i=1}^\dims \alpha_i \jacob_{xy_i}\mapG(x, y) \jacob_x\mapG(x, y)^{-1}}} \leq 2\dims\vecnorm{\alpha}{2} \internsteps^2 \step^2 \hessianLip.
\end{align*}
This is valid for any $\alpha \in \real^\dims$, as a consequence, we have
\begin{align*}
  \vecnorm{\begin{bmatrix}\trace\parenth{[\jacob_x \mapG(x, \hmcstate_0)]^{-1} \jacob_{xy_1} \mapG(x, \hmcstate_0)} \\
  \vdots \\
  \trace\parenth{[\jacob_x \mapG(x, q_0)]^{-1} \jacob_{xy_\dims} \mapG(x, \hmcstate_0)}
  \end{bmatrix}}{2}
 \leq 2\dims \internsteps^2 \step^2 \hessianLip.
\end{align*}
This concludes the proof of Lemma~\ref{lem:grady_log_det_jacob_G_x}.


\subsubsection{Proof of Lemma~\ref{lem:q_j_recursion_bounds}} 
\label{ssub:proof_of_lemma_lem:q_j_recursion_bounds}
We first show equation~\eqref{eq:q_j_diff_recursion_bound} by induction. Then equation~\eqref{eq:q_j_recursion_bound} is a direct consequence of equation~\eqref{eq:q_j_diff_recursion_bound} by summing $k$ terms together.
\paragraph{Case $k=0$:} We first examine the case $k=0$. According to the definition of $\mapFn{k}$ in equation~\eqref{eq:hmcstate_integrate_expression_K}, we have
\begin{align*}
  \mapFn{1}(\hmcnoise_0, \hmcstate_0) = \hmcstate_0 + \step\hmcnoise_0 - \frac{\step^2}{2}\gradf(\hmcstate_0).
\end{align*}
Then the case $k=0$ is verified automatically via triangle inequality,
\begin{align*}
  \vecnorm{\mapFn{1}(\hmcnoise_0, \hmcstate_0) - \hmcstate_0}{2} \leq \step\vecnorm{\hmcnoise_0}{2} + \frac{\step^2}{2}\vecnorm{\gradf(\hmcstate_0)}{2}.
\end{align*}
\paragraph{Case $k$ to $k+1$:} Assume that the statement is verified until $k \geq 0$. For $k+1$, using $\mapFn{j}$ as the shorthand for $\mapFn{j}(\hmcnoise_0, \hmcstate_0)$, we obtain
\begin{align*}
  &\mapFn{k+2} - \mapFn{k+1}\\
  =& \step \hmcnoise_0 - \frac{\step^2}{2} \gradf(\hmcstate_0) - \step^2 \sum_{j=1}^{k+1}\gradf(\mapFn{j}).
\end{align*}
Taking the norm, we have
\begin{align*}
  \vecnorm{\mapFn{k+2} - \mapFn{k+1}}{2}
  &\leq \step \vecnorm{\hmcnoise_0}{2} + \frac{(2k+3)\step^2}{2} \vecnorm{\gradf(\hmcstate_0)}{2} + \step^2 \sum_{j=1}^{k+1}\vecnorm{\gradf(\mapFn{j}) - \gradf(\hmcstate_0)}{2} \\
  &\stackrel{(i)}{\leq} \step \vecnorm{\hmcnoise_0}{2} + \frac{(2k+3)\step^2}{2} \vecnorm{\gradf(\hmcstate_0)}{2} + \step^2 \sum_{j=1}^{k+1} \sum_{l=0}^j \vecnorm{\gradf(\mapFn{l+1}) - \gradf(\mapFn{l})}{2}\\
  &\stackrel{(ii)}{\leq} \step \vecnorm{\hmcnoise_0}{2} + \frac{(2k+3)\step^2}{2} \vecnorm{\gradf(\hmcstate_0)}{2} + \step^2 \smoothness \sum_{j=1}^{k+1} \sum_{l=0}^j \vecnorm{\mapFn{l+1} - \mapFn{l}}{2} \\
  &\stackrel{(iii)}{\leq} \step \vecnorm{\hmcnoise_0}{2} + \frac{(2k+3)\step^2}{2} \vecnorm{\gradf(\hmcstate_0)}{2} + \step^2 \smoothness \sum_{j=1}^{k+1} \sum_{l=0}^j \parenth{2 \step \vecnorm{\hmcnoise}{2} + 2(l+1)\step^2\vecnorm{\gradf(\hmcstate_0)}{2}} \\
  &\stackrel{(iv)}{\leq} 2\step \vecnorm{\hmcnoise_0}{2} + (2k+2)\step^2 \vecnorm{\gradf(\hmcstate_0)}{2}.
\end{align*}
Inequality (i) uses triangular inequality. Inequality (ii) uses $\smoothness$-smoothness. Inequality (iii) applies the hypothesis of the induction and inequalities relies on the condition $\internsteps^2\step^2 \leq \frac{1}{4\smoothness}$.
This completes the induction.


\section{Proof of Corollary~\ref{cor:hmc_mixing_sc_fixedK}} 
\label{sec:proof_of_corollary_cor:hmc_mixing_sc_fixedk}

In order to prove Corollary~\ref{cor:hmc_mixing_sc_fixedK}, we first state a
more general corollary of Theorem~\ref{thm:hmc_mixing_general} that does not specify the explicit choice of
step size~$\step$ and leapfrog steps~$\internsteps$. Then we specify two choices of the initial distribution $\initial$ and hyper-parameters $(\internsteps, \step)$ to obtain part (a) and part (b) of Corollary~\ref{cor:hmc_mixing_sc_fixedK}.

\begin{corollary}
  \label{cor:hmc_mixing_sc_general}
  Consider an $(\smoothness, \hessianLip, \scparam)$-strongly log-concave target
  distribution $\target$ (cf. Assumption~\ref{itm:assumptionB}). Fix $\res = \frac{\errorparam^2}{3\warmparam}$. Then the $\frac{1}{2}$-lazy HMC algorithm with initial distribution $\initialstar = \Normal(x^*, \frac{1}{\smoothness}\Ind_\dims)$, step size~$\step$ and leapfrog steps~$\internsteps$ chosen under the condition
  \begin{align}
    \label{eq:hmc_step_choice_sc}
    \step^2 \leq \frac{1}{c\smoothness} \min\braces{\frac{1}{\internsteps^2 \dims^{\frac{1}{2}}}, \frac{1}{\internsteps^2 \dims^{\frac{2}{3}}}\frac{\smoothness}{\hessianLip^{\frac{2}{3}}}, \frac{1}{\internsteps \dims^{\frac{1}{2}}}, \frac{1}{\internsteps^{\frac{2}{3}}\dims^{\frac{2}{3}} \condition^{\frac{1}{3}}\radius(\res)^{\frac{2}{3}}}, \frac{1}{\internsteps \dims^{\frac{1}{2}}\condition^{\frac{1}{2}}\radius(\res)}, \frac{1}{\internsteps^{\frac{2}{3}}\dims} \frac{\smoothness}{\hessianLip^{\frac{2}{3}}}, \frac{1}{\internsteps^{\frac{4}{3}} \dims^{\frac{1}{2}}\condition^{\frac{1}{2}}\radius(\res) } \parenth{\frac{\smoothness}{\hessianLip^{\frac{2}{3}}}}^{\frac{1}{2}} }
  \end{align}
  satisfies the mixing time bounds
  \begin{align*}
    \Tmix_2^\taghmc(\errorparam; \initial) \leq
       c\cdot \max\braces{\log{\warmparam}, \frac{1}{\internsteps^2\step^2\scparam } \log\parenth{\frac{\dims\log{\condition}}{\errorparam}}}.
  \end{align*}
\end{corollary}
\paragraph{Proof of part (a) in Corollary~\ref{cor:hmc_mixing_sc_fixedK}:} 
\label{par:proof_of_part_in_cor1}
Taking the hyper-parameters $\internsteps = \dims^{\frac{1}{4}}$ and $\step = \step_\text{\tiny warm}$ in equation~\eqref{eq:step_hmc_waram_vs_feasible}, we verify that $\step$ satisfies the condition~\eqref{eq:hmc_step_choice_sc}. Given the warmness parameter $\warmparam = O\parenth{\exp\parenth{\dims^{\frac{2}{3}}\condition}}$, we have
\begin{align*}
  \frac{1}{\internsteps^2 \step^2 \scparam} \geq \log(\warmparam).
\end{align*}
Plugging in the choice of $\internsteps$ and $\step$ into Corollary~\ref{cor:hmc_mixing_sc_general}, we obtain the desired result.
\paragraph{Proof of part (b) in Corollary~\ref{cor:hmc_mixing_sc_fixedK}:} 
\label{par:proof_of_part_in_cor2}
We notice that the initial distribution $\initialstar = \Normal(\xstar,
     \frac{1}{\smoothness}\Ind_\dims)$ is $\condition^{\dims/2}$-warm (see Corollary~1 in~\cite{dwivedi2018log}). It is sufficient to plug in the hyper-parameters $\internsteps = \condition^{\frac{3}{4}}$ and $\step = \step_\text{\tiny feasible}$ into Corollary~\ref{cor:hmc_mixing_sc_general}  to obtain the desired result.

Now we turn back to prove Corollary~\ref{cor:hmc_mixing_sc_general}. In order to prove Corollary~\ref{cor:hmc_mixing_sc_general}, we
require the the following lemma, which relates a
$(\smoothness, \hessianLip, \scparam)$-strongly-logconcave target
distribution to a regular target distribution.

\begin{lemma}
\label{lem:strongly_logconcave_regular_to_general}
An $(\smoothness, \hessianLip, \scparam)$-strongly log-concave distribution is
  $(\smoothness, \hessianLip, \res, \isoconst_{\frac{1}{2}}, \gradbound)$-general
  with high mass set $\convset = \truncballres$,
  log-isoperimetric constant $\isoconst_{\frac{1}{2}}
  = \scparam^{-\frac{1}{2}}$ and $\gradbound
  = \smoothness\parenth{\frac{\dims}{\scparam}}^{\frac{1}{2}} \radius(\res)$,
  where the radius is defined in equation~\eqref{eq:def_radius} and
  the convex measurable set $\truncballres$ defined in
  equation~\eqref{eq:def_truncball}.
\end{lemma}

Taking Lemma~\ref{lem:strongly_logconcave_regular_to_general} as
given, Corollary~\ref{cor:hmc_mixing_sc_general} is a direct
consequence of Theorem~\ref{thm:hmc_mixing_general} by plugging the
specific values of $(\convset, \isoconst_{\frac{1}{2}}, \gradbound)$
as a function of strong convexity parameter $\scparam$. The optimal
choices of step-size~$\step$ and leapfrog steps~$\internsteps$ in
Corollary~\ref{cor:hmc_mixing_sc_general} are discussed in
Appendix~\ref{sub:parameter_choice_in_cor_hmc_mixing_sc_general}.

We now proceed to prove
Lemma~\ref{lem:strongly_logconcave_regular_to_general}.


\subsection{Proof of Lemma~\ref{lem:strongly_logconcave_regular_to_general}}
\label{sub:proof_of_lemma_4}

We now prove Lemma~\ref{lem:strongly_logconcave_regular_to_general},
which shows that any
\mbox{$(\smoothness, \hessianLip, \scparam)$-strongly-logconcave} target
distribution is in fact
\mbox{$(\smoothness, \hessianLip, \res, \isoconst_{\frac{1}{2}}, \gradbound)$-regular.}

First, we set $\convset$ to $\truncballres$ as defined in
equation~\eqref{eq:def_truncball}. It is known that this ball has
probability under the target distribution lower bounded as
$\target(\truncballres) \geq 1 - \res$ (e.g. Lemma 1 in the
paper~\cite{dwivedi2018log}).  Second, the gradient bound is a
consequence of the bounded domain. For any $x \in \truncballres$, we
have
\begin{align}
  \label{eq:sc_gradbound} \vecnorm{\gradf(x)}{2} = \vecnorm{\gradf(x)
  - \gradf(\xstar)}{2} \leq \smoothness\vecnorm{x
  - \xstar}{2} \leq \smoothness \parenth{\frac{\dims}{\scparam}}^{\frac{1}{2}} \radius(\res).
\end{align}
Third, we make use of a logarithmic isoperimetric inequality for
log-concave distribution.  We note that the logarithmic isoperimetric
inequality has been introduced in \cite{kannan2006blocking} for the uniform distribution on convex
body and in \cite{lee2018stochastic} for log-concave
distribution with a diameter.  We extend this inequality to strongly
log-concave distribution on $\real^\dims$ following a similar road-map
and provide explicit constants.


\paragraph{Improved logarithmic isoperimetric inequality}
\label{par:improved_logarithmic_isoperimetric_inequality}

We now state the improved logarithmic isoperimetric inequality for
strongly log-concave distributions.
\begin{lemma}
  \label{lem:logarithmic_isoperimetric_ineq} Let $\gamma$ denote the
density of the standard Gaussian distribution
$\Normal\parenth{0, \sigma^2 \Ind_\dims}$, and let $\target$ be a
distribution with density $\targetdensity = q \cdot \gamma$, where $q$
is a log-concave function.  Then for any partition $S_1, S_2, S_3$ of
$\real^\dims$, we have
\begin{align}
\label{eq:logarithmic_isoperimetric_ineq}
\target(S_3) \geq \frac{d(S_1,
 S_2)}{2 \sigma} \min \braces{\target(S_1), \target(S_2)} \log^{\frac{1}{2}} \parenth{1
+ \frac{1}{\min\braces{\target(S_1), \target(S_2)}}}.
\end{align}
\end{lemma}
\noindent See Appendix~\ref{sub:proof_of_lemma_lem:logarithmic_isoperimetric_ineq}
for the proof.\\

Taking Lemma~\ref{lem:logarithmic_isoperimetric_ineq} as given for the
moment, we turn to prove the logarithmic isoperimetric inequality for
the $\convset$-restricted distribution $\target_\convset$ with density
\begin{align*}
\targetdensity_\convset(x)
  = \frac{\targetdensity(x) \indicator_\convset(x)}{\target(\convset)}.
\end{align*}
Since $\targetf$ is $\scparam$-strongly convex, the function
$x \rightarrow f(x) - \frac{\scparam}{2}\vecnorm{x - \xstar}{2}^2$ is
convex. Noting that the class of log-concave function is closed under
multiplication and that the indicator function $\indicator_\convset$
is log-concave, we conclude that the restricted density
$\targetdensity_\convset$ can be expressed as a product of a
log-concave density and the density of the Gaussian distribution
$\Normal(\xstar, \frac{1}{\scparam}\Ind_\dims)$. Applying
Lemma~\ref{lem:logarithmic_isoperimetric_ineq} with $\sigma
= \parenth{\frac{1}{\scparam}}^{\frac{1}{2}}$, we obtain the desired
logarithmic isoperimetric inequality with
$\isoconst_{\frac{1}{2}}=\parenth{\frac{1}{\scparam}}^{\frac{1}{2}}$,
which concludes the proof of
Lemma~\ref{lem:strongly_logconcave_regular_to_general}.


\subsection{Proof of Lemma~\ref{lem:logarithmic_isoperimetric_ineq}}
\label{sub:proof_of_lemma_lem:logarithmic_isoperimetric_ineq}

The main tool for proving general isoperimetric inequalities is the
localization lemma introduced by \cite{lovasz1993random}. Similar result for the
infinitesimal version of
equation~\eqref{eq:logarithmic_isoperimetric_ineq} have appeared as
Theorem 1.1 in~\cite{ledoux1999concentration} and Theorem
30 in~\cite{lee2018stochastic}. Intuitively, the
localization lemma reduces a high-dimensional isoperimetric inequality
to a one-dimensional inequality which is much easier to verify
directly.  In a few key steps, the proof follows a similar road map as
the proof of logarithmic Cheeger inequality~\citep{kannan2006blocking}.

We first state an additional lemma that comes in handy for the proof.
\begin{lemma}
  \label{lem:1d_logarithmic_isoperimetric_ineq}
Let $\gamma$ be the density of the one-dimensional Gaussian
distribution $\Normal\parenth{\nu, \sigma^2}$ with mean $\nu$ and
variance $\sigma^2$. Let $\locnewdensity$ be a one-dimensional
distribution with density given by $\locnewdensity = q \cdot \gamma$,
where $q$ is a log-concave function supported on $[0, 1]$. Let $J_1,
J_2, J_3$ partition $[0, 1]$,
then
\begin{align}
\label{eq:1d_logarithmic_isoperimetric_ineq}
\locnewdensity(J_3) \geq \frac{d(J_1,
J_2)}{2\sigma} \min\braces{\locnewdensity(J_1),\locnewdensity(J_2)} \log^{\frac{1}{2}} \parenth{1
+ \frac{1}{\min\braces{\locnewdensity(J_1), \locnewdensity(J_2)}}}.
\end{align}
\end{lemma}
\noindent See Appendix~\ref{sub:proof_of_lemma_lem:1d_logarithmic_isoperimetric_ineq}
for the proof.\\

We now turn to proving Lemma~\ref{lem:logarithmic_isoperimetric_ineq}
via contradiction: We assume that the
claim~\eqref{eq:logarithmic_isoperimetric_ineq} is not true for some
partition, and then using well known localization techniques, we
construct a one-dimensional distribution that violates
Lemma~\ref{lem:1d_logarithmic_isoperimetric_ineq} resulting in a
contradiction.

Suppose that there exists a partition $S_1, S_2, S_3$ of
$\real^\dims$, such that
\begin{align}
  \label{eq:log_iso_contradiction_hypothesis}
\target(S_3)  < \frac{d(S_1, S_2)}{2 \sigma} \min \braces{\target(S_1), \target(S_2)}
\log^{\frac{1}{2}} \parenth{1 + \frac{1}{\min \braces{\target(S_1), \target(S_2)}}}.
\end{align}
Let $\nu > 0$ denote a sufficiently small number (to be specified
exactly later), such that \mbox{$\nu
< \min\braces{\target(S_1), \target(S_2)}$}.

We now explain the construction of the one-dimensional density that is
crucial for the rest of the argument.  We define two functions
$g: \statespace \rightarrow \real$ and $h: \statespace
\rightarrow \real$ as follows
\begin{align*}
  g(x)
  = \frac{\targetdensity(x) \cdot \indicator_{S_1}(x)}{\target(S_1)
  - \nu} - \targetdensity(x) \quad\text{and}\quad h(x)
  = \frac{\targetdensity(x) \cdot \indicator_{S_2}(x)}{\target(S_2)
  - \nu} - \targetdensity(x).
\end{align*}
Clearly, we have
\begin{align*}
  \int_{\statespace} g(x) dx >
  0\quad\text{and}\quad \int_{\statespace} h(x) dx > 0.
\end{align*}
By the localization lemma (Lemma 2.5 in~\cite{lovasz1993random}; see the corrected form stated as Lemma
2.1 in~\cite{kannan1995isoperimetric}), there exist two
points $a \in \real^\dims, b \in \real^\dims$ and a linear function
$l: [0, 1] \rightarrow \real_+$, such that
\begin{align}
  \label{eq:localization_lemma_g_h} \int_0^1 l(t)^{\dims-1}
  g\parenth{(1-t)a+tb} dt > 0 \quad \text{ and }\quad \int_0^1
  l(t)^{\dims-1} h\parenth{(1-t)a+tb} dt > 0.
\end{align}
Define the one-dimensional density $\locnewdensity: [0,
1] \rightarrow \real^+$ and the sets $J_i, i\in\braces{1, 2, 3}$ as
follows:
\begin{align}
  \label{eq:localziation_lemma_new_density}
\locnewdensity(t) &
  = \frac{l(t)^{\dims-1} \targetdensity\parenth{(1-t)a + tb}}{\int_0^1
  l(u)^{\dims-1} \targetdensity\parenth{(1-u)a + ub}
  du}, \quad\text{and} \quad \\
J_i & = \braces{t\in [0, 1] \mid (1-t)a + tb \in S_i} \quad\text{for }
  i \in \braces{1, 2, 3}.
\end{align}

We now show how the
hypothesis~\eqref{eq:log_iso_contradiction_hypothesis} leads to a
contradiction for the density~$\locnewdensity$.  Plugging in the
definiton of $g$ and $h$ into
equation~\eqref{eq:localization_lemma_g_h}, we find that
\begin{align*}
  \locnewdensity(J_1) > \target(S_1)
  - \nu \quad\text{and}\quad \locnewdensity(J_2) > \target(S_2) - \nu.
\end{align*}
Since $J_1, J_2, J_3$ partition $[0, 1]$, it follows that
\begin{align*}
  \locnewdensity(J_3) < \target(S_3) + 2\nu.
\end{align*}
Since the function $x\mapsto x\log^{\frac{1}{2}}(1+1/x)$ is
monotonically increasing on $[0, 1]$, we have
\begin{align*}
  & \frac{d(S_1,
  S_2)}{2\sigma} \min\braces{\locnewdensity(J_1), \locnewdensity(J_2)} \log^{\frac{1}{2}}\parenth{1+\frac{1}{\min\braces{\locnewdensity(J_1), \locnewdensity(J_2)}}}
  - \locnewdensity(J_3)\\ &\quad\quad\geq \frac{d(S_1,
  S_2)}{2\sigma} \min\braces{\parenth{\locnewdensity
  (S_1)-\nu}, \parenth{\locnewdensity(S_2)-\nu}} \cdot \\
& \quad\quad\quad\quad\quad\quad\quad\quad \log^{\frac{1}{2}}\parenth{1+\frac{1}
  {\min\braces{\parenth{\locnewdensity(S_1)-\nu}, \parenth{\locnewdensity(S_2)
  - \nu}}}} - \parenth{\locnewdensity(S_3)+2\nu}
\end{align*}
The hypothesis~\eqref{eq:log_iso_contradiction_hypothesis} of the
contradiction implies that we can find $\nu$ sufficiently small such
that the RHS in the inequality above will be strictly positive.
Consequently, we obtain
\begin{align}
  \label{eq:log_iso_contradiction_1d_hypothesis_S12} \frac{d(S_1,
  S_2)}{2\sigma} \min\braces{\locnewdensity(J_1), \locnewdensity
  (J_2)} \log^{\frac{1}{2}}\parenth{1+\frac{1}{\min\braces{\locnewdensity(J_1), \locnewdensity(J_2)}}}
  > \locnewdensity(J_3).
\end{align}
Additionally, for $t_1 \in J_1, t_2 \in J_2$, we have $(1-t_1)a + t_1
b \in S_1$ and $(1-t_2)a + t_2 b \in S_2$.  As a result, we have
\begin{align*}
  \abss{t_1-t_2}
  = \frac{1}{\vecnorm{b-a}{2}} \vecnorm{\brackets{(1-t_1)a + t_1 b}
  - \brackets{(1-t_2)a + t_2 b}}{2} \geq \frac{1}{\vecnorm{b-a}{2}}
  d(S_1, S_2),
\end{align*}
which implies that
\begin{align}
  \label{eq:log_iso_relate_distance_J_S} d(J_1,
  J_2) \geq \frac{1}{\vecnorm{b-a}{2}} d(S_1, S_2).
\end{align}
Combining equations~\eqref{eq:log_iso_contradiction_1d_hypothesis_S12}
and~
\eqref{eq:log_iso_relate_distance_J_S}, we obtain that
\begin{align}
  \label{eq:log_iso_contradiction_1d_hypothesis_J} \frac{\vecnorm{b-a}{2} \cdot
  d(J_1, J_2)}{2\sigma} \min\braces{\locnewdensity
  (J_1), \locnewdensity(J_2)} \log^{\frac{1}{2}}\parenth{1+\frac{1}{\min\braces{\locnewdensity(J_1), \locnewdensity(J_2)}}}
  > \locnewdensity(J_3),
\end{align}
which contradicts Lemma~\ref{lem:1d_logarithmic_isoperimetric_ineq}.
Indeed, this contradiction is immediate once we note that the new
density $\locnewdensity$ can also be written as a product of
log-concave density and a Gaussian density with variance
$\frac{\sigma^2}{\vecnorm{b-a}{2}^2}$.


\subsection{Proof of Lemma~\ref{lem:1d_logarithmic_isoperimetric_ineq}}
\label{sub:proof_of_lemma_lem:1d_logarithmic_isoperimetric_ineq}

We split the proof into three cases. Each one is more general than the
previous one. First, we consider the case when $q$ is a constant
function on $[0, 1]$ and the sets $J_1, J_2, J_3$ are all intervals.
In the second case, we consider a general log-concave $q$ supported on
$[0, 1]$ while we still assume that the sets $J_1, J_2, J_3$ are all
intervals.  Finally, in the most general case, we consider a general
log-concave $q$ supported on $[0, 1]$ and $J_1, J_2, J_3$ consist of
an arbitrary partition of $[0, 1]$. The proof idea follows roughly
that of Theorem 4.6 in~\cite{kannan2006blocking}.

Our proof makes use of the Gaussian isoperimetric inequality which we
now state (see e.g., equation~(1.2)
in~\cite{bobkov1999isoperimetric}): Let $\Gamma$ denote the standard
univariate Gaussian distribution and let $\phi_\Gamma$ and
$\Phi_\Gamma^{-1}$ denote its density and inverse cumulative
distribution function respectively.  Given a measurable set
$A\subset\real$, define its $\Gamma$-perimeter $\Gamma^+(A)$ as
\begin{align*}
   \Gamma^+ (A) = {\lim\inf}_{h \rightarrow 0^+} \frac{\Gamma(A+h)
   - \Gamma (A)}{h}, \end{align*} where $A+h
   = \braces{t \in \real \mid \exists a \in A, \abss{t - a} < h}$
   denotes an $h$-neighborhood of $A$. Then, we have
\begin{align}
  \label{eq:gaussian_isoperimetry} \Gamma^+(A) \geq \phi_\Gamma(\Phi_\Gamma^{-1}(\Gamma(A))),
\end{align}
Furthermore, standard Gaussian tail bounds\footnote{E.g., see the
discussion before equation~1 in~\cite{barthe2000some}. The
constant $1/2$ was estimated by plotting the continuous function on
the left hand side via Mathematica.}  estimate imply that
\begin{align}
  \label{eq:gaussian_isoperimetry_elementary_bound} \phi_\Gamma(\Phi_\Gamma^{-1}(t)) \geq \frac{1}{2}
  t \log^{\frac{1}{2}} \parenth{1 + \frac{1}{t}},\quad \text{ for }
  t \in (0, \frac{1}{2}].
\end{align}


\paragraph{Case 1:}
\label{par:case_1_}

First, we consider the case when the function $q$ is constant on $[0,
1]$ and all of the sets $J_1, J_2, J_3$ are intervals. Without loss of
generality, we can shift and scale the density function by changing
the domain, and assume that the density $\locnewdensity$ is of the
form $\locnewdensity(t) \propto e^{-\frac{t^2}{2}}\indicator_{[a,
d]}$.  Additionally, we can assume that $J_1, J_2, J_3$ are of the
form
\begin{align}
\label{eq:three_intervals}
  J_1 = [a, b),\quad J_3 = [b, c], \quad\text{and} \quad J_2 = (c, d],
\end{align}
because the case when $J_3$ is not in the middle is a trivial case.

Applying the inequalities~\eqref{eq:gaussian_isoperimetry} and~\eqref{eq:gaussian_isoperimetry_elementary_bound} with $A=J_2=(c, d]$, we obtain that
\begin{align}
\label{eq:bound_on_phi_c}
  \phi_\gamma(c) = \Gamma^+(J_2) \geq \phi_\gamma
  (\Phi_\gamma^{-1}(\Gamma(J_2))) \geq \frac{\Gamma(J_2)}{2} \log^{\frac{1}{2}} \parenth{1
  + \frac{1}{\Gamma(J_2)}}.
\end{align}
Note that $\locnewdensity(t)
= \frac{\phi_\gamma(t)}{\Phi_\gamma(d)-\Phi_\gamma (a)}\indicator_{[a,
d]}(t)$ and $\locnewdensity(J_2)
= \frac{\Gamma(J_2)}{\Phi_\gamma(d)-\Phi_\gamma (a)}$. We have
\begin{align*}
   \locnewdensity(J_3) = \int_{b}^c \locnewdensity(t) dt \geq
  (c-b) \cdot \locnewdensity(c)
  &=(c-b) \frac{\phi_\gamma(c)}{\Phi_\gamma(d)-\Phi_\gamma(a)}\\
  &\stackrel{(i)}{\geq} \frac{(c-b)}{2} \frac{\Gamma(J_2)}{\Phi_\gamma(d)-\Phi_\gamma(a)} \log^{\frac{1}{2}}\parenth{1+\frac{1}{\Gamma(J_2)}}\\
  & \stackrel{(ii)}{\geq} \frac{c-b}{2} \locnewdensity(J_2) \log^{\frac{1}{2}}\parenth{1+\frac{\Phi_\gamma(d)-\Phi_\gamma(a)}{\Gamma(J_2)}} \\
  &\stackrel{(iii)}{=} \frac{c-b}{2} \locnewdensity(J_2) \log^{\frac{1}{2}}\parenth{1
  + \frac{1}{\locnewdensity(J_2)}} \\
  & \stackrel{(iv)}{\geq} \frac{c-b}{2} \min\braces{\locnewdensity(J_1), \locnewdensity(J_2)} \log^{\frac{1}{2}}\parenth{1
  + \frac{1}{\min\braces{\locnewdensity(J_1), \locnewdensity(J_2)}}},
\end{align*}
where step~(i) follows from the bound~\eqref{eq:bound_on_phi_c} and
step~(ii) follows from the relationship between $\locnewdensity$ and
$\Gamma$ and the facts that $\log$ is an increasing function and that
$\Phi_\gamma(d)-\Phi_\gamma(a) \leq 1$.  Step~(iii) follows from the
definition of $\locnewdensity$ and finally step~(iv) follows from the
increasing nature of the map $t \mapsto t \log^
{1/2}\parenth{1+\frac{1}{t}}$.  This concludes the argument for Case
1.\\

\paragraph{Case 2:}
\label{par:case_2_}

We now consider the case when $q$ is a general log-concave function on
$[0, 1]$ and $J_1, J_2, J_3$ are all intervals. Again we can assume
that $J_1, J_2, J_3$ are of the form~\eqref{eq:three_intervals}, i.e.,
they are given by $J_1 = [a, b), J_3 = [b, c]$, and $J_2 = (c, d]$.

We consider a function $h(t) = \alpha e^{\beta t
- \frac{t^2}{2\sigma^2}}$ such that $h(b) = q(b)$ and $h(c) =
q(c)$.\footnote{This idea of introducing exponential function appeared
in Corollary 6.2 of Kannan et al.~\cite{kannan2006blocking}.} Define
$Q(t_1, t_2) = \int_{t_1}^{t_2} q(t) dt$ and $H(t_1, t_2)
= \int_{t_1}^{t_2} h(t) dt$. Then since $q$ has an extra log-concave
component compared to $h$, we have
\begin{align}
  \label{eq:HQ_naive_bounds} H(a, b) \geq Q(a, b),\quad H(c, d) \geq
  Q(c, d), \text{ but } H(b, c) \leq Q(b, c).
\end{align}
Using the individual bounds in equation~\eqref{eq:HQ_naive_bounds}, we have
\begin{align*}
  \frac{H(a, b)}{H(b, c)} + \frac{H(c, d)}{H(b, c)} \geq \frac{Q(a,
  b)}{Q(b, c)} + \frac{Q(c, d)}{Q(b, c)}.
\end{align*}
From the equation above and the fact that $H(a, b) + H(b, c) + H(c, d) = H(a, d)$, we obtain
\begin{align}
  \label{eq:HQ_elementary_comparison_2}
  \frac{H(b, c)}{H(a, d)} \leq \frac{Q(b, c)}{Q(a, d)}.
\end{align}
To prove the inequality in Case 2, here are two subcases depending on whether $H(a, d) \geq Q(a, d)$ or $H(a, d) < Q(a, d)$.
\begin{itemize}
  \item If $H(a, d) \geq Q(a, d)$, then
\begin{align*}
  \frac{Q(b, c)}{Q(a, d)} & \stackrel{(i)}{\geq} \frac{H(b, c)}{Q(a, d)} \\
  & \stackrel{(ii)}{\geq} \frac{c-b}{2} \cdot \frac{H(a, d)}{Q(a, d)} \cdot \frac{\min(H(a, b), H(c, d))}{H(a, d)} \cdot \log^{\frac{1}{2}}\parenth{1 + \frac{H(a, d)}{{\min(H(a, b), H(c, d))}}} \\
  & \stackrel{(iii)}{\geq} \frac{c-b}{2} \cdot \frac{H(a, d)}{Q(a, d)} \cdot \frac{\min(Q(a, b), Q(c, d))}{H(a, d)} \cdot \log^{\frac{1}{2}}\parenth{1 + \frac{H(a, d)}{{\min(Q(a, b), Q(c, d))}}} \\
  & \stackrel{(iv)}{\geq} \frac{c-b}{2} \cdot \frac{\min(Q(a, b), Q(c, d))}{Q(a, d)} \cdot \log^{\frac{1}{2}}\parenth{1 + \frac{Q(a, d)}{{\min(Q(a, b), Q(c, d))}}}.
\end{align*}
Inequality (i) follows from equation~\eqref{eq:HQ_naive_bounds}; inequality (ii) follows from equation Case 1 because $H$ is covered by Case 1; inequality (iii) uses the fact that the
function $t \mapsto t \log^{\frac{1}{2}}\parenth{1+\frac{1}{t}}$ is
increasing; inequality (iv) follows from the assumption in this subcase $H(a, d) \geq Q(a, d)$.
\item Otherwise $H(a, d) < Q(a, d)$, then we have from equation~\eqref{eq:HQ_naive_bounds}
\begin{align*}
  \frac{H(a, b)}{H(a, d)} \geq \frac{Q(a, b)}{Q(a, d)},\quad \frac{H(c, d)}{Q(a, d)} \geq \frac{Q(c, d)}{Q(a, d)}.
\end{align*}
\begin{align*}
  \frac{Q(b, c)}{Q(a, d)} & \stackrel{(i)}{\geq} \frac{H(b, c)}{H(a, d)} \\
  & \stackrel{(ii)}{\geq} \frac{c-b}{2} \cdot \frac{\min(H(a, b), H(c, d))}{H(a, d)} \cdot \log^{\frac{1}{2}}\parenth{1 + \frac{H(a, d)}{{\min(H(a, b), H(c, d))}}} \\
  & \stackrel{(iii)}{\geq} \frac{c-b}{2} \cdot \frac{\min(Q(a, b), Q(c, d))}{Q(a, d)} \cdot \log^{\frac{1}{2}}\parenth{1 + \frac{Q(a, d)}{{\min(Q(a, b), Q(c, d))}}}.
\end{align*}
Inequality (i) follows from equation~\eqref{eq:HQ_elementary_comparison_2}; inequality (ii) follows from equation Case 1; inequality (iii) uses the fact that the
function $t \mapsto t \log^{\frac{1}{2}}\parenth{1+\frac{1}{t}}$ is
increasing.
\end{itemize}
In both subcases above, we conclude Case 2 using the results established in Case 1.

\paragraph{Case 3:}
\label{par:case_3_}

Finally, we deal with the general case where $J_1, J_2, J_3$ each can
be union of intervals and $q$ is a general log-concave function on
$[0, 1]$.  We show that this case can be reduced to the case of three
intervals, namely, the previous case.

Let $\braces{(b_i, c_i)}_{i \in \indMaxInters}$ be all non-empty
maximal intervals contained in $J_3$. Here the intervals can be either
closed, open or half. That is, $(\cdot, \cdot)$ can be
$[\cdot, \cdot]$, $]\cdot, \cdot[$, $[\cdot, \cdot[$ or
$]\cdot, \cdot]$. For an interval $(b_i, c_i)$, we define its left
surround $LS((b_i, c_i))$ as
\begin{align*}
  LS((b_i, c_i)) = \begin{cases} 2, \text{ if } \exists x_2 \in J_2
  , \parenth{x_2 \leq b_i} \text{ and } \parenth{\nexists x_1 \in J_1,
  x_2 < x_1 \leq b_i}\\ 1, \text{ if } \exists x_1 \in J_1
  , \parenth{x_1 \leq b_i} \text{ and } \parenth{\nexists x_2 \in J_2,
  x_1 < x_2 \leq b_i}\\ 0, \text{ otherwise }.  \end{cases}
\end{align*}
Similarly, we define $RS((b_i, c_i))$ as
\begin{align*}
  RS((b_i, c_i)) = \begin{cases} 2, \text{ if } \exists x_2 \in J_2
  , \parenth{x_2 \geq c_i} \text{ and } \parenth{\nexists x_1 \in J_1,
  x_2 > x_1 \geq c_i}\\ 1, \text{ if } \exists x_1 \in J_1
  , \parenth{x_1 \geq c_i} \text{ and } \parenth{\nexists x_2 \in J_2,
  x_1 > x_2 \geq c_i}\\ 0, \text{ otherwise }.  \end{cases}
\end{align*}
We distinguish two types of intervals. Denote
$G_2 \subset \indMaxInters$ the set containing the indices of all
intervals that are surrounded by either $1$ or $2$ but different.
\begin{align*}
  G_2 \defn \braces{i \in \indMaxInters \mid (LS((b_i, c_i)), RS((b_i,
  c_i))) = (1, 2) \text{ or } (2, 1)}.
\end{align*}
Denote $G_1 \defn \indMaxInters \setminus G_2 $ to be its complement.
By the result settled in case 2, for $i \in G_2$, we have
\begin{align*}
  \locnewdensity([b_i, c_i]) \geq \frac{d(J_1,
  J_2)}{2\sigma} \locnewdensity(I_i) \log^{\frac{1}{2}}\parenth{1
  + \frac{1}{\locnewdensity(I_i)}}
\end{align*}
where $I_i$ is either $[a, b_i]$ or $[c_i, d]$. Summing over all $i\in G_2$, we have
\begin{align}
  \locnewdensity(J_3) \geq \sum_{i \in G_2} \locnewdensity([b_i, c_i])
  &\geq \frac{d(J_1, J_2)}{2\sigma} \sum_{i \in
  G_2} \locnewdensity(I_i) \log^{\frac{1}{2}}\parenth{1
  + \frac{1}{\locnewdensity(I_i)}} \notag \\ &\geq \frac{d(J_1,
  J_2)}{2\sigma} \locnewdensity(\cup_{i\in
  G_2}I_i) \log^{\frac{1}{2}}\parenth{1
  + \frac{1}{\locnewdensity(\cup_{i\in
  G_2}I_i)}}.  \label{eq:log_iso_J3_general_case}
\end{align}
The last inequality follows from the sub-additivity of the map:
$x \mapsto x \log^{\frac{1}{2}}(1+x)$, i.e., for $x > 0$ and $y > 0$,
we have
\begin{align*}
   x \log^{\frac{1}{2}}\parenth{1+\frac{1}{x}} +
   y \log^{\frac{1}{2}}\parenth{1+\frac{1}{y}} \geq
   (x+y) \log^{\frac{1}{2}}\parenth{1+\frac{1}{x+y}}.
\end{align*}
Indeed the sub-additivity follows immediately from the following
observation:
\begin{align*}
  &x \log^{\frac{1}{2}}\parenth{1+\frac{1}{x}} +
  y \log^{\frac{1}{2}}\parenth{1+ \frac{1}{y}} -
  (x+y) \log^{\frac{1}{2}}\parenth{1+\frac{1}{x+y}} \\ &\quad=
  x \brackets{\log^{\frac{1}{2}}\parenth{1+\frac{1}{x}}
  -\log^{\frac{1} {2}} \parenth{1+\frac{1}{x+y}}} +
  y \brackets{\log^{\frac{1}{2}}\parenth{1+\frac{1}{y}}
  - \log^{\frac{1}{2}}\parenth{1+\frac{1}{x+y}}}\\ &\quad\geq 0.
\end{align*}
Finally, we remark that either $J_1$ or $J_2$ is a subset of
$\cup_{i\in G_2} I_i$. If not, there exists $u \in
J_1 \setminus \cup_{i\in G_2} I_i$ and $v \in J_2 \setminus \cup_{i\in
G_2} I_i$, such that $u$ and $v$ are separated by some inverval
$(b_{i^*}, c_{i^*}) \subset J_3$ with $i^* \in G_2$. This is
contradictory with the fact that either $u$ or $v$ must be included in
$I_{i^*}$. Given equation~\eqref{eq:log_iso_J3_general_case}, we use
the fact that the function $x \mapsto
x \log^{\frac{1}{2}}\parenth{1+\frac{1} {x}}$ is monotonically
increasing:
\begin{align*}
  \locnewdensity(J_3) \geq \frac{d(J_1,
  J_2)}{2\sigma} \min\braces{\locnewdensity(J_1), \locnewdensity(J_2)} \log^{\frac{1}{2}}\parenth{1
  + \frac{1}{\min\braces{\locnewdensity(J_1), \locnewdensity(J_2)}}}
\end{align*}
to conclude the proof.


\section{Beyond strongly log-concave target distributions}
\label{sec:beyond_strongly_log_concave}

In this appendix, we continue the discussion of mixing time bounds of
Metropolized HMC from
Section~\ref{sub:mixing_time_bounds_for_hmc_for_general_regular_target_distributions}.
In the next two subsections, we discuss the case when the target is
weakly log-concave distribution or a perturbation of log-concave
distribution, respectively.

\subsection{Weakly log-concave target}
\label{ssub:hmc_mixing_time_bounds_for_weakly_log_concave_target}
The mixing rate in the weakly log-concave case differs depends on further
structural assumptions on the density. We now consider two different scenarios
where either a bound on fourth moment is known or the covariance of the
distribution is well-behaved:
\begin{enumerate}[label=(\Alph*)]
  \setcounter{enumi}{2}
  \item \label{itm:assumptionC}
  The negative log density of the target distribution is $\smoothness$-smooth~\eqref{eq:assumption_smoothness}
  and has $\hessianLip$-Lipschitz Hessian~\eqref{eq:assumption_hessianLip}.
  Additionally for some point $\xstar$, its fourth moment satisfies the
  bound
    \begin{align}
    \label{eq:assumption_fourth_moment}
    \int_{\real^\dims} \vecnorm{x - \xstar}{2}^4 \targetdensity(x) dx \leq
    \frac{\dims^2 \nu^2}{\smoothness}.
  \end{align}
  \item \label{itm:assumptionD} The negative log density of the target distribution is $\smoothness$-smooth~\eqref{eq:assumption_smoothness}
  and has $\hessianLip$-Lipschitz Hessian~\eqref{eq:assumption_hessianLip}.
  Additionally, its covariance matrix satisfies
\begin{align}
\label{eq:assumption_covariance}
  \matsnorm{\int_{x \in \real^\dims} (x - \Exs[x])(x - \Exs[x])\tp\targetdensity
    (x) dx}{\text{op}} \leq 1,
\end{align}
and the norm of the gradient of the negative log density $\targetf$ is bounded
by a constant in the ball $\ball\parenth{\Exs\brackets{x}, \log\parenth{
\frac{1}{\res}} \dims^{3/4} }$ for small enough $\res \geq \res_0$.
\end{enumerate}

When the distribution satisfies assumption~\ref{itm:assumptionC} we consider
HMC chain with slightly modified target and assume that the $\initial$
is $\warmparam$-warm with respect to this modified target distribution
(see the discussion after Corollary~\ref{cor:hmc_mixing_wc} for details).
Moreover, In order to simplify the bounds in the next result, we assume
that $\hessianLip^{2/3} = O(\smoothness)$. A more general result
with without this condition can be derived in a similar fashion.

\begin{corollary}[HMC mixing for weakly-log-concave]
  \label{cor:hmc_mixing_wc}
 Let $\initial$ be a $\warmparam$-warm start,
  $\errorparam \in (0, 1)$ be fixed and consider $\frac{1}{2}$-lazy HMC
  chain with leapfrog steps $\internsteps= \dims^{\frac{1}{2}}$ and step
  size $\step^2=\frac{1}{c\smoothness\dims^{\frac{4}{3}}}$.
  \begin{itemize}
    \item[(a)] If the distribution satisfies assumption~\ref{itm:assumptionC},
    then we have
      \begin{align}
      \label{eq:weak_bound_c}
    \Tmix_{\smalltv}^\taghmc(\errorparam; \initial) \leq
       c\cdot \max\braces{\log{\warmparam}, \frac{\dims^{\frac{4}{3}}\nu}{\errorparam} \log\parenth{\frac{\log{\warmparam}}{\errorparam}}}.
  \end{align}
    \item[(b)] If the distribution satisfies assumption~\ref{itm:assumptionD}
    such that $\res_0 \leq \frac{\errorparam^2}{3\warmparam}$,
    then we have
    \begin{align}
    \label{eq:weak_bound_d}
    \Tmix_2^\taghmc(\errorparam; \initial) \leq
       c\cdot \dims^{\frac{5}{6}} \log\parenth{\frac{\log \warmparam}{\errorparam}}.
  \end{align}
  \end{itemize}
\end{corollary}

As an immediate consequence, we obtain that the number of gradient evaluations
in the two cases is bounded as
\begin{align*}
   \mathcal{B}_1 = \max \braces{ \dims^{\frac{1}{2}} \log \warmparam,
   \frac{\dims^{\frac{11}{6}}\nu}{\errorparam} \log\parenth{\frac{\log{\warmparam}}{\errorparam}}}
   \quad\text{and}\quad
   \mathcal{B}_2 =
   \dims^{\frac{4}{3}}\log\parenth{\frac{\log{\warmparam}}{\errorparam}}.
\end{align*}
We remark that the bound $\mathcal{B}_1$ for HMC chain improves upon the
bound for number of gradient evaluations required by MALA to mix in a similar
set-up. \cite{dwivedi2018log} showed that under assumption~\ref{itm:assumptionC}
(without the Lipschitz-Hessian condition), MALA takes $O({\frac{\dims^2}
{\nu\errorparam}}\log{\frac{\warmparam}{\errorparam}})$ steps to mix.
Since each step of MALA uses one gradient evaluation, our result shows that HMC takes $O(d^{\frac{1}{6}})$ fewer gradient evaluations.
On the other hand, when the target satisfies assumption~\ref{itm:assumptionD}, \cite{mangoubi2019nonconvex} showed that MALA takes $O(
{\dims^{\frac{3}{2}}\log{\frac{\warmparam}{\errorparam}}})$ steps.\footnote{Note
that \cite{mangoubi2019nonconvex}
assume an infinity-norm third order smoothness
which is a stronger assumption than the $\hessianLip$-Lipschitz Hessian
assumption that we made here. Under our setting, the infinity norm third
order smoothness is upper bounded by $\sqrt{\dims} \hessianLip$ and plugging
in this bound changes their rate of MALA from $d^{7/6}$ to $d^{3/2}$.}
Thus even for this case, our result shows that HMC takes $O(d^{\frac{1}
{6}})$ fewer gradient evaluations when compared to MALA.

\paragraph{Proof sketch:} 
\label{par:proof_sketch_}
When the target distribution has a bounded fourth moment (assumption~\ref{itm:assumptionC}),
proceeding as in~\cite{dalalyan2016theoretical},
we can approximate the target distribution $\target$ by a strongly log-concave
distribution $\tilde{\Pi}$ with density given by
\begin{align*}
  \tilde{\pi}(x) = \frac{1}{\int_{\real^\dims} e^{-\tilde{f}(y)} dy } e^{-\tilde{f}(x)} \quad\text{ where } \tilde{f}(x) = f(x) + \frac{\lambda}{2}\vecnorm{x - \xstar}{2}^2.
\end{align*}
Setting $\lambda \defn \frac{2\smoothness\errorparam}{\dims \nu}$ yields
that $\tilde{f}$ is  $\lambda/2$-strongly convex, $\smoothness + \lambda/2$
smooth and $\hessianLip$-Hessian Lipschitz and that the TV distance
$\tvdist{\target}{\tilde{\Pi}} \leq \errorparam/2$ is small. The new condition
number becomes $\widetilde{\condition} \defn 1
+ \dims \nu /\errorparam$. The new logarithmic-isoperimetric constant is
$\widetilde{\isoconst}_{1/2} = (\dims\nu/(\smoothness\errorparam))^{1/2}$.
Thus, in order to obtain an $\errorparam$-accurate sample with respect to
$\target$, it is sufficient to run HMC chain on the new strongly
log-concave distribution $\widetilde{\Pi}$ upto $\errorparam/2$-accuracy.
Invoking Corollary~\ref{cor:hmc_mixing_sc_fixedK} for $\widetilde{\Pi}$
and doing some algebra yields the bound~\eqref{eq:weak_bound_c}.

For the second case (assumption~\ref{itm:assumptionD}), \cite{lee2017eldan} showed
that when the
covariance of $\target$
has a bounded operator norm, it satisfies isoperimetry inequality~\eqref{eq:assumption_isoperimetric}
with $\isoconst_0 \leq O(\dims^{\frac{1}{4}})$.
Moreover, using the Lipschitz concentration~\citep{gromov1983topological},
we have
\begin{align*}
  \Prob_{x \sim \target }\parenth{\vecnorm{x - \Exs_{\target}\brackets{x}}{2} \geq t\isoconst_0 \cdot \sqrt{\dims} } \leq e^{- ct},
\end{align*}
which implies that there exists a constant $c$, for $\convset_\res = \ball\parenth{\Exs_{\target}\brackets{x},
\frac{1}{c}\log\parenth{\frac{1}{\res}}\isoconst_0 \cdot \sqrt{\dims} }$, we have $\target(\convset_\res) \geq 1 - s$.
In addition, assuming that the gradient is bounded in this ball $\convset_\res$
for $\res=\frac{\errorparam^2}{3\warmparam}$ enables us to invoke Theorem~
\ref{thm:hmc_mixing_general}
and obtain the bound~\eqref{eq:weak_bound_d} after plugging in the values
of $\isoconst_0, \internsteps$ and $\step$.

\subsection{Non-log-concave target}
\label{ssub:hmc_mixing_time_bounds_for_nearly_logconcave_target}
We now briefly discuss how our mixing time bounds in
Theorem~\ref{thm:hmc_mixing_general} can be applied for distributions
whose negative log density may be non-convex.  Let $\Pi$ be a
log-concave distribution with negative log density as $\targetf$ and
isoperimetric constant $\isoconst_0$.  Suppose that the target
distribution $\widetilde{\Pi}$ is a perturbation of $\Pi$ with target
density $\widetilde{\pi}(x)$ such that $\widetilde{\pi} (x) \propto
e^{-\targetf(x) - \xi(x)}$, where the perturbation
$\xi: \real^\dims \rightarrow \real$ is uniformly lower bounded by
some constant $-b$ with $b \geq 0$. Then it can be shown that the
distribution $\widetilde{\Pi}$ satisfies isoperimetric
inequality~\eqref{eq:assumption_isoperimetric} with a constant
$\widetilde{\isoconst}_0 \geq e^{-2b}\isoconst_0$.  For example, such
type of a non-log-concave distribution distribution arises when the
target distribution is that of a Gaussian mixture model with several
components where all the means of different components are close to
each other (see e.g.~\cite{ma2018sampling}).  If a bound on
the gradient is also known, Theorem~\ref{thm:hmc_mixing_general} can
be applied to obtain a suitable mixing time bound. However deriving
explicit bounds in such settings is not the focus of the paper and
thereby we omit the details here.




\section{Optimal choice for HMC hyper-parameters}
\label{sec:discussion_around_corollary_cor:hmc_mixing_sc_fixedK}

In this section, we provide a detailed discussion about the optimal leapfrog
steps choice for Metropolized HMC with strongly log-concave target distribution
(Corollary~\ref{cor:hmc_mixing_sc_fixedK}).
We also discuss a few improved convergence rates for Metropolized HMC
under additional assumptions on the target distribution.
Finally, we compare our results for Metropolized HMC with other versions
of HMC namely unadjusted HMC and ODE-solved based HMC in
Subsection~\ref{sub:comparison_with_guarantees_for_other_versions_of_hmc}.

\subsection{Optimal choices for Corollary~\ref{cor:hmc_mixing_sc_general}}
\label{sub:parameter_choice_in_cor_hmc_mixing_sc_general}

Corollary~\ref{cor:hmc_mixing_sc_general} provides an implicit
condition that the step size $\step$ and leapfrog steps $\internsteps$
should satisfy and provides a generic mixing time upper bound that depends
on the choices made. We claim that the optimal choices of $\step$ and
$\internsteps$ according to Table~\ref{tab:hmc_cor1_params} lead to the
following upper bound on number of gradient evaluations required by HMC
to mix to $\errorparam$-tolerance:
\begin{align}
  \label{eq:optimal_choice_cor1} \internsteps \cdot \Tmix_{\smalltv}^\taghmc(\errorparam; \initial)
  &\leq
  O\parenth{\max\braces{ \dims \condition^{\frac{3}{4}}, \dims^{\frac{11}{12}} \condition, \dims^{\frac{3}{4}} \condition^{\frac{5}{4}}, \dims^{\frac{1}{2}} \condition^{\frac{3}{2}}} \cdot \log\frac{1}{\errorparam}}.
\end{align}
This (upper) bound shows that HMC always requires fewer gradient
evaluations when compared to MALA for mixing in total variation distance.
However, such a bound requires a delicate choice of the leap frog steps
$\internsteps$ and $\step$ depending on the condition number
$\condition$ and the dimension $\dims$, which might
be difficult to implement in practice. We summarize these optimal choices
in Table~\ref{tab:hmc_cor1_params}.

\begin{table}[ht]
\centering
  \begin{tabular}{ccl} \toprule \textbf{Case} & $\internsteps$ &
  $\step^2$ \\[2mm] \midrule $\condition \in (0, \dims^{\frac{1}{3}})$
  & $\displaystyle \condition^{\frac{3}{4}}$ &
  $\displaystyle\frac{1}{c\smoothness} \cdot \dims^{-1}\condition^{-\frac{1}{2}}$\\[3mm]
  $\condition \in [\dims^{\frac{1}{3}}, \dims^{\frac{2}{3}}]$ &
  $\displaystyle \dims^{\frac{1}{4}}$ &
  $\displaystyle\frac{1}{c\smoothness} \cdot \dims^{-\frac{7}{6}}$\\[3mm]
  $\condition \in (\dims^{\frac{2}{3}}, \dims]$ &
  $\displaystyle \dims^{\frac{3}{4}}\condition^{-\frac{3}{4}}$ &
  $\displaystyle \frac{1}{c\smoothness} \cdot \dims^{-\frac{3}{2}}\condition^{\frac{1}{2}}$ \\[3mm]
  $\condition \in (\dims, \infty)$ & 1 &
  $\displaystyle\frac{1}{c\smoothness} \cdot \dims^{-\frac{1}{2}}\condition^{-\frac{1}{2}}$ \\[1mm] \bottomrule \end{tabular} \caption{Optimal
  choices of leapfrog steps $\internsteps$ and the step size $\step$
  for the HMC algorithm for an
  $(\scparam, \smoothness, \hessianLip)$-regular target distribution
  such that $\hessianLip~=~O (\smoothness^{\frac{3}{2}})$ used for the
  mixing time bounds in
  Corollary~\ref{cor:hmc_mixing_sc_general}. Here $c$ denotes a
  universal constant.}  \label{tab:hmc_cor1_params}
\end{table}


\paragraph{Proof of claim~\eqref{eq:optimal_choice_cor1}:}
\label{par:proof_of_claim_eq:optimal_choice_cor1}

Recall that under the condition~\eqref{eq:hmc_step_choice_sc}
(restated for reader's convenience)
\begin{align*}
  \step^2 \leq \frac{1}{c\smoothness} \min\braces{\frac{1}{\internsteps^2 \dims^{\frac{1}{2}}}, \frac{1}{\internsteps^2 \dims^{\frac{2}{3}}}\frac{\smoothness}{\hessianLip^{\frac{2}{3}}}, \frac{1}{\internsteps \dims^{\frac{1}{2}}}, \frac{1}{\internsteps^{\frac{2}{3}}\dims^{\frac{2}{3}} \condition^{\frac{1}{3}}\radius(\res)^{\frac{2}{3}}}, \frac{1}{\internsteps \dims^{\frac{1}{2}}\condition^{\frac{1}{2}}\radius(\res)}, \frac{1}{\internsteps^{\frac{2}{3}}\dims} \frac{\smoothness}{\hessianLip^{\frac{2}{3}}}, \frac{1}{\internsteps^{\frac{4}{3}} \dims^{\frac{1}{2}}\condition^{\frac{1}{2}}\radius(\res)
  } \parenth{\frac{\smoothness}{\hessianLip^{\frac{2}{3}}}}^{\frac{1}{2}}
  },
\end{align*}
Corollary~\ref{cor:hmc_mixing_sc_fixedK} guarantees that the HMC mixing
time for the $\condition^{\frac{\dims}{2}}$-warm initialization
$\initialstar=\Normal(\xstar, \smoothness^{-1}\Ind_\dims)$, is
\begin{align*}
  \TMIX_{2}^\taghmc(\errorparam; \initial) = O\parenth{\dims
  + \frac{\condition}{\internsteps^2 \step^2 \smoothness }},
\end{align*}
where we have ignored logarithmic factors.
In order to compare with MALA and other sampling methods, our goal is to
optimize the number of gradient evaluations $\mathcal{G}_{\text{eval}}$ taken
by HMC to mix:
\begin{align}
  \label{eq:number_grad_evals_appendix}
  \mathcal{G}_{\text{eval}}:= \internsteps \cdot \TMIX_{\smalltv}^\taghmc
  (\errorparam; \initial)
  = O\parenth{\internsteps \dims
  + \frac{\condition}{ \internsteps \step^2 \smoothness}}.
\end{align}
Plugging in the condition on $\step$ stated above, we obtain
\begin{align}
  \label{eq:step_condition_explicit_K_appendix}
  \mathcal{G}_{\text{eval}} \leq
\max \bigg\{ \underbrace{\internsteps \dims}_{=:T_1},\quad \underbrace{\internsteps
\max\parenth{\dims^{\frac{1}{2}} \condition, \dims^{\frac{2}{3}} \condition
\hessianLipRatio}}_{=:T_2},\quad \underbrace{\dims^{\frac{1}{2}}\condition^
{\frac{3}{2}}}_{=:T_3},\quad \underbrace{\internsteps^{-\frac{1}{3}}\dims^
{\frac{2}{3}}\condition^{\frac{4}{3}}}_{=:T_4},\quad \underbrace{\internsteps^
{-\frac{1}{3}}\dims\condition\cdot \hessianLipRatio}_{=:T_5},\quad \underbrace{\internsteps^{\frac{1}{3}}\dims^{\frac{1}{2}}\condition^{\frac{3}{2}}\cdot \hessianLipRatio^{\frac{1}{2}}}_{=:T_6}
  \bigg\}
\end{align}
where $\hessianLipRatio = {\hessianLip^{\frac{2}{3}}}/{\smoothness}$. Note
that this bound depends only on the relation between $\dims$, $\condition$
and the choice of $\internsteps$.
We now summarize the source of all of these terms in our proofs:
\begin{itemize}
\item $T_1$: This term is attributed to the warmness of the initial distribution.
The distribution $\initialstar$ is $O(\condition^\dims)$-warm.  This
term could be improved if we have a warmer initial distribution.
\item $T_2$: This term appears in the proposal overlap bound from
equation~\eqref{eq:proposal_closeness} of Lemma~\ref{lem:transition_closeness}
and more precisely, it comes from equation~\eqref{eq:proposal_closeness_stronger}.
\item $T_3,T_4, T_5$ and $T_6$: These terms pop-out from the accept-reject
bound from
  equation~\eqref{eq:transition_proposal_overlap} of
  Lemma~\ref{lem:transition_closeness}. More precisely, $T_3$ and $T_4$
  are a consequence of the first three terms in equation~\eqref{eq:accept_rate_lower_bound_only_p_0},
  and $T_5$ and $T_6$ arise the last two terms in equation~\eqref{eq:accept_rate_lower_bound_only_p_0}.
\end{itemize}
In Table~\ref{tab:hmc_param_choice_explicit}, we summarize
how these six terms can be traded-off to derive
the optimal parameter choices for Corollary~\ref{cor:hmc_mixing_sc_general}.
The effective bound  on $\mathcal{G}_{\text{eval}}$-the number of gradient
evaluations required by HMC to mix, is given by the largest of the six terms.
\begin{table}[H]
\centering
  \begin{tabular}{c|c|cccccc}
  \toprule
  $\condition$ versus $\dims$ & optimal choice $\internsteps$

  & $T_1$ & $T_2$ & $T_3$ & $T_4$ & $T_5$ & $T_6$ \\[3mm]
  & & $\internsteps \dims$
  & $\internsteps \dims^{\frac{2}{3}} \condition$ &
  $\dims^{\frac{1}{2}}\condition^{\frac{3}{2}}$
  &$\internsteps^{-\frac{1}{3}}\dims^{\frac{2}{3}}\condition^{\frac{4}{3}}$
  & $\internsteps^{-\frac{1}{3}}\dims\condition$
  & $\internsteps^{\frac{1}{3}}\dims^{\frac{1}{2}}\condition^{\frac{3}{2}}$
  \\[3mm]
  $\condition \in [1, \dims^{\frac{1}{3}})$ & $\internsteps = \condition^
  {\frac{3}{4}}$ &$\red{\mathbf{\dims\condition^{\frac{3}{4}}}}$
  & $\dims^{\frac{2}{3}}\condition^{\frac{7}{4}}$
  & $\dims^{\frac{1}{2}}\condition^{\frac{3}{2}}$
  & $\dims^{\frac{2}{3}}\condition^{\frac{13}{12}}$
  & $\red{\mathbf{\dims\condition^{\frac{3}{4}}}}$
  & $\dims^{\frac{1}{2}}\condition^{\frac{7}{4}}$\\[3mm]
  $\condition \in [\dims^{\frac{1}{3}}, \dims^{\frac{2}{3}}]$
  & $\internsteps = \dims^{\frac{1}{4}}$
  & $\dims^{\frac{5}{4}}$
  & $\red{\mathbf{\dims^{\frac{11}{12}}\condition}}$
  & $\dims^{\frac{1}{2}}\condition^{\frac{3}{2}}$
  & $\dims^{\frac{7}{12}}\condition^{\frac{4}{3}}$
  & $\red{\mathbf{\dims^{\frac{11}{12}}\condition}}$
  & $\dims^{\frac{7}{12}}\condition^{\frac{3}{2}}$\\[3mm]
  $\condition \in (\dims^{\frac{2}{3}}, \dims]$
  & $\internsteps = \dims^{\frac{3}{4}}\condition^{-\frac{3}{4}}$
  & $\dims^{\frac{7}{4}}\condition^{-\frac{3}{4}}$
  & $\dims^{\frac{19}{12}}\condition^{\frac{1}{4}}$
  & $\dims^{\frac{1}{2}}\condition^{\frac{3}{2}}$
  & $\dims^{\frac{5}{12}}\condition^{\frac{19}{12}}$
  & $\red{\mathbf{\dims^{\frac{3}{4}}\condition^{\frac{5}{4}}}}$
  & $\red{\mathbf{\dims^{\frac{3}{4}}\condition^{\frac{5}{4}}}}$\\[3mm]
  $\condition \in (\dims, \infty]$ & $\internsteps = 1$  & $\dims$
  & $\dims^{\frac{2}{3}}\condition$
  & $\red{\mathbf{\dims^{\frac{1}{2}}\condition^{\frac{3}{2}}}}$
  & $\dims^{\frac{2}{3}}\condition^{\frac{4}{3}}$
  & $\dims\condition$
  & $\red{\mathbf{\dims^{\frac{1}{2}}\condition^{\frac{3}{2}}}}$\\[3mm]
  \bottomrule
  \end{tabular}
  \caption{Trade-off between the the six terms $T_i, i =1, \ldots 6$, from
  the bound~\eqref{eq:step_condition_explicit_K_appendix}
  under the assumption $\hessianLipRatio={\hessianLip^{{2}/{3}}}/{\smoothness}
  \leq 1$.
  In the second column, we provide the optimal choice of $\internsteps$
  for the condition on $\condition$ stated in first column
  such that the maximum of the $T_i$'s is smallest.
  For each row the dominant (maximum) term, and equivalently the effective
  bound on $\mathcal{G}_
  {\text{eval}}$ is displayed in bold (red).}
  \label{tab:hmc_param_choice_explicit}
\end{table}

\subsubsection{Faster mixing time bounds} 
\label{ssub:faster_mixing_time_bounds}
We now derive several mixing time bounds under additional assumptions:
(a) when a warm start is available, and (b) the Hessian-Lipschitz constant
is small.

\paragraph{Faster mixing time with warm start:} 
\label{par:faster_mixing_}
When a better initialization with warmness
\mbox{$\warmparam \leq O(e^{\dims^{\frac{2}{3}}\condition})$} is available,
and suppose that $\condition$ is much smaller than $\dims$.
In such a case, the optimal choice turns out to be
$\internsteps = \dims^ {\frac{1}{4}}$ (instead of $\condition^{\frac{3}
{4}}$) which implies a bound of  $O\parenth{\dims^{\frac{11}{12}}\condition
\log\parenth{\frac{1}{\errorparam}}}$ on $\mathcal{G}_{\text{eval}}$ (this
bound was also stated in  Table~\ref{tab:mixing_times_all}).

\paragraph{Faster mixing time with small $\hessianLip$:}
Suppose in addition to warmness being not too large, \mbox{$\warmparam
\leq O(e^{\dims^{\frac{2}{3}}\condition})$}, the Hessian-Lipschitz constant
$\hessianLip$ is small enough $\hessianLip^{\frac{2}{3}} \ll \smoothness$.
In such a scenario, the terms $T_5$ and $T_6$ become negligible because
of small $\hessianLip$ and $T_1$ is negligible because of small $\warmparam$.
The terms $T_3$ and $T_4$ remain unchanged, and the term $T_2$ changes
slightly. More precisely, for the case $\hessianLip^{\frac{2}{3}} \leq
\frac{\smoothness}{\dims^{\frac{1}{2}}\condition^{\frac{1}{2}}}$ we obtain
a slightly modified trade-off for the terms in
the~\eqref{eq:step_condition_explicit_K_appendix} for
$\mathcal{G}_{\text{eval}}$ (summarized in
Table~\ref{tab:hmc_param_choice_explicit_L3}). If $\condition$ is small
too, then we obtain a mixing time bound of order $\dims^{\frac{5}{8}}$.
Via this artificially constructed example, we wanted to demonstrate two
things. First, faster convergence rates are possible to derive under additional
assumptions directly from our results. Suitable adaptation of our proof
techniques might provide a faster rate of mixing for Metropolized HMC under
additional assumptions like infinity semi-norm regularity condition made
in other works~\citep{mangoubi2018dimensionally} (but we leave a detailed
derivation for future work). Second, it also demonstrates the looseness
of our proof techniques since we were unable to recover an $O(1)$ mixing time bound
for sampling from a Gaussian target.

\begin{table}[ht]
\centering
  \begin{tabular}{c|c|cccccc} \toprule $\condition$ versus $\dims$ &
  $\internsteps$ optimal choice & $T_1$ & $T_2$ & $T_3$ & $T_4$ &
  $T_5$ & $T_6$ \\[3mm] & & - &
  $\orange{\internsteps \dims^{\frac{1}{2}} \condition}$ &
  $\dims^{\frac{1}{2}}\condition^{\frac{3}{2}}$ &
  $\internsteps^{-\frac{1}{3}}\dims^{\frac{2}{3}}\condition^{\frac{4}{3}}$
  & - & - \\[3mm] $\condition \in (0, \dims^{\frac{1}{2}})$ &
  $\internsteps = \dims^{\frac{1}{8}}\condition^{\frac{1}{4}}$
  & - &
  $\red{\mathbf{\dims^{\frac{5}{8}}\condition^{\frac{5}{4}}}}$ &
  $\dims^{\frac{1}{2}}\condition^{\frac{3}{2}}$ &
  $\red{\mathbf{\dims^{\frac{5}{8}}\condition^{\frac{5}{4}}}}$
  & - & - \\[3mm] \bottomrule
  \end{tabular} \caption{Six terms in the HMC
  number of gradient evaluations bound under small hessian-Lipschitz
  constant and very warm start. The dominant term is highlighted in
  red.}  \label{tab:hmc_param_choice_explicit_L3}
\end{table}

\paragraph{Linearly transformed HMC (effect of mass function):}
\label{sub:the_effect_of_linear_transformations}
In practice, it is often beneficial to apply linear transformations in
HMC (cf. Section 4 in~\cite{neal2011mcmc}).
At a high level, such a transformation can improve the conditioning of the
problem and help HMC mix faster. For the target distribution $\target$ with
density proportional to $e^{-\targetf}$, we can define a new
distribution $\Pi_{\targetg}$ with density $e^ {-\targetg}$ (up to
normalization) such that $\targetg (x) = \targetf(\hmcmass^
{-\frac{1}{2}}x)$ where $\hmcmass \in \real^{\dims\times\dims}$
is an invertible matrix. Then for a random sample $\tilde{\hmcstate}
\sim \Pi_\targetg$, the distribution of
$\hmcmass^{\frac{1}{2}} \tilde{\hmcstate}$ is $\target$. When the new
distribution $\targetg$ has a better condition number $\condition_\targetg$
than the condition number $\condition$ of
$\targetf$, we can use HMC to draw approximate sample from $\Pi_\targetg$
and then transform the samples using the matrix $\hmcmass$.
Clearly the bound from Corollary~\ref{cor:hmc_mixing_sc_general} guarantees
that when $\condition_\targetg $ is much smaller than $\condition$, HMC
on the new target $\Pi_\targetg$ would mix much faster than the HMC chain on $\target$.
This transformation is equivalent to the HMC
algorithm with modified kinetic energy
\begin{align*}
  \frac{d\hmcstate_t}{dt}
  = \hmcmass^{-1} \hmcnoise_t
  \quad\text{and}\quad
  \frac{d\hmcnoise_t}{dt}
  = - \gradf(\hmcstate_t),
\end{align*}
which is easier to implement in practice.
For a detailed discussion of this implementation,
we refer the readers to the paper by \cite{neal2011mcmc}.


\subsection{Comparison with guarantees for unadjusted versions of HMC}
\label{sub:comparison_with_guarantees_for_other_versions_of_hmc}

In this appendix, we compare our results with mixing time guarantees
results on unadjusted and ODE solver based HMC chains.  We summarize
the number of gradient evaluations needed for Metropolized HMC to mix
and those for other existing sampling results in
Table~\ref{tab:mixing_times_with_poly}. Note that all the results
summarized here are the best upper bounds in the literature for
log-concave sampling.  We present the results for a
$(\smoothness, \hessianLip,
\scparam)$-regular target distribution. We remark that all methods presented
in Table~\ref{tab:mixing_times_with_poly} requires the regularity
assumptions~\eqref{eq:assumption_smoothness}
and~\eqref{eq:assumption_scparam}, even though some do not require
assumption~\eqref{eq:assumption_hessianLip}.

\begin{table}[ht]
    \centering
    {\renewcommand{\arraystretch}{.8} \begin{tabular}{ccc} \toprule \thead{ \bf
    Sampling algorithm} & \thead{ \bf \#Grad. evals} \\ \midrule
        %
        \thead{ $^{\ddagger, \diamond}$Unadjusted HMC with \\  leapfrog integrator~\citep{mangoubi2018dimensionally}}
        & $\dims^{\frac{1}{4}}\condition^{\frac{11}{4}} \cdot
        \frac{1}{\errorparam^{1/2}}$ \\[4mm]
        \thead{$^{\ddagger}$Underdamped Langevin~\citep{cheng2017underdamped}}
        & $\dims^{\frac{1}{2}} \condition^{2} \cdot \frac{1}
        {\errorparam}$ \\[4mm]
        \thead{ $^{\ddagger}$HMC with ODE solver, Thm 1.6 in~\citep{lee2018algorithmic}}
        & $\dims^{\frac{1}{2}}\condition^{\frac{7}{4}} \cdot \frac{1}{\errorparam}$ \\[4mm]
        \thead{ $^\star$MALA~\citep{dwivedi2018log}[this paper]}
        & $\max\braces{\dims \condition, \dims^{\frac{1}{2}} \condition^{\frac{3}{2}}} \cdot \log\frac{1}{\errorparam}$\\[4mm]
        \thead{$^\star$Metropolized HMC with \\ leapfrog integrator [this paper]}
        & $\max\braces{\dims \condition^{\frac{3}{4}},
        \dims^{\frac{11}{12}} \condition,
        \dims^{\frac{3}{4}} \condition^{\frac{5}{4}},
        \dims^{\frac{1}{2}} \condition^{\frac{3}{2}}
        } \cdot \log\frac{1}
        {\errorparam}$\\[2mm]
        \bottomrule
    \end{tabular}
    }
    \caption{Summary of the number of gradient evaluations needed for
    the sampling algorithms to converge to a
    $(\scparam,\smoothness, \hessianLip)$-regular target distribution
    with $\hessianLip = O(\smoothness^{\frac{3}{2}})$ within
    $\errorparam$ error from the target distribution (in
    total-variation distance$^\star$ or $1$-Wasserstein
    distance$^\ddagger$) (and $\diamond$ certain additional regularity
    conditions for the result by~\cite{mangoubi2018dimensionally}).  Note that the unadjusted
    algorithms suffer from an exponentially worse dependency on
    $\errorparam$ when compared to the Metropolis adjusted chains. For
    MALA, results by~\cite{dwivedi2018log} had an extra
    $\dims$ factor which is sharpened in
    Theorem~\ref{thm:mala_mrw_mixing} of this paper.
    } \label{tab:mixing_times_with_poly}
\end{table}

Two remarks are in order.
First, the error metric for the guarantees in the works~\citep{mangoubi2018dimensionally,cheng2017underdamped,lee2018algorithmic}
is $1$-Wasserstein distance, while our results make use of $\Ell_2$ or TV
distance. As a result, a direct comparison between these results is not
possible although we provide an indirect comparison below.
Second, the previous guarantees have a polynomial dependence on the inverse
of error-tolerance $1/\errorparam$. In contrast, our results for MALA and
Metropolized HMC have a logarithmic dependence $\log(1/\errorparam)$. For a
well-conditioned target, i.e., when $\condition$ is a constant, all
prior results have a better dependence on $\dims$ when compared to our bounds.

\paragraph{Logarithmic vs polynomial dependence on $1/\errorparam$:} 
\label{par:logarithmic_vs_polynomial_dependence_on_}
We now provide an indirect comparison, between prior guarantees based on
Wasserstein distance and our results based on TV-distance,
for estimating expectations of Lipschitz-functions on bounded domains.
MCMC algorithms are used to estimate expectations of certain functions of
interest. Given an arbitrary function $g$ and an MCMC algorithm, one of
the ways to estimate \mbox{$\target(g):= \Exs_{X\sim\target}[g(X)]$} is to
use the $k$-th iterate from $N$ independent runs of the chain.
Let  $X_i^{(k)}$ for $i=1, \ldots, N$ denote the $N$ i.i.d. samples
at the $k$-th iteration of the chain and let $\mu_k$ denote the distribution
of $X_i^{(k)}$, namely the distribution of the chain after $k$ iterations.
Then for the estimate $\widehat{\Pi}_k(g) \defn \frac{1}{N} \sum_{i=1}^N
g(X_i^{(k)})$, the estimation error can be decomposed as
\begin{align}
\label{eq:error_expectation_estimation}
\target(g) - \widehat{\Pi}_k(g) &= \int_{\real^\dims}
  g(x) \targetdensity(x) dx - \frac{1}{N} \sum_{i=1}^N g(X_i^{(k)}) \notag
  \\ &
  = \underbrace{\int_{\real^\dims} g(x) \brackets{\targetdensity(x)
  - \mu_k(x)} dx}_{=:J_1\ \text{(Approximation bias)}}
  + \underbrace{\Exs_{\mu_k}\brackets{g(X)}
  - \frac{1}{N} \sum_{i=1}^N g(X_i^{(k)})}_{=:J_2\ \text{(Finite sample
  error)}}.
\end{align}
To compare different prior works, we assume that $\Var_{\mu_k}\brackets{g
(X_1)}$ is bounded and thereby that the finite sample error $J_2$ is negligible
for large enough $N$.\footnote{Moreover, this error should be usually similar
across different sampling algorithms since several algorithms are designed
in a manner agnostic to a particular function $g$.}
It remains to bound the error $J_1$ which can be done in two different ways
depending on the error-metric used to provide mixing time guarantees for
the Markov chain.

If the function $g$  is
$\omega$-Lipschitz and $k$ is chosen such that $\mathcal{W}_1(\target,\mu_k)
\leq \errorparam$, then we have $J_1 \leq \omega\errorparam =:J_{\text{Wass}}$.
On the other hand, if the function $g$ is bounded by $B$, and $k$ is chosen
such that $\tvdist{\target}{\mu_k} \leq \errorparam$, then we obtain the
bound $J_1 \leq B\errorparam =:J_{\text{TV}}$. We make use of these two
facts to compare the number of gradient evaluations needed by
unadjusted HMC or ODE solved based HMC and Metropolized HMC.
Consider an $\omega$-Lipschitz function $g$ with support on a ball of radius
$R$. Note that this function is uniformly bounded by $B=\omega R$.
Now in order to to ensure that $J_1 \leq \delta$ (some user-specified small
threshold),
the choice of $\errorparam$ in the two cases (Wasserstein and TV distance)
would be different leading to different number of gradient evaluations required
by the two chains. More precisely, we have
\begin{align*}
  J_1&\leq J_{\text{Wass}} = \omega\errorparam \leq \delta \quad\Longrightarrow
  \quad
  \errorparam_{\text{wass}} = \frac{\delta}{\omega}\quad\text{and}\\
  J_1&\leq J_{\text{TV}} = B\errorparam=\omega R\errorparam \leq \delta
  \quad\Longrightarrow \quad
  \errorparam_{\text{TV}} = \frac{\delta}{\omega R}.
\end{align*}
To simplify the discussion, we consider well-conditioned
(constant $\condition$) strongly log-concave distributions such that most
of the mass is concentrated on a ball of radius
$O(\sqrt{d})$ (cf. Appendix~\ref{sub:proof_of_lemma_4}) and consider $R=
\sqrt{d}$.
Then plugging the error-tolerances from the display above in Table~\ref{tab:mixing_times_with_poly},
we obtain that the number of gradient evaluations $\mathcal{G}_{\text{MC}}$
for different chains\footnote{The results for other HMCs often assume
(different) additional conditions so that a direct comparison should be
taken with a fine grain of salt.} would scale as
\begin{align*}
   \mathcal{G}_{\text{unadj.-HMC}}\leq O(\sqrt{\frac{d\omega}{\delta}}),\quad
   \mathcal{G}_{\text{ODE-HMC}}\leq O(\frac{\omega\sqrt{d}}{\delta}),\text{and}\quad
   \mathcal{G}_{\text{Metro.-HMC}}\leq O(d\log\frac{\omega\sqrt{d}}
   {\delta})
\end{align*}
Clearly, depending on $\omega$ and the threshold $\delta$, different chains
would have better guarantees. When $\omega$ is large or $\delta$ is small,
our results ensure the superiority of Metropolized-HMC over other versions.
For example, higher-order moments can be functions of interest, i.e.,
$g(x) = \Vert x\Vert^{1+\nu}$ for which the Lipschitz-constant $\omega=O
(d^{\nu})$ scales with $\dims$. For this function, we obtain the bounds:
\begin{align*}
  \mathcal{G}_{\text{unadj.-HMC}}\leq O(\frac{d^{\frac{1+\nu}{2}}}{\sqrt{\delta}}),
  \quad
   \mathcal{G}_{\text{ODE-HMC}}\leq
   O(\frac{d^{\frac{1}{2}+\nu}}{\delta}), \quad \text{and}\quad
   \mathcal{G}_{\text{Metro.-HMC}}\leq O(d(1+\nu)\log\frac{d}{\delta})
\end{align*}
and thus Metropolized HMC takes fewer gradient evaluations than ODE-based
HMC for $\nu>1/2$ and unadjusted HMC for $\nu>1$ (to ensure $J_1\leq\delta$~\eqref{eq:error_expectation_estimation}).
We remark that the bounds for unadjusted-HMC require additional
regularity conditions. From this informal comparison, we demonstrate that both the dimension dependency $\dims$ and error dependency $\errorparam$ should be accounted for comparing unadjusted algorithms and Metropolized algorithms. Especially for estimating high-order moments, Metropolized algorithms with $\log(\frac{1}{\errorparam})$ dependency will be advantageous.
